%% file: main.tex
\documentclass[10pt]{article}
\usepackage[margin=1in]{geometry}
\usepackage{mathpazo}
\usepackage{booktabs}

\usepackage[utf8]{inputenc} 
\usepackage{url}            
\usepackage{amsfonts}       
\usepackage{nicefrac}       
\usepackage{microtype}      
\usepackage{xcolor}         

\usepackage{bbm}
\usepackage{amsmath,amssymb,amsthm}
\usepackage{xspace}
\usepackage{graphicx}
\usepackage{color}
\usepackage[colorlinks=true,linkcolor=blue,citecolor=blue,urlcolor=blue]{hyperref}
\usepackage{todonotes}
\usepackage{subcaption}
\usepackage{comment}
\usepackage{algorithm,algpseudocode}
\usepackage[T1]{fontenc}
\usepackage[backend=biber,style=alphabetic,natbib=true,maxbibnames=99,minalphanames=3]{biblatex}
\usepackage{makecell}

\usepackage{mathtools}

\newcommand{\saachi}[1]{}

\title{A Data-Based Perspective on Transfer Learning}
\author{
	Saachi Jain\footnotemark[1] \\
	MIT \\
	\texttt{saachij@mit.edu}
    \and
    Hadi Salman\thanks{Equal contribution.} \\
	MIT \\
	\texttt{hady@mit.edu} \\
	\and
    Alaa Khaddaj \footnotemark[1] \\
	MIT \\
    \texttt{alaakh@mit.edu}
    \and
    Eric Wong \\
    MIT \\
    \texttt{wongeric@mit.edu}
    \and
    Sung Min Park \\
	MIT \\
    \texttt{sp765@mit.edu}
    \and
	Aleksander M\k{a}dry \\
	MIT \\
    \texttt{madry@mit.edu}
}
\date{}
\makeatletter
\renewcommand{\paragraph}{%
  \@startsection{paragraph}{4}%
  {\z@}{1.5ex \@plus 1ex \@minus .2ex}{-1em}%
  {\normalfont\normalsize\bfseries}%
}
\makeatother

\begin{document}
    \maketitle
    \begin{abstract}
        \input{abstract}
    \end{abstract}

    \section{Introduction}
    \label{sec:intro}
    \input{sections/intro}
    \section{A Data-Based Framework for Studying Transfer Learning}
    \label{sec:framework}
    \input{sections/framework}

    \section{Identifying the Most Influential Classes of the Source Dataset}
    \label{sec:counterfactuals}
    \input{sections/counterfactuals}

    \section{Probing the Impact of the Source Dataset on Transfer Learning}
    \label{sec:science}
    \input{sections/science}

    \section{Related Work}
    \label{sec:related}
    \input{sections/related}

    \section{Conclusions}
    \label{sec:conclusion}
    \input{sections/conclusion}

    \section{Acknowledgements}
    \input{sections/ack.tex}

    \clearpage
    \printbibliography

    \clearpage
    \appendix

    \section{Experimental Setup}
    \label{app:setup}
    \input{sections/app_setup.tex}

    \clearpage
    \section{Variants of Computing Influences}
    \label{app:variants-influences}
    \input{sections/app_variants_influences.tex}

    \clearpage
    \section{Full Counterfactual Experiment}
    \label{app:counterfactual_all_datasets}
    \input{sections/app_counterfactual.tex}

    \clearpage
    \section{Adapting our Framework to Compute the Effect of Every Source Datapoint on Transfer Learning}
    \label{app:example-based}

\input{sections/app_example_based.tex}

    \clearpage
    \section{Omitted Results}
    \label{app:omitted}
    \input{sections/app_omitted.tex}

    \clearpage
    \section{Further Convergence Analysis}
    \label{app:convergence_analysis_extended}
    \input{sections/app_convergence.tex}

\end{document}

%% file: abstract.tex
It is commonly believed that in transfer learning including more pre-training data translates into better performance. However, recent evidence suggests that \emph{removing} data from the source dataset can actually help too. In this work, we take a closer look at the role of the source dataset's composition in transfer learning and present a framework for probing its impact on downstream performance. Our framework gives rise to new capabilities such as pinpointing transfer learning brittleness as well as detecting pathologies such as data-leakage and the presence of misleading examples in the source dataset. In particular, we demonstrate that removing detrimental datapoints identified by our framework improves transfer learning performance from ImageNet on a variety of target tasks.\footnote{Code is available at \url{https://github.com/MadryLab/data-transfer}}

%% file: sections/intro.tex
Transfer learning enables us to adapt a model trained on a \textit{source} dataset to perform better on a downstream \textit{target} task. This technique is employed in a range of machine learning applications including radiology \cite{wang2017chestx, ke2021chextransfer}, autonomous driving \cite{kim2017end, du2019self}, and satellite imagery analysis \cite{xie2016transfer, wang2019crop}. Despite its successes, however, it is still not clear what the drivers of performance gains brought by transfer learning actually are.

So far, a dominant approach to studying these drivers focused on the role of the source \textit{model}---i.e., the model trained on the source dataset. The corresponding works involve investigating the source model's architecture~\cite{ke2021chextransfer}, accuracy~\cite{kornblith2019better}, adversarial vulnerability~\cite{salman2020adversarially, utrera2020adversarially}, and training procedure~\cite{jang2019learning, kumar2022fine}. This line of work makes it clear that the properties of the source model has a significant impact on transfer learning. There is some evidence, however, that the source \textit{dataset} might play an important role as well~\cite{huh2016makes,ngiam2018domain,kolesnikov2019big}. For example, several works have shown that while increasing the size of the source dataset generally boosts transfer learning performance, \emph{removing} specific classes can help too~\cite{huh2016makes,ngiam2018domain,kolesnikov2019big}. All of this motivates a natural question:

\begin{center}
    \textit{How can we pinpoint the exact impact of the source dataset in transfer learning?}
\end{center}



\paragraph{Our Contributions.} In this paper, we present a framework for measuring and analyzing the impact of the source dataset's composition on transfer learning performance. In particular, our framework allows us to study how the quality of the transfer learning model's predictions changes when subsets of the source dataset are removed. This enables us, in turn, to automatically find subsets of the source dataset that---positively or negatively---impact downstream behavior. Using our framework, we can:

\begin{itemize}
    \item Pinpoint what parts of the source dataset are most utilized by the downstream task.
    \item Automatically extract granular subpopulations in the target dataset through projection of the fine-grained labels of the source dataset.
    \item Remove detrimental data from the source dataset to improve transfer learning performance.
\end{itemize}

Finally, we demonstrate how our framework can be used to find subsets of ImageNet~\cite{deng2009imagenet} that, when removed, give rise to better downstream performance on a variety of image classification tasks.

%% file: sections/framework.tex
In order to pinpoint the role of the source dataset in transfer learning, we need to understand how the composition of that source dataset impacts the downstream model's performance.
To do so, we draw inspiration from supervised machine learning approaches that study the impact of the training data on the model's subsequent predictions. In particular, these approaches capture this impact via studying (and approximating) the counterfactual effect of excluding certain training datapoints. This paradigm underlies a number of techniques, from influence functions~\cite{cook1982residuals, koh2017understanding, feldman2020what}, to datamodels~\cite{ilyas2022datamodels}, to data Shapley values~\cite{kwon2021beta, karlavs2022data, ghorbani2019data}.

Now, to adapt this paradigm to our setting, we study the counterfactual effect of excluding datapoints from the {\em source} dataset on the downstream, {\em target} task predictions. In our framework, we will focus on the inclusion or exclusion of entire {\em classes} in the source dataset, as opposed to individual examples\footnote{In Section~\ref{sec:example_wise}, we adapt our framework to calculate more granular influences of individual source examples too.}. This is motivated by the fact that, intuitively, we expect these classes
to be the ones that embody whole concepts and thus drive  the formation of (transferred) features. We therefore anticipate the removal of entire classes to have a more measurable impact on the representation learned by the source model (and consequently on the downstream model’s predictions).



Once we have chosen to focus on removal of entire source classes, we can design counterfactual experiments to estimate their influences\saachi{Alt option: Now that we have chosen to focus on source classes, how can we measure their influences?}. A natural approach here, the \textit{leave-one-out} method ~\cite{cook1982residuals, koh2017understanding}, would involve removing each individual class from the source dataset separately and then measuring the change in the downstream model's predictions. However, in the transfer learning setting, we suspect that removing a single class from the source dataset won't significantly change the downstream model's performance. Thus, leave-one-out methodology may be able to capture meaningful influences only in rare cases. This is especially so as many common source datasets contain highly redundant classes. For example, ImageNet contains over 100 dog-breed classes. The removal of a single dog-breed class might thus have a negligible impact on transfer learning performance, but the removal of all of the dog classes might significantly change the features learned by the downstream model. For these reasons, we adapt the \textit{subsampling}~\cite{feldman2020what,ilyas2022datamodels} approach, which revolves around removing a random collection of source classes at once.


\paragraph{Computing transfer influences.} In the light of the above, our methodology for computing the influence of source classes on transfer learning performance involves training a large number\footnote{In this paper, we train 7540 models. Details are in Appendix~\ref{app:setup}.} of models with random subsets of the source classes removed, and fine-tuning these models on the target task. We then estimate the influence value of a source class $\mathcal{C}$ on a target example $t$ as the expected difference in the transfer model's performance on example $t$ when class $\mathcal{C}$ was either included in or excluded from the source dataset:
\begin{align}
    \label{eq:influence}
    \text{Infl}[\mathcal{C} \rightarrow t] = \mathbb{E}_{S} \left[ f(t; S) ~|~ \mathcal{C} \subset S \right] - \mathbb{E}_{S} \left[ f(t; S) ~|~ \mathcal{C} \not \subset S \right]\text{,}
\end{align}
where $f(t; S)$ is the softmax output\footnote{We experiment with other outputs such as logits, margins, or correctness too. We discuss the corresponding results in Appendix~\ref{app:variants-influences}.} of a model trained on a subset $S$ of the source dataset. A positive influence value indicates that including the source class $\mathcal{C}$ helps the model predict the target example $t$ correctly. On the other hand, a negative influence value suggests that the source class $\mathcal{C}$ actually hurts the model's performance on the target example $t$.  We outline the overall procedure in Algorithm~\ref{alg:main}, and defer a detailed description of our approach to Appendix~\ref{app:setup}.


\begin{algorithm}[!hbtp]
    \caption{Estimation of source dataset class influences on transfer learning performance.}
    \label{alg:main}
    \begin{algorithmic}[1]
    \Require Source dataset $\mathcal{S} = \cup_{k=1}^{K}~\mathcal{C}_k$ (with $K$ classes), a target dataset $\mathcal{T} = (t_1, t_2,\cdots,t_n$), training algorithm $\mathcal{A}$, subset ratio $\alpha$, and number of models $m$
    \State Sample $m$ random subsets $S_1,S_2,\cdots,S_m \subset \mathcal{S}$ of size $\alpha \cdot |\mathcal{S}|$:
    \For{$i \in  1$ to $m$}
        \State Train model $f_i$ by running algorithm $\mathcal{A}$ on $S_{i}$
    \EndFor
    \For{$k \in  1$ to $K$}
    \For{$j \in  1$ to $n$}
        \State$ \text{Infl}[\mathcal{C}_k \rightarrow t_j] =
        \frac{
            \sum_{i=1}^{m} f_i(t_j; S_i) \mathbbm{1}_{\mathcal{C}_k \subset S_i}
        }{\sum_{i=1}^m \mathbbm{1}_{\mathcal{C}_k \subset S_i}
        } - \frac{
            \sum_{i=1}^{m} f_i(t_j; S_i) \mathbbm{1}_{\mathcal{C}_k \not \subset S_i}
        }{\sum_{i=1}^m \mathbbm{1}_{\mathcal{C}_k \not \subset S_i}} $
    \EndFor
    \EndFor
    \State
    \Return $\text{Infl}[\mathcal{C}_k \rightarrow t_j]\text{,}~~\text{for all}~j\in[n], k\in[K]$
    \end{algorithmic}
\end{algorithm}

%% file: sections/counterfactuals.tex
In Section~\ref{sec:framework}, we presented a framework for pinpointing the role of the source dataset in transfer learning by estimating the influence of each source class on the target model's predictions. Using these influences, we can now take a look at the classes from the source dataset that have the largest positive or negative impact on the overall transfer learning performance. We focus our analysis on the fixed-weights transfer learning setting (and defer results for full model fine-tuning to Appendix~\ref{app:omitted}).

\begin{figure}[!t]
    \centering
        \includegraphics[width=0.9\linewidth]{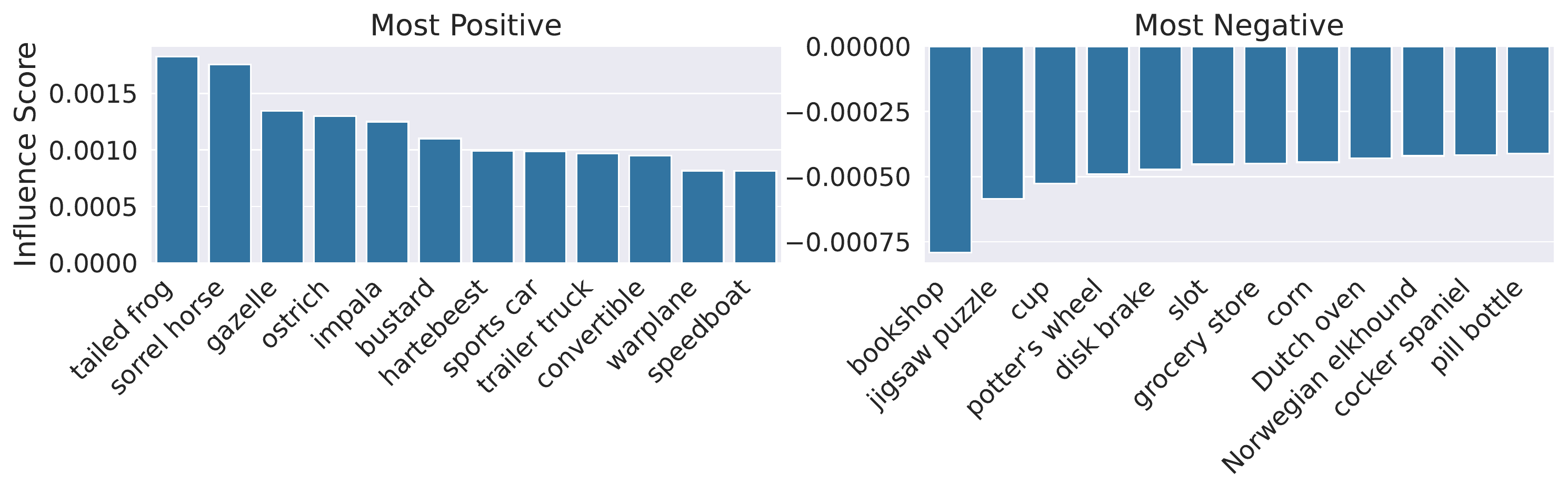}
    \caption{Most positive and negative ImageNet classes ordered based on their overall influence on the CIFAR-10 dataset. The top source classes (e.g., tailed frog and sorrel horse) turn out to be semantically relevant to the target classes (e.g., frog and horse).}
    \label{fig:influencers-cifar-overall}
\end{figure}

As one might expect, not all source classes have large influences. Figure \ref{fig:influencers-cifar-overall}  displays the most influential classes of ImageNet with CIFAR-10 as the target task. Notably, the most positively influential source classes turn out to be directly related to classes in the target task (e.g., the ImageNet label ``tailed frog'' is an instance of the CIFAR class ``frog''). This trend holds across all of the target datasets and transfer learning settings we considered (see Appendix~\ref{app:counterfactual_all_datasets}). Interestingly, the source dataset also contains classes that are overall negatively influential for the target task (e.g., ``bookshop'' and ``jigsaw puzzle'' classes). (In Section~\ref{sec:science}, we will take a closer look at the factors that can cause a source class to be negatively influential for a target prediction.)

\begin{figure}[!t]
    \begin{subfigure}[c]{0.44\linewidth}
        \vspace{0pt}
        \includegraphics[width=\linewidth]{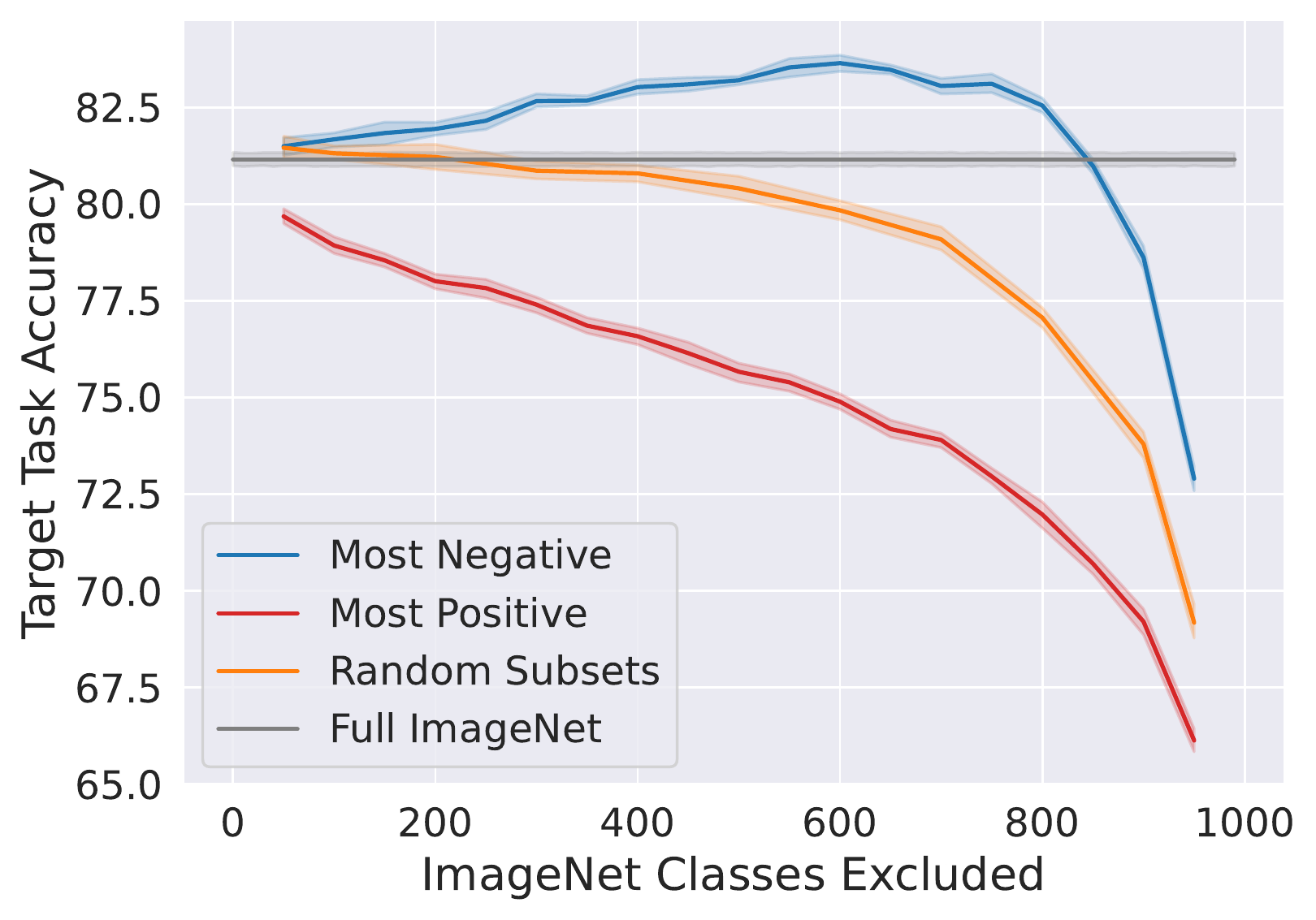}\hfill
        \caption{CIFAR-10 results}
        \label{fig:cifar_detailed_fixed_cf}
    \end{subfigure}\hfill
    \begin{subfigure}[c]{0.54\linewidth}
        \vspace{0pt}
        \resizebox{\linewidth}{!}{
        \begin{tabular}{c|ccc}
            \toprule
            & \multicolumn{3}{c}{\textit{Source Dataset}}\\
            \textit{Target Dataset} & Full ImageNet & \makecell{Removing \\Bottom Infl.} & Hand-picked\\
            \midrule
            AIRCRAFT & $36.08 \pm 1.07$ & $\mathbf{36.88 \pm 0.74}$ & N/A \\
            BIRDSNAP & $38.42 \pm 0.40$ & $\mathbf{39.19 \pm 0.38}$ & $26.74 \pm 0.31$\\
            CALTECH101 & $86.69 \pm 0.79$ & $\mathbf{87.03 \pm 0.30}$ & $82.28 \pm 0.40$\\
            CALTECH256 & $74.97 \pm 0.27$ & $\mathbf{75.24 \pm 0.21}$ & $67.42 \pm 0.39$\\
            CARS & $39.55 \pm 0.32$ & $\mathbf{40.59 \pm 0.57}$ & $21.71 \pm 0.40$\\
            CIFAR10 & $81.16 \pm 0.30$ & $\mathbf{83.64 \pm 0.40}$ & $75.53 \pm 0.42$\\
            CIFAR100 & $59.37 \pm 0.58$ & $\mathbf{61.46 \pm 0.59}$ & $55.21 \pm 0.52$\\
            FLOWERS & $\mathbf{82.92 \pm 0.52}$ & $82.89 \pm 0.48$ &N/A\\
            FOOD & $56.19 \pm 0.14$ & $\mathbf{56.85 \pm 0.27}$ & $39.36 \pm 0.39$\\
            PETS & $83.41 \pm 0.55$ & $\mathbf{87.59 \pm 0.24}$ & $87.16 \pm 0.24$\\
            SUN397 & $50.15 \pm 0.23$ & $\mathbf{51.34 \pm 0.29}$ &N/A\\
            \bottomrule
        \end{tabular}
        }
        \caption{Summary of 11 target tasks}
        \label{fig:summary_fixed_cf}
    \end{subfigure}

    \caption{Target task accuracies after removing the K most positively or negatively influential ImageNet classes from the source dataset. Mean/std are reported over 10 runs. \textbf{(a)} Results with CIFAR-10 as the target task after removing different numbers of classes from the source dataset. We also include baselines of using the full ImageNet dataset and removing random classes. One can note that, by removing negatively influential source classes, we can obtain a test accuracy that is 2.5\% larger than what using the entire ImageNet dataset would yield.  Results for other target tasks can be found in Appendix~\ref{app:counterfactual_all_datasets}. \textbf{(b)} Peak performances when removing the most negatively influential source classes across a range of other target tasks. We compare against using the full ImageNet dataset or a relevant subset of classes (hand-picked, see Appendix~\ref{app:setup} for details).}
    \label{fig:cf_fixed_cifar}
\end{figure}

\paragraph{How important are the most influential source classes?} We now remove each of the most influential classes from the source dataset to observe their actual impact on transfer learning performance (Figure~\ref{fig:cifar_detailed_fixed_cf}). As expected, removing the most positively influential classes severely degrades transfer learning performance as compared to removing random classes. This counterfactual experiment confirms that these classes are indeed important to the performance of transfer learning. On the other hand, removing the most negatively influential classes actually improves the overall transfer learning performance \emph{beyond what using the entire ImageNet dataset provides} (see Figure~\ref{fig:summary_fixed_cf}).

%% file: sections/science.tex
\begin{figure}[!t]
    \centering
    \begin{subfigure}[t]{0.9\linewidth}
        \centering
        \includegraphics[width=\linewidth]{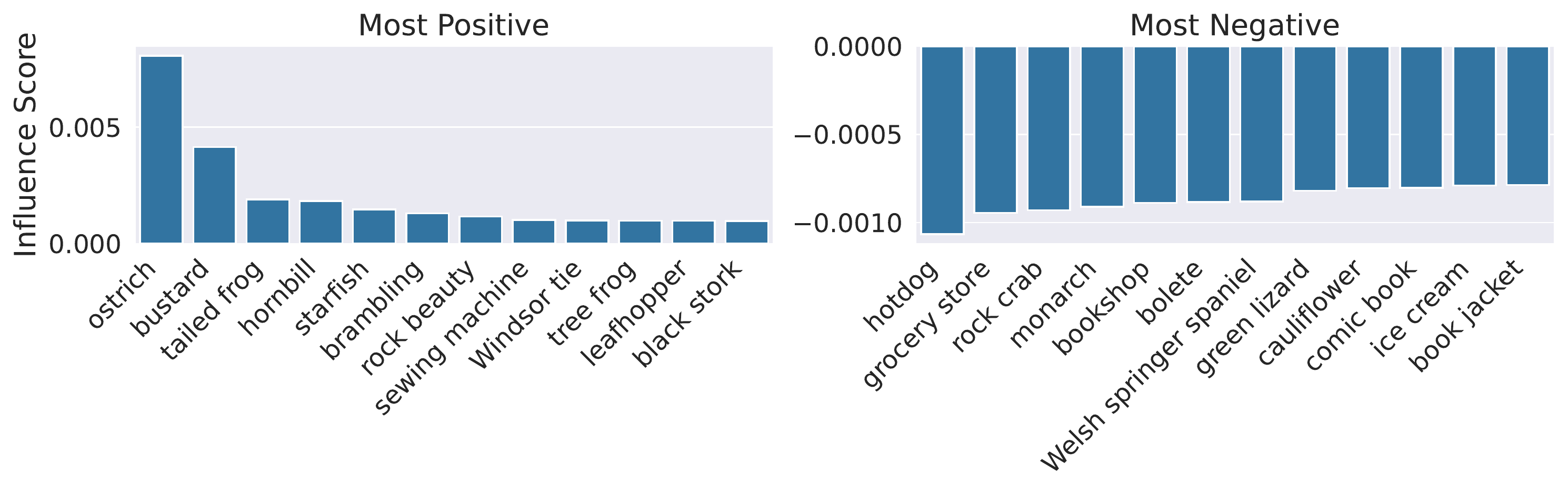}
    \end{subfigure}
    \caption{Most positive and negative influencing ImageNet classes for the CIFAR-10 class ``bird''. These are calculated by averaging the influence of each source class over all bird examples. We find that the most positively influencing ImageNet classes (e.g., ``ostrich'' and ``bustard'') are related to the CIFAR-10 class ``bird''. See Appendix~\ref{app:omitted} for results on other CIFAR-10 classes.}
    \label{fig:class-barplots}
\end{figure}

\begin{figure}[!t]
    \centering
    \includegraphics[width=\linewidth]{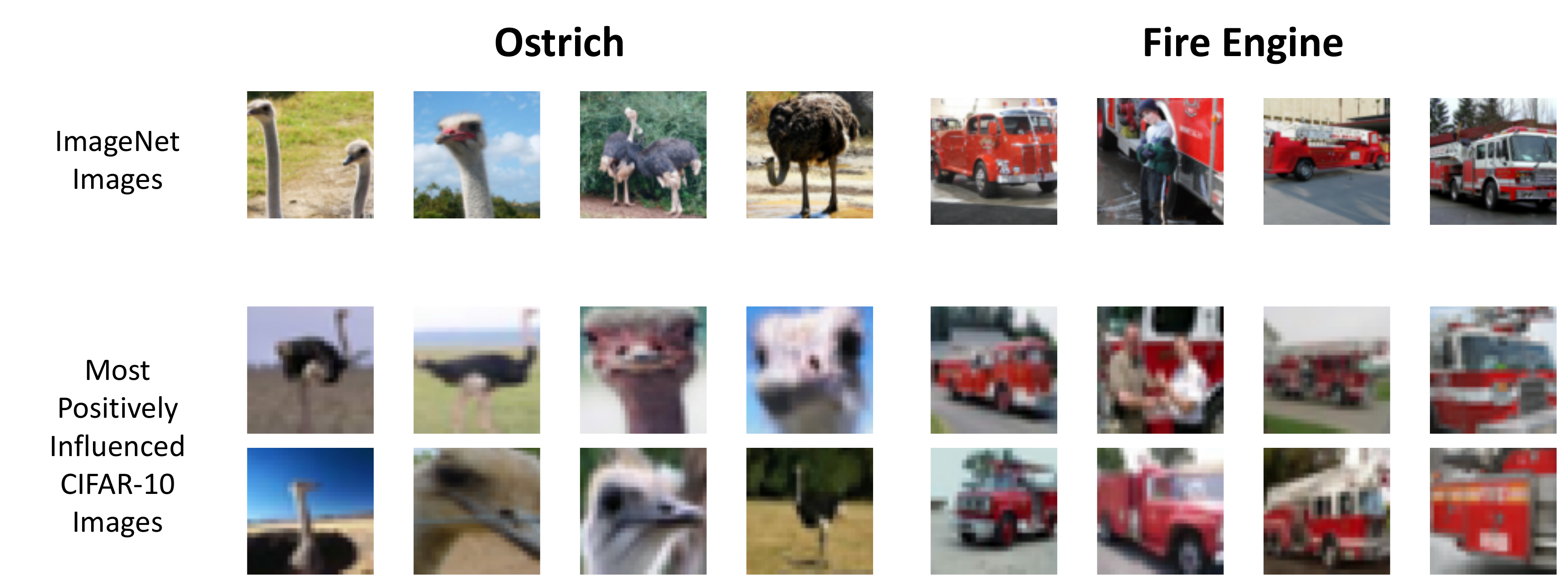}
    \caption{\textbf{Projecting source labels onto the target dataset.} The CIFAR-10 images that were most positively influenced by the ImageNet classes ``ostrich'' and ``fire engine.'' We find that these images look similar to the corresponding images in the source dataset.
    }
    \label{fig:extracting-subpopulations}
\end{figure}
In Section~\ref{sec:counterfactuals}, we developed a methodology for identifying source dataset classes that have the most impact on transfer learning performance. Now, we demonstrate how this methodology can be extended into a framework for probing and understanding transfer learning, including: (1) identifying granular target subpopulations that correspond to source classes, (2) debugging transfer learning failures, and (3) detecting data leakage between the source and target datasets. We focus our demonstration of these capabilities on a commonly-used transfer learning setting: ImageNet to CIFAR-10 (experimental details are in Appendix~\ref{app:setup}).

\subsection{Capability 1: Extracting target subpopulations by projecting source class labels}
Imagine that we would like to find all the ostriches in the CIFAR-10 dataset. This is not an easy task as CIFAR-10 only has ``bird'' as a label, and thus lacks sufficiently fine-grained annotations. Luckily, however, ImageNet \textit{does} contain an ostrich class! Our computed influences enable us to ``project" this ostrich class annotation (and, more broadly, the fine-grained label hierarchy of our source dataset) to find this subpopulation of interest in the target dataset.

Indeed, our examination from Section~\ref{sec:counterfactuals} suggests that the most positively influencing source classes are typically those that directly overlap with the target classes (see Figure~\ref{fig:influencers-cifar-overall}). In particular, for our example, ``ostrich'' is highly positively influential for the ``bird'' class (see Figure~\ref{fig:class-barplots}). To find ostriches in the CIFAR-10 dataset, we thus need to simply surface the CIFAR-10 images which were most positively influenced by the ``ostrich'' source class (see Figure~\ref{fig:extracting-subpopulations}).



\begin{figure}[!t]
    \centering
    \includegraphics[width=\linewidth]{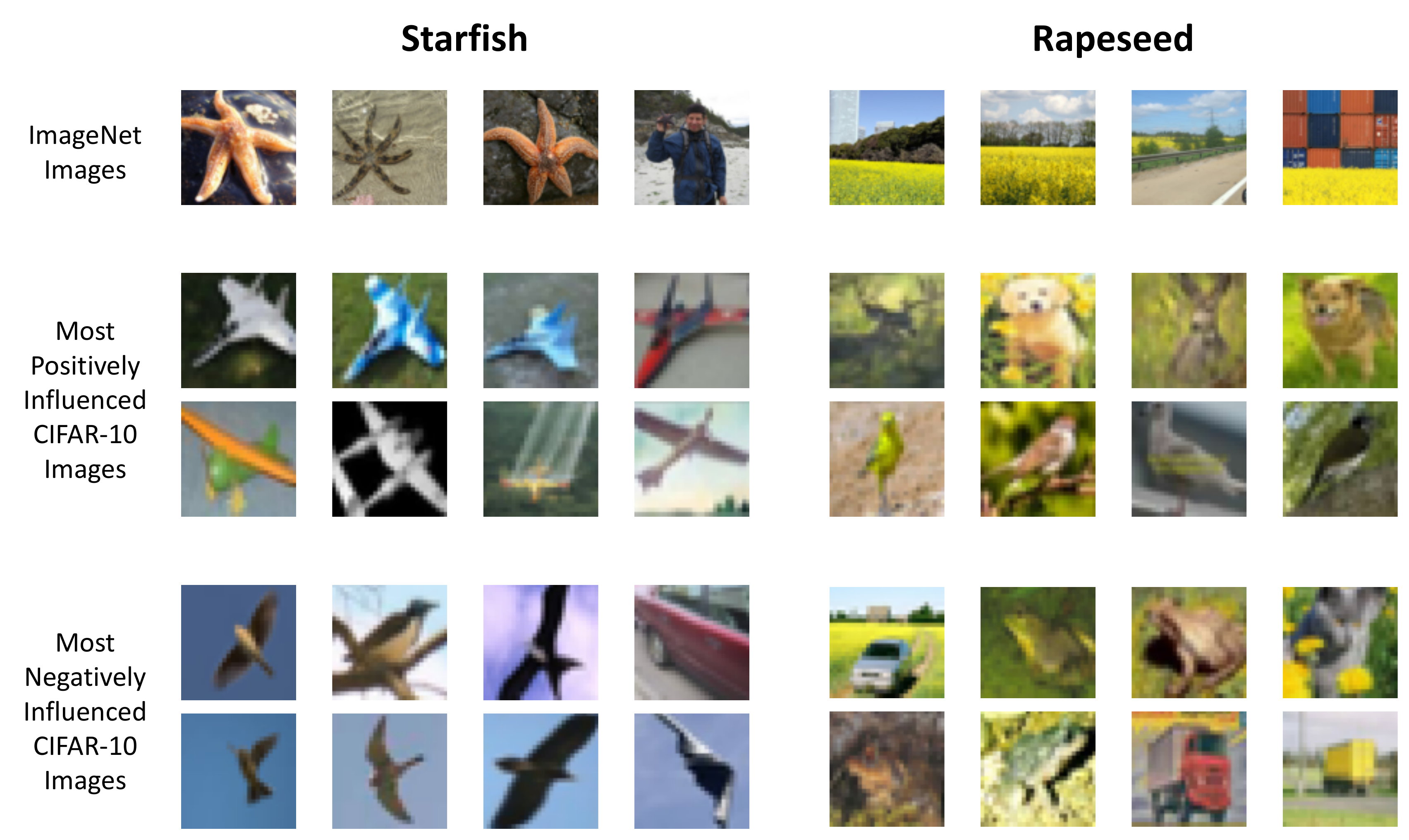}
    \caption{The CIFAR-10 images that were most positively (or negatively) influenced by the ImageNet classes ``starfish'' and ``rapeseed.'' CIFAR-10 images that are highly influenced by the ``starfish'' class have similar shapes, while those influenced by ``rapeseed'' class have yellow-green colors.
    }
    \label{fig:explaining-imagenet-classes-transfer}
\end{figure}

It turns out that this type of projection approach can be applied more broadly. Even when the source class is not a direct sub-type of a target class, the downstream model can still leverage salient features from this class --- such as shape or color --- to predict on the target dataset. For such classes, projecting source labels can extract target subpopulations which share such features.\saachi{aleks is this what we are going for} To illustrate this, in Figure~\ref{fig:explaining-imagenet-classes-transfer}, we display the CIFAR-10 images that are highly influenced by the classes ``starfish'' and ``rapeseed'' (both of which do not directly appear in the CIFAR-10 dataset). For these classes, the most influenced CIFAR-10 images share the same shape (``starfish'') or color (``rapeseed'') as their ImageNet counterparts. More examples of such projections can be found in Appendix~\ref{app:omitted}.

\subsection{Capability 2: Debugging the failures of a transferred model}
\label{sec:debugging-mistakes}
\begin{figure}[!t]
    \centering
    \includegraphics[width=\linewidth]{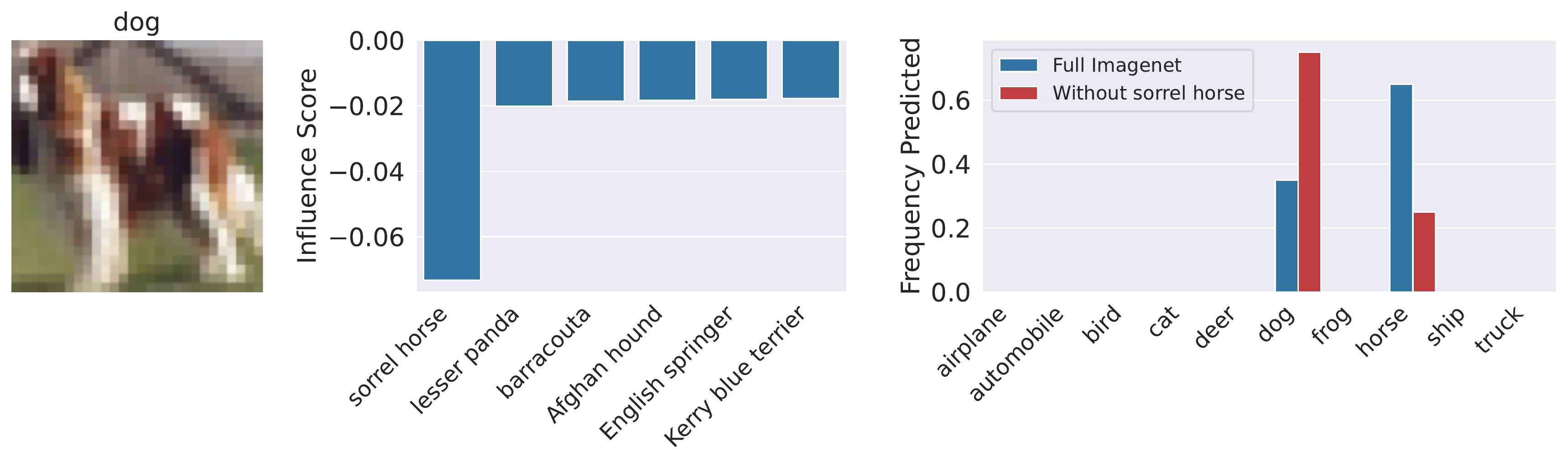}
    \includegraphics[width=\linewidth]{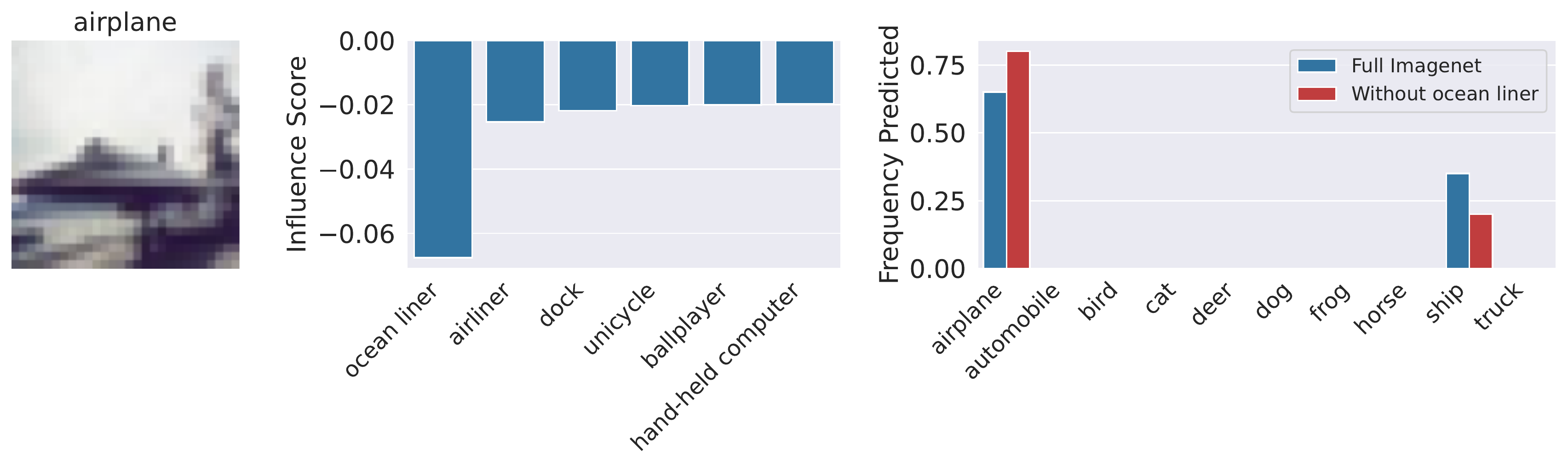}
    \caption{Pinpointing highly negatively influential source classes can help explain model mistakes. \textbf{Left:} For two CIFAR-10 images, we plot the most negatively influential source classes. \textbf{Right:} Over 20 runs, the fraction of times that our downstream model predicts each label for the given CIFAR-10 image. When the most negatively influential class is removed, the model predicts the correct label more frequently. More examples can be found in Appendix~\ref{app:omitted}.}
    \label{fig:explaining-mistakes}
\end{figure}

Our framework enables us to also reason about the possible mistakes of the transferred model caused by source dataset classes. For example, consider the CIFAR-10 image of a dog in Figure~\ref{fig:explaining-mistakes}, which our transfer learning model often mispredicts as a horse. Using our framework, we can demonstrate that this image is strongly negatively influenced by the source class ``sorrel horse.''  Thus, our downstream model may be misusing a feature introduced by this class. Indeed, once we remove ``sorrel horse'' from the source dataset, our model predicts the correct label more frequently. (See Appendix~\ref{app:omitted} for more examples, as well as a  quantitative analysis of this experiment.)

\subsection{Capability 3: Detecting data leakage and misleading source examples}
\label{sec:example_wise}
Thus far, we have focused on how the \textit{classes} in the source dataset influence the predictions of the transferred model on target examples. In this section, we extend our analysis to the \textit{individual} datapoints of the source dataset. We do so by adapting our approach to measure the influence of each individual source datapoint on each target datapoint. Further details on how these influences are computed can be found in Appendix~\ref{app:example-based}.\saachi{this was a comment, but we use datamodels for this, which we do not want to get into here.}

\begin{figure}[!t]
    \centering
        \includegraphics[width=\linewidth]{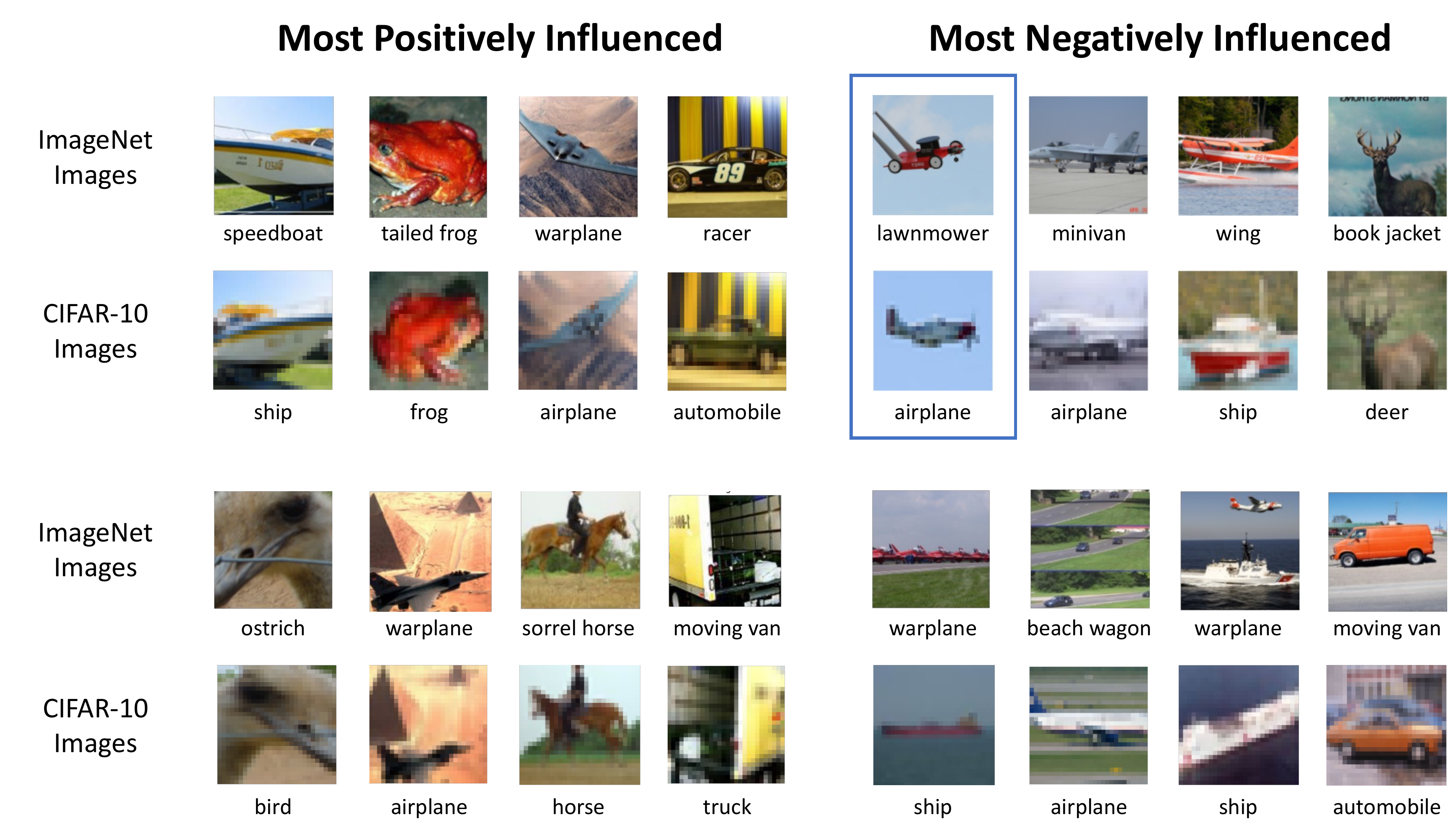}
    \caption{ImageNet training images with highest positive (\textbf{left}) or negative (\textbf{right}) example-wise (average) influences on CIFAR-10 test images.
        We find that ImageNet images that are highly positively influential often correspond to data leakage, while ImageNet images that are highly negatively influential are often either mislabeled, ambiguous, or otherwise misleading. For example, the presence of a flying lawnmower in the ImageNet dataset hurts the downstream performance on a similarly shaped airplane (boxed).}
    \label{fig:data-leakage}
\end{figure}

Figure~\ref{fig:data-leakage} displays the ImageNet training examples that have highly positive or negative influences on CIFAR-10 test examples. We find that the source images that are highly positively influential are often instances of \textit{data leakage} between the source training set and the target test set. On the other hand, the ImageNet images that are highly negatively influential are typically mislabeled, misleading, or otherwise surprising. For example, the presence of the ImageNet image of a flying lawnmower hurts the performance on a CIFAR-10 image of a regular (but similarly shaped) airplane (see Figure~\ref{fig:data-leakage}).

%% file: sections/related.tex
\paragraph{Transfer learning.}
Transfer learning is a technique commonly used in domains ranging from medical imaging \citep{mormont2018comparison, ke2021chextransfer}, language modeling \citep{conneau2018senteval}, to object detection \citep{ren2015faster, dai2016r, girshick2014rich, chen2017deeplab}. Therefore, there has been considerable interest in understanding the drivers of transfer learning's success. For example, by performing transfer learning on block-shuffled images, \citet{neyshabur2020being} demonstrate that at least some of the benefits of transfer learning come from low-level image statistics of source data. There is also an important line of work studying transfer learning by investigating the relationship between different properties of the source model and performance on the target task~\cite{ke2021chextransfer, kornblith2019better, salman2020adversarially, utrera2020adversarially}.

The works that are the most relevant to ours are those which studied how modifying the source dataset can affect the downstream performance. For example, \citet{kolesnikov2019big} showed that pre-training with an enormous source dataset (approximately 300 million) of noisily labeled images can outperform pre-training with ImageNet. \citet{azizpour2015factors, huh2016makes} investigated the importance of the number of classes and the number of images per class in transfer learning. Finally, \citet{ngiam2018domain} demonstrated that  more pre-training data does not always help, and transfer learning can be sensitive to the choice of pre-training data. They also presented a framework for reweighting the source datapoints in order to boost transfer learning performance.

\paragraph{Influence functions and datamodels.}

Influence functions are well-studied statistical tools that have been recently applied in machine learning settings~\citep{hampel2011robust,cook1982residuals,koh2017understanding}. For a given model, influence functions analyze the effect of a training input on the model's predictions by estimating the expected change in performance when this training input is added or removed.  A recent line of work estimates this quantity more efficiently by training on different subsets of the training set~\citep{feldman2020what}. In a similar vein, \citet{ghorbani2019data} proposed running a Monte Carlo search to estimate the effect of every training input via Shapley values. More recently, \citet{ilyas2022datamodels} proposed datamodeling framework as an alternative way to estimate the effect of a training input on the models' prediction. Datamodels are represented using parametric functions (typically, linear functions) that aim to map a subset of the training set to the model's output.

%% file: sections/conclusion.tex



In this work, we presented a new framework for examining the impact of the source dataset in transfer learning. Specifically, our approach estimates the influence of a source class (or datapoint) that captures how including that class (or datapoint) in the source dataset impacts the downstream model's predictions.  Leveraging these estimates, we demonstrate that we can improve the transfer learning performance on a range of downstream tasks by identifying and removing detrimental datapoints from the source dataset. Furthermore, our framework enables us to identify granular subpopulations in the target dataset by projecting fine-grained labels from the source dataset, better understand model failures on the downstream task and detect potential data-leakages from the source to the downstream dataset. We believe our framework provides a new perspective on transfer learning: one that enables us to perform a fine-grained analysis of the impact of the source dataset.




%% file: sections/ack.tex
Work supported in part by the NSF grants CCF-1553428 and CNS-1815221, and Open Philanthropy. This material is based upon work supported by the Defense Advanced Research Projects Agency (DARPA) under Contract No. HR001120C0015.

Research was sponsored by the United States Air Force Research Laboratory and the United States Air Force Artificial Intelligence Accelerator and was accomplished under Cooperative Agreement Number FA8750-19-2-1000. The views and conclusions contained in this document are those of the authors and should not be interpreted as representing the official policies, either expressed or implied, of the United States Air Force or the U.S. Government. The U.S. Government is authorized to reproduce and distribute reprints for Government purposes notwithstanding any copyright notation herein.

We thank the MIT Supercloud cluster~\citep{reuther2018interactive} for providing computational resources that supported part of this work.

%% file: sections/app_setup.tex
\subsection{ImageNet Models}
\label{app:models}
In this paper, we train a large number of models on various subsets of ImageNet in order to estimate the influence of each class of ImageNet on the model's transfer performance for multiple downstream tasks. We focus on the ResNet-18 architecture from PyTorch's official implementation found here \url{https://pytorch.org/vision/stable/models.html}\footnote{Our framework is agnostic to the choice of the model's architecture.}.

\paragraph{Training details.}
We fix the training procedure for all of our models. Specifically, we train our models from scratch using SGD to minimize the standard cross-entropy multi-class classification loss. We use a batch size of 1024, momentum of $0.9$, and weight decay of $5\times 10^{-4}$.
The models are trained for 16 epochs using a Cyclic learning rate schedule with an initial learning rate of $0.5$ and learning rate peak epoch of $2$. We use standard data-augmentation: \textit{RandomResizedCrop} and
\textit{RandomHorizontalFlip} during training, and \textit{RandomResizedCrop} during testing. Our implementation and configuration files are available in our code \url{https://github.com/MadryLab/data-transfer}.

\subsection{ImageNet transfer to classification datasets}
\label{app:transfer-to-small-datasets}

\paragraph{Datasets.}
\label{app:classification-datasets}
\begin{table}[!h]
    \begin{center}
        \begin{small}
        \begin{tabular}{@{}lccc@{}}
            \toprule
            \textbf{Dataset}           & \textbf{Classes} & \textbf{Train Size} & \textbf{Test Size} \\ \midrule
            Birdsnap \citep{berg2014birdsnap}                  & 500     & 32,677 & 8,171        \\
            Caltech-101 \citep{fei2004learning} & 101 & 3,030 & 5,647 \\
            Caltech-256 \citep{griffin2007caltech} & 257 & 15,420 & 15,187 \\
            CIFAR-10 \citep{krizhevsky2009learning}                  & 10      & 50,000 & 10,000     \\
            CIFAR-100 \citep{krizhevsky2009learning}                 & 100     & 50,000 & 10,000       \\
            FGVC Aircraft \citep{maji2013fine}             & 100     & 6,667 & 3,333         \\
            Food-101 \citep{bossard2014food}                  & 101     & 75,750 & 25,250       \\
            Oxford 102 Flowers \citep{nilsback2008automated}        & 102     & 2,040 & 6,149           \\
            Oxford-IIIT Pets \cite{parkhi2012cats} & 37 & 3,680 & 3,669 \\
            SUN397 \citep{xiao2010sun}                    & 397     & 19,850 & 19,850       \\
            Stanford Cars \citep{krause2013collecting}              & 196     & 8,144 & 8,041        \\
            \bottomrule
            \end{tabular}
        \end{small}
    \end{center}
    \caption{Image classification datasets used in this paper.}
    \label{table:datasets}
\end{table}

We consider the transfer image classification tasks that are used in \citep{salman2020adversarially,kornblith2019better},
which vary in size and number of classes. See Table~\ref{table:datasets} for the details of these datasets. 
We consider two transfer learning settings for each dataset: \textit{fixed-feature} and \textit{full-network} transfer learning.

\paragraph{Fixed-feature transfer.}
\label{app:logistic-regression-params}
For this setting, we \textit{freeze} the layers of the ImageNet source model\footnote{For all of our experiments, we do not freeze the batch norm statistics. We only freeze the weights of the model, similar to~\citet{salman2020adversarially}.}, except for the last layer, which we replace with a random initialized linear layer whose output matches the number of classes in the transfer dataset.
We now train only this new layer for using SGD, with a batch size of 1024 using cyclic learning rate. For more details and hyperparameter for each dataset, please see config files in our code \url{https://github.com/MadryLab/data-transfer}.

\paragraph{Full-network transfer.}
\label{app:finetuning-params}
For this setting, we \textit{do not freeze} any of the layers of the ImageNet source model, and all the model weights are updated. We follow the exact same hyperparameters as the fixed-feature setting.

\subsection{Compute and training time.}
We leveraged the FFCV data-loading library for fast training of the ImageNet models~\cite{leclerc2022ffcv}\footnote{Using FFCV, we can train a model on the ImageNet dataset in around 1 hour, and reach $\sim$63\% accuracy}. Our experiments were run on two GPU clusters: an A100 and a V100 cluster.


\paragraph{Training ImageNet models and influence calculation.} We trained 7,540 ImageNet models on random subsets of ImageNet, each containing half of ImageNet classes. On a single V100, training a single model takes around 30 minutes. After training these ImageNet models, we compute the influence of each class as outlined in Algorithm~\ref{alg:main}. Computing the influences is fast, and takes few seconds on a single V100 GPU.


\subsection{Handpicked baseline details}
In our counterfactual experiments in Section~\ref{sec:counterfactuals}, we automatically selected, via our framework, the most influential subsets of ImageNet classes for various downstream tasks. We then removed the classes that are detrimental to the transfer performance, and measured the transfer accuracy improvement after removing these classes. The results are summarized in Table~\ref{fig:summary_fixed_cf}.

\paragraph{ What happens if we hand-pick the source dataset classes that are relevant to the target dataset?}


Indeed, \citet{ngiam2018domain} found that hand-picking the source dataset classes can sometimes boost transfer performance. We validate this approach for our setting using the WordNet hierarchy~\citep{miller1995wordnet}. Specifically, for each class from the target task, we look up all the ImageNet classes that are either children or parents of this target class. The set of all such ImageNet classes are used as the handpicked most influential classes. Following this manual selection, we train an ImageNet model on these classes, then apply transfer learning to get the baseline performance that we report in Table~\ref{fig:summary_fixed_cf}.

\subsection{Error and Convergence Analysis}
\label{app:convergence}
We compute our class influence values using 7,540 source models, each of which were trained using 500 randomly chosen ImageNet classes. How many models do we actually need to compute our transfer influences? In order to analyze the convergence of the transfer influences, we track the standard deviation of the influence values after bootstrap resampling.

We consider the ImageNet $\rightarrow$ CIFAR-10 transfer setting with fixed-feature fine-tuning. Given $N$ models, we randomly sample, with replacement, $N$ models to recompute our transfer influences. Specifically, we evaluate the overall transfer influences (i.e., the influence value of each ImageNet class averaged over all target examples). We perform this resampling 500 times, and measure the standard deviation of the computed overall transfer influence value for each class over these 500 resamples.

Below, we plot this standard deviation (averaged over the 1000 classes) for various number of models $N$. We find that the standard deviation goes down as more models are used, indicating that our estimate of the influence values has less variance. This metric roughly plateaus by the time we are using 7000 models.
\begin{figure}[htbp!]
    \centering
    \includegraphics[width=0.5\linewidth]{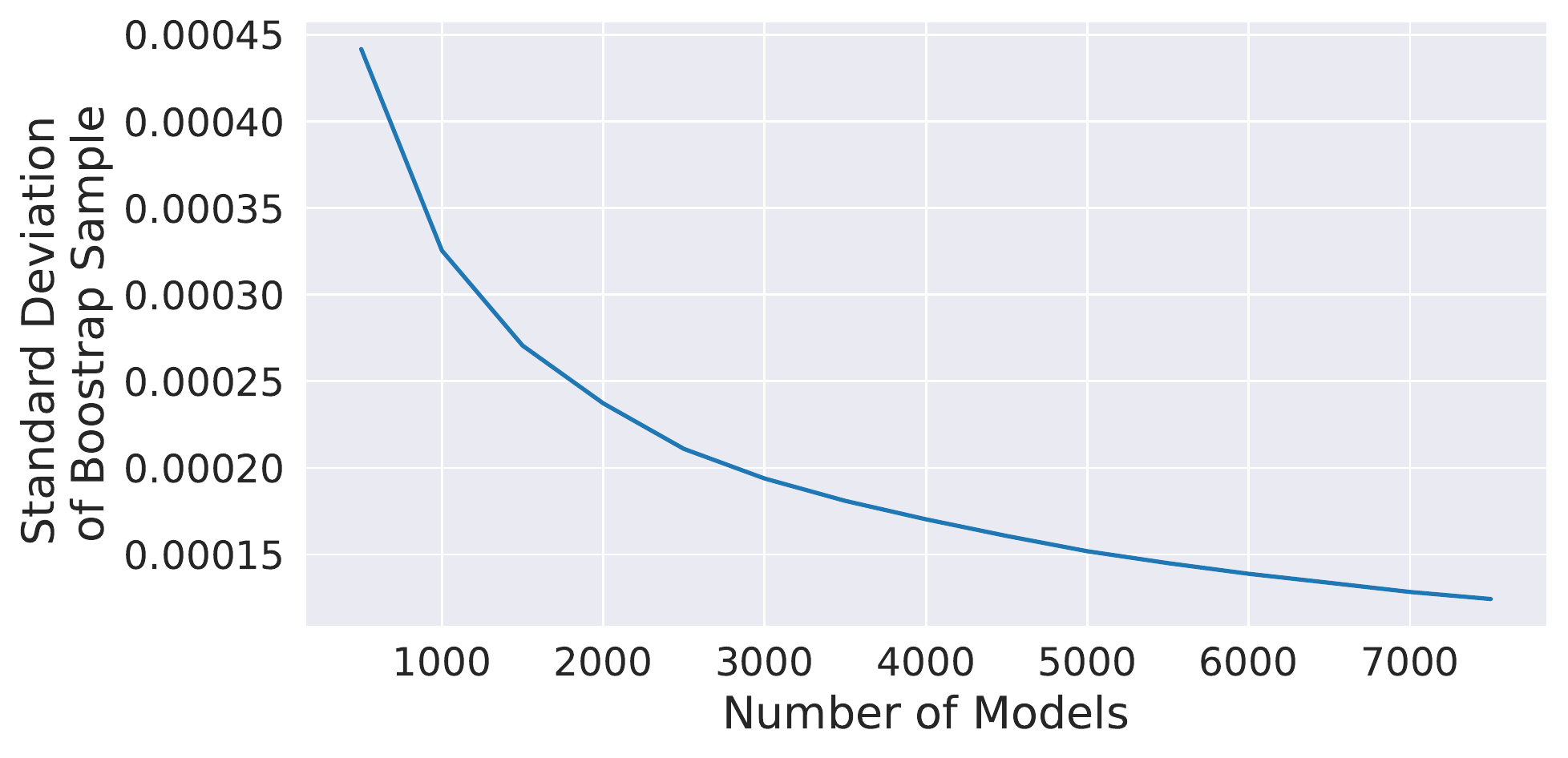}
    \caption{Standard deviation of the overall influence values (averaged over classes) after bootstrap resampling for various numbers of models. }
\end{figure}

%% file: sections/app_variants_influences.tex
\subsection{Variations of targets for computing transfer influences}
\label{app:vary_model_outputs}
In our paper, we used the softmax output of the groundtruth class as the target for our influence calculation. What happens if we use a different target? We compare using the following types of targets.
\begin{itemize}
    \item Softmax Logits: the softmax output of the groundtruth class
    \item Is Correct: the binary value of whether the image was predicted correctly
    \item Raw Margins: the difference in raw output between the correct class and the most confidently predicted incorrect class
    \item Softmax Margins: the same as raw margins, but use the output after softmax
\end{itemize}

\begin{figure}[h!]
    \centering
    \includegraphics[width=0.9\textwidth]{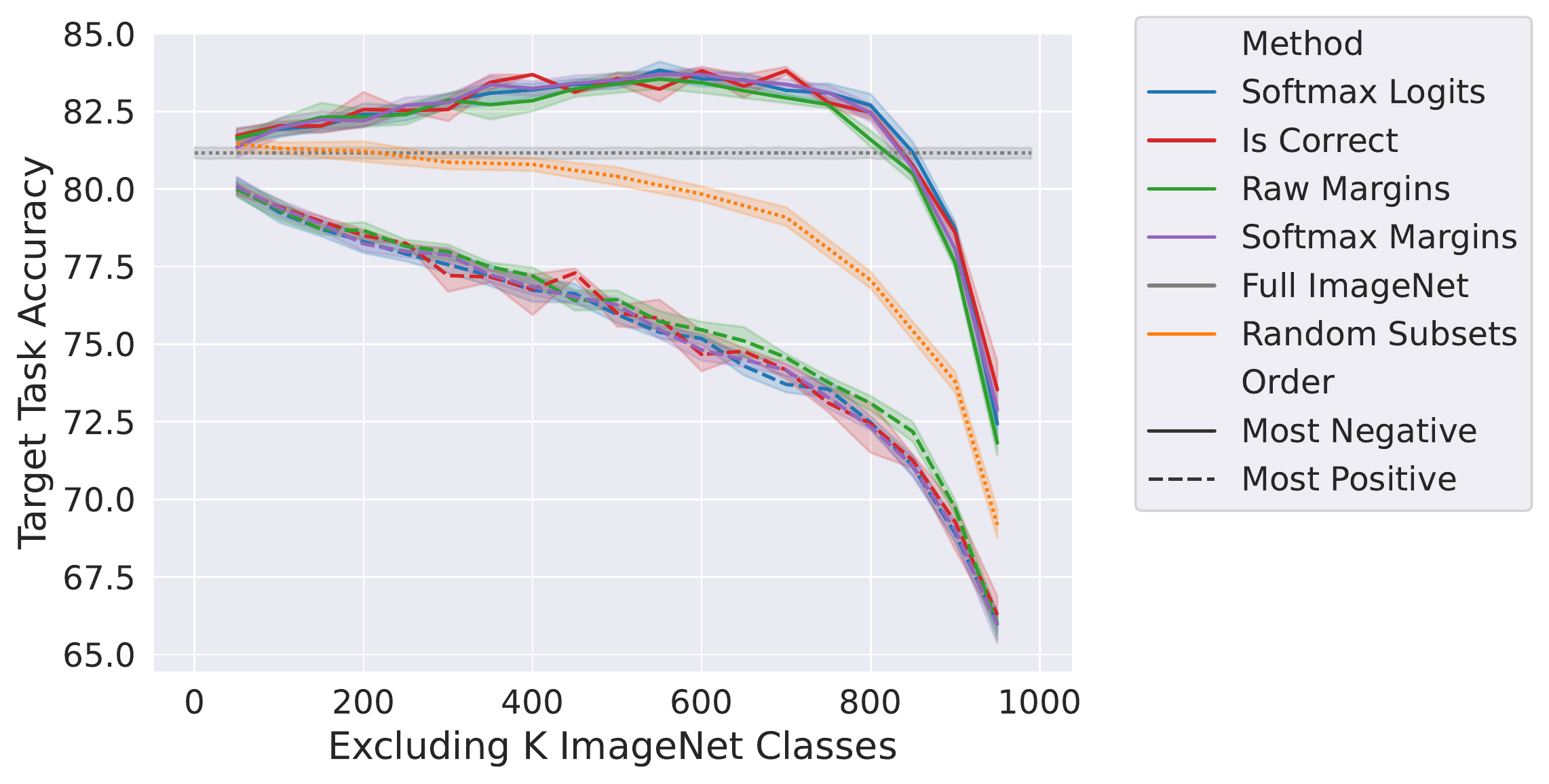}
    \caption{Target task accuracies after removing the most positively or negatively influential ImageNet classes from the source dataset with various influence targets.}
    \label{app_fig:compare_targets}
\end{figure}

In Figure~\ref{app_fig:compare_targets}, we replicate the counterfactual experiment from the main paper in Figure~\ref{fig:summary_fixed_cf} using these different targets. Specifically (over 2 runs), we rank the overall influence of the ImageNet classes on CIFAR-10 for fixed-feature transfer. We then remove the classes in order most most or least influence.

We find that our method is not particularly sensitive to the individual target used. We found that using the softmax logits provided the highest benefit when removing negative influencers, and thus used that target for the reset of our experiments. 

\paragraph{Datamodels vs. Influences.}
Datamodels~\cite{ilyas2022datamodels} is another method that, similar to influences, seeks to compute the importance of a training point on a test set prediction. Specifically, instead of computing the difference in the expected accuracy of the model when a training point is removed, the method fits a linear model that, given a binary vector that denotes the composition of the training dataset, predicts the raw margin (i.e., the difference in raw output between the correct class and the most confidently predicted incorrect class). The importance of each training point is then the coefficient of the linear model for that particular example. 

We adapt this method to our framework by training a linear model with ridge regression to predict the softmax output of the transfer model on the target images given a binary vector that denotes which source classes were included in the source dataset. However, we find that datamodels were more effective for computing example-based values (see Appendix~\ref{app:example-based}).

In Figure~\ref{app_fig:class_dm}, we compare using influence values (as described in the main paper) to using these adapted datamodels.  Specifically (over 5 runs), we rank the overall importance of the ImageNet classes on CIFAR-10 for fixed-feature transfer using influences or datamodels. We then remove the classes in order most most or least influence. We find that our framework is not sensitive to the choice of datamodels or influences. However, influences performed marginally better in this counterfactual experiment, so we used influences for all other experiments in this paper.

\begin{figure}[h!]
    \centering
    \includegraphics[width=0.9\textwidth]{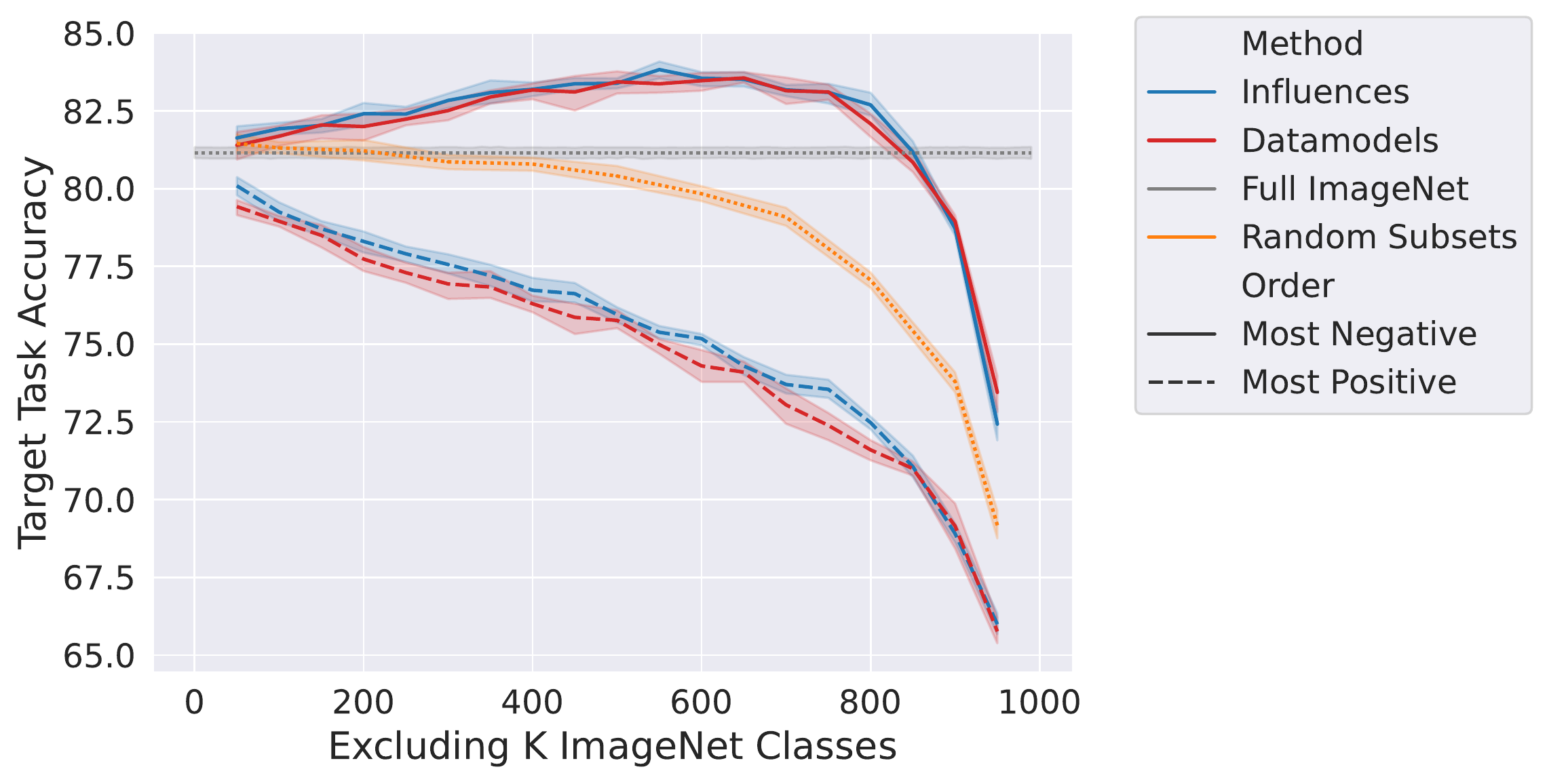}
    \caption{Target task accuracies after removing the most positively or negatively influential ImageNet classes from the source dataset using datamodels or influences.}
    \label{app_fig:class_dm}
\end{figure}

%% file: sections/app_counterfactual.tex
In this section, we display the full results for the counterfactual experiment in the main paper 
(Figure ~\ref{fig:summary_fixed_cf}). Specifically, for each target task, we display the target task accuracies after removing the most positive (top) and negative (bottom) influencers from the dataset over 10 runs. We find that our results hold across datasets. 
\begin{figure}[h!]
    \centering
    \includegraphics[width=0.9\linewidth]{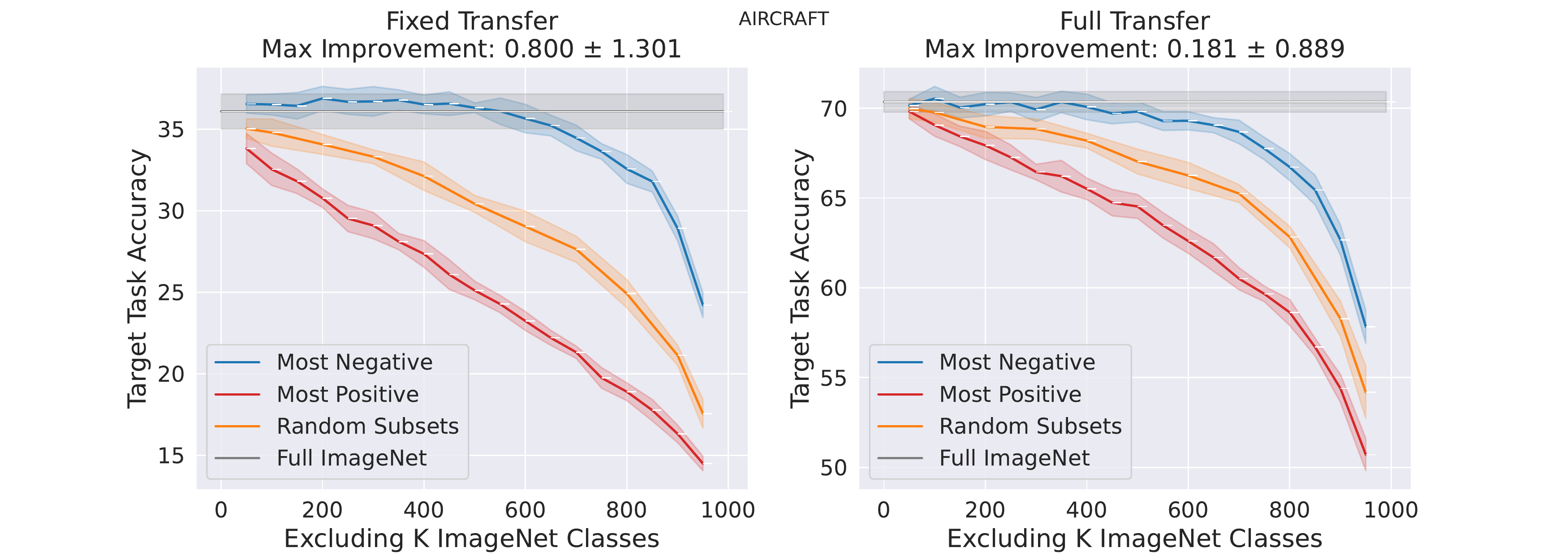}
    \caption{AIRCRAFT}
\end{figure} \begin{figure}[h!]
    \centering
    \includegraphics[width=0.9\linewidth]{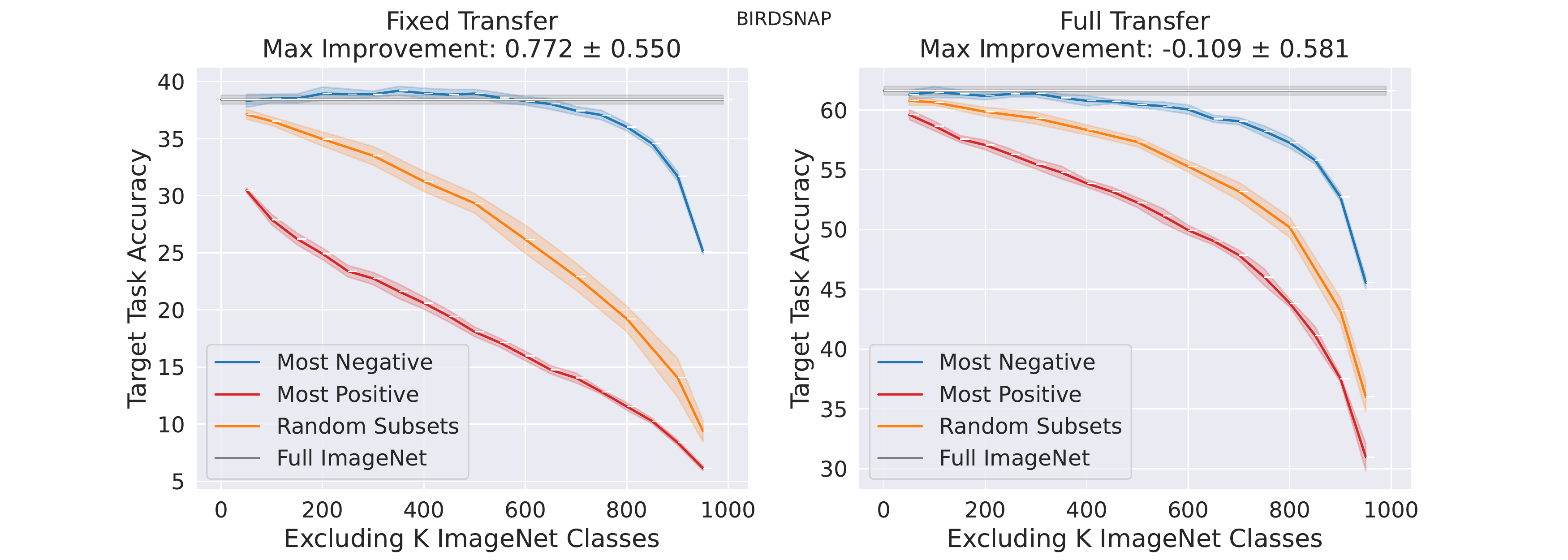}
    \caption{BIRDSNAP}
\end{figure} \begin{figure}[h!]
    \centering
    \includegraphics[width=0.9\linewidth]{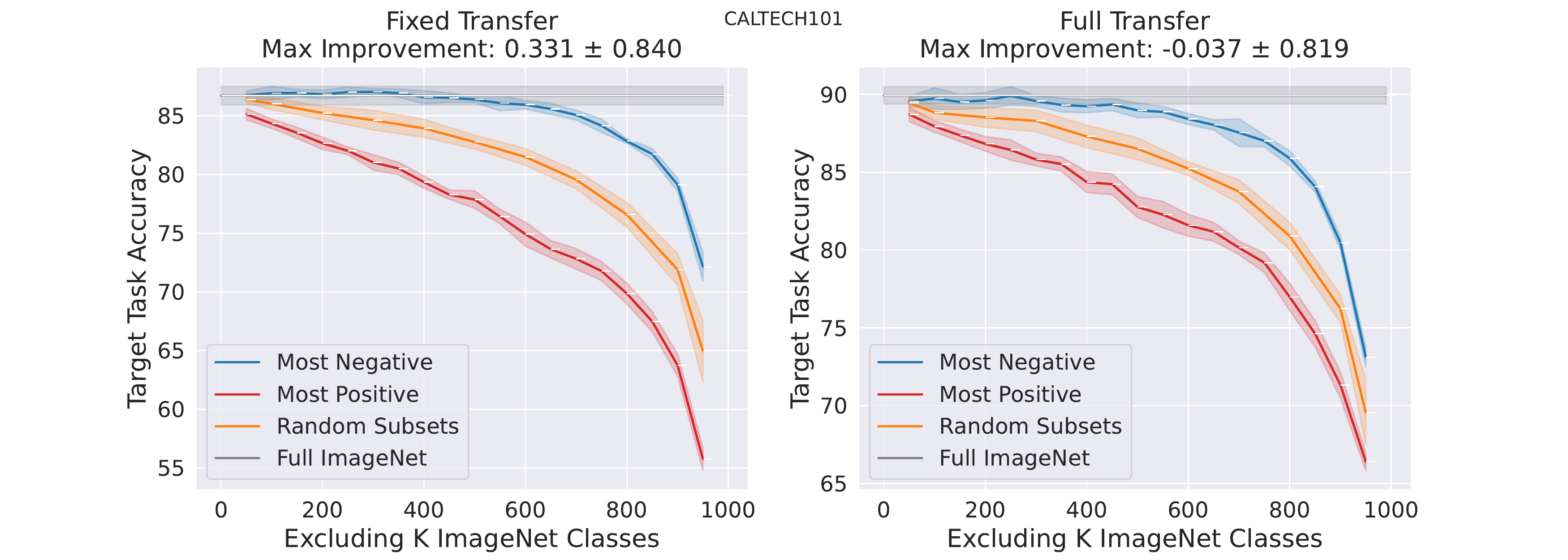}
    \caption{CALTECH101}
\end{figure} \begin{figure}[t!]
    \centering
    \includegraphics[width=0.9\linewidth]{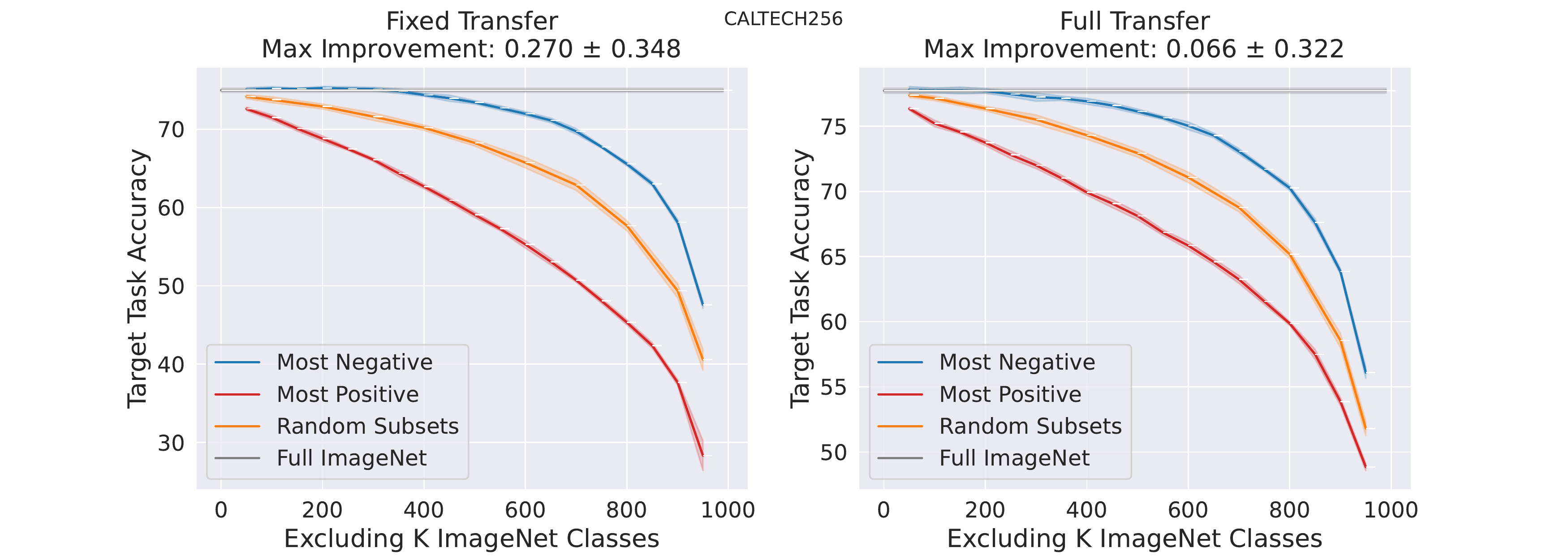}
    \caption{CALTECH256}
\end{figure} \begin{figure}[t!]
    \centering
    \includegraphics[width=0.9\linewidth]{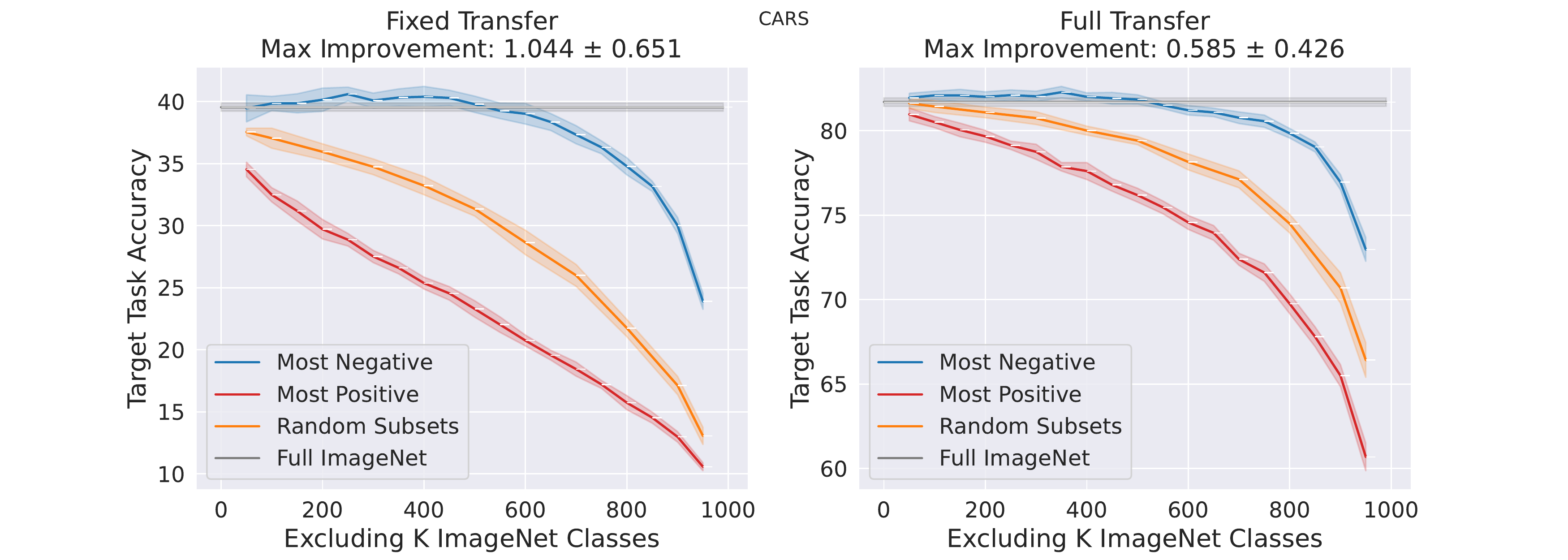}
    \caption{CARS}
\end{figure} \begin{figure}[t!]
    \centering
    \includegraphics[width=0.9\linewidth]{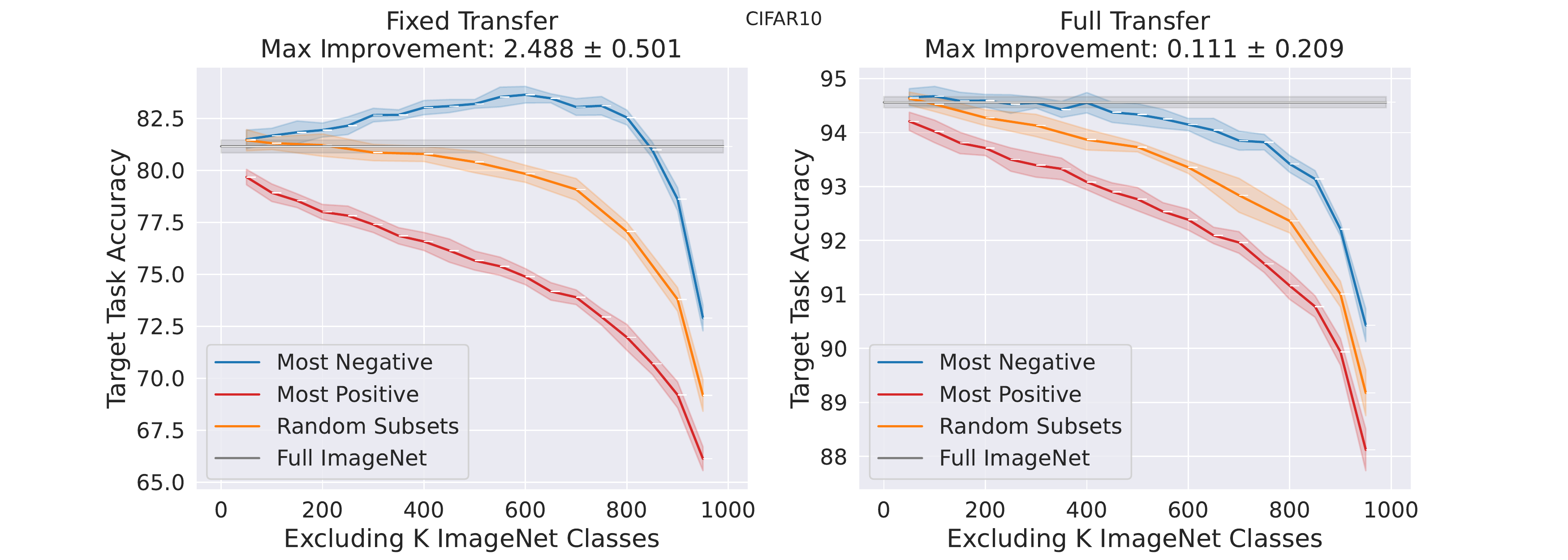}
    \caption{CIFAR10}
\end{figure} \begin{figure}[t!]
    \centering
    \includegraphics[width=0.9\linewidth]{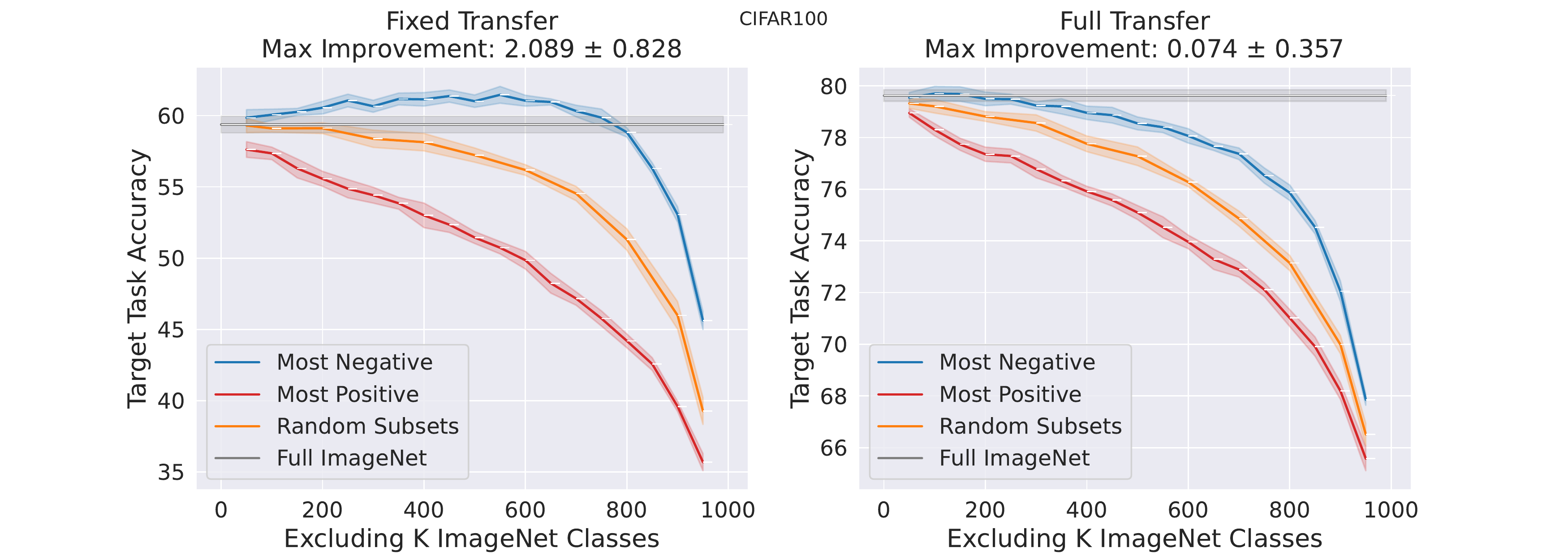}
    \caption{CIFAR100}
\end{figure} \begin{figure}[t!]
    \centering
    \includegraphics[width=0.9\linewidth]{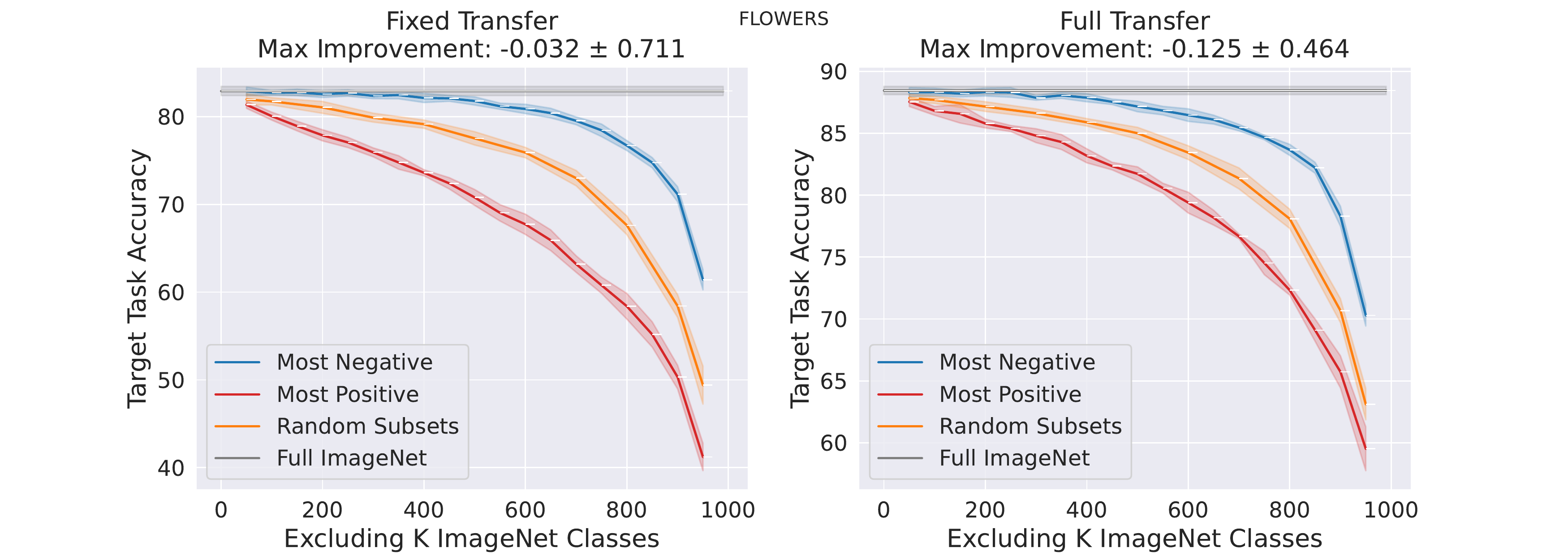}
    \caption{FLOWERS}
\end{figure} \begin{figure}[t!]
    \centering
    \includegraphics[width=0.9\linewidth]{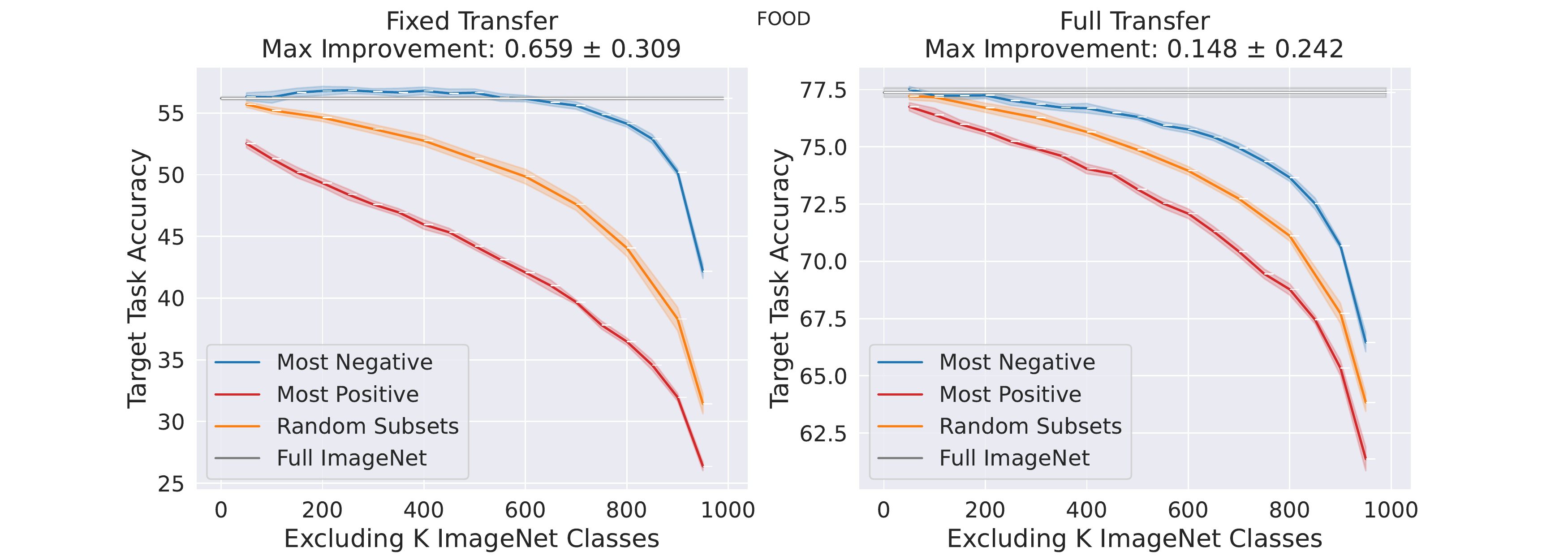}
    \caption{FOOD}
\end{figure} \begin{figure}[t!]
    \centering
    \includegraphics[width=0.9\linewidth]{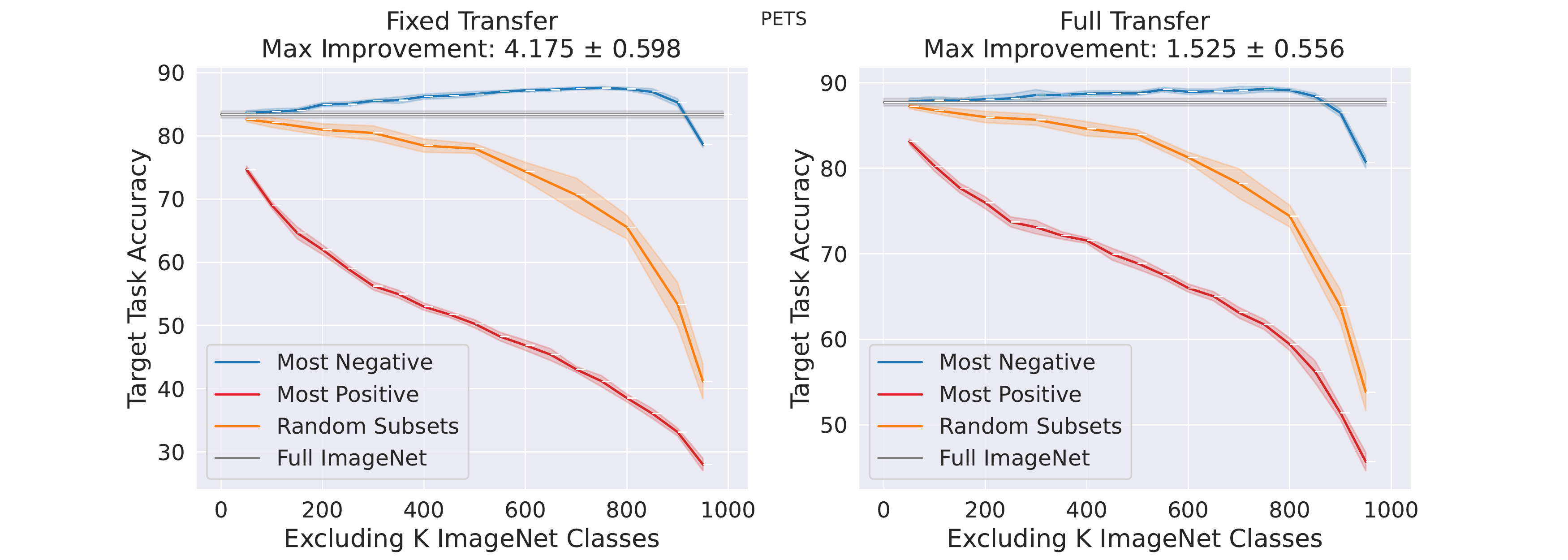}
    \caption{PETS}
\end{figure} \begin{figure}[t!]
    \centering
    \includegraphics[width=0.9\linewidth]{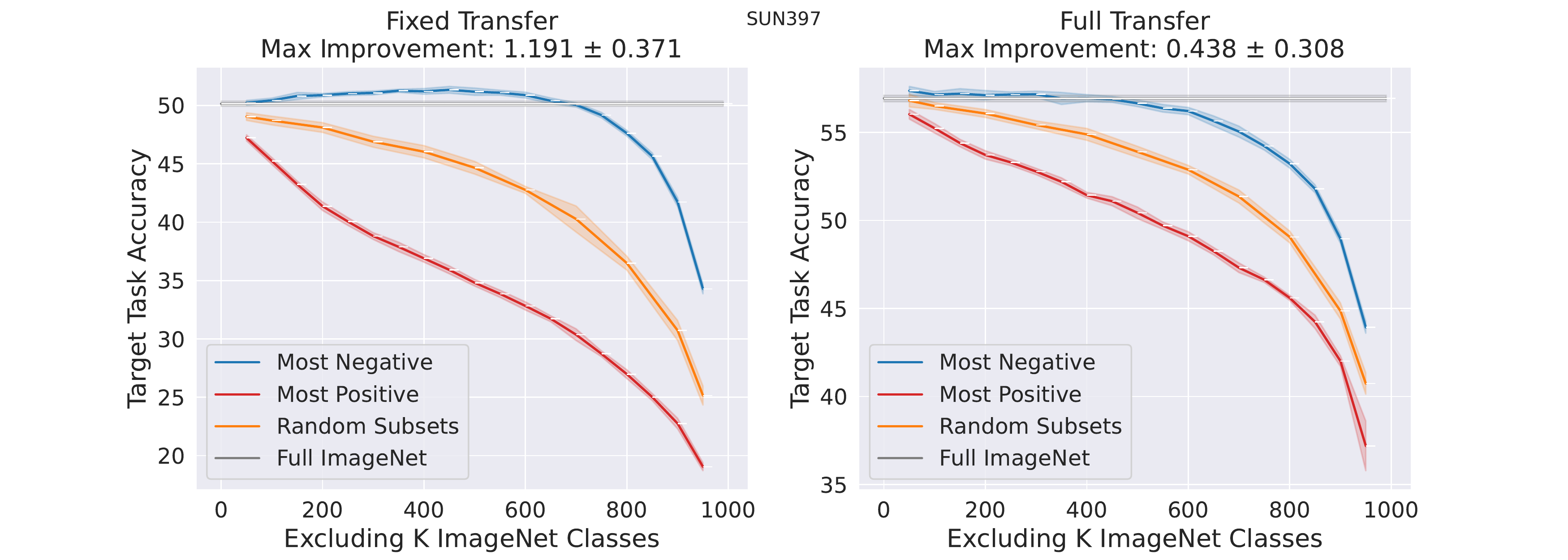}
    \caption{SUN397}
\end{figure}

%% file: sections/app_example_based.tex
We have presented in the main paper how to compute the influences of every class in the source dataset on the predictions of the model on the target dataset. In that setup, we demonstrated multiple capabilities of our framework, such as improving overall transfer performance, detecting particular subpopulations in the target dataset, etc. Given the wide range of capabilities class-based influences provide, one natural question that arises: Can we compute the influence of every source datapoint on the predictions of the model on the target dataset? Furthermore, what do these influences tell us about the transfer learning process?

Mathematically, the computation of example-based influences (i.e., the influence of every source datapoint) is very similar to the computation of class-based influences. Specifically, to compute example-based influences, we start by training a large number of models on different subsets of the source datapoints (as opposed to source classes for class-based influences). Next, we estimate the influence value of a source datapoint $s$ on a target example $t$ as the expected difference in the transfer model's performance on example $t$ when datapoint $s$ was either included or excluded from the source dataset:
\begin{align}
    \label{eq:influence_example}
    \text{Infl}[s \rightarrow t] = \mathbb{E}_{S} \left[ f(t; S) ~|~ s \in S \right] - \mathbb{E}_{S} \left[ f(t; S) ~|~ s \not \in S \right]
\end{align}
where $f(t; S)$ is the softmax output of a model trained on a subset $S$ of the source dataset. Similar to class-based influences, a positive (resp. negative) influence value indicates that including the source datapoint $s$ improves (resp. hurts) the model's performance on the target example $t$.

While example-based influences provide some insights about the transfer process, we found that---in this regime---datamodels~\citep{ilyas2022datamodels} provide cleaner results and better insights. Generally, influences and datamodels measure similar properties: the effect of the source datapoints on the target datapoints. For a particular target datapoint $t$, we measure the effect of every source datapoint $s$ with datamodels by solving a regression problem. Specifically, we train a large number of models on different subsets of the source dataset. For every model $f_i$, we record 1) a binary mask $\mathbbm{1}_{\mathcal{S}_i}$ that indicates which source datapoints were included in the subset $\mathcal{S}_i$ of the source dataset, and 2) the transfer performance $f_i(t; \mathcal{S}_i)$ of the model $f_i$ on the target datapoint $t$ after fine-tuning on the target dataset. Following the training and the fine-tuning stages, we fit a linear model $g_w$ that predicts the transfer performance $f(t; \mathcal{S})$ from a random subset $\mathcal{S}$ of the source dataset as follows: $f(t; \mathcal{S}) \approx g_w(\mathbbm{1}_{\mathcal{S}}) = w^\top \mathbbm{1}_{\mathcal{S}}$. Given this framework, $w = (w_1, w_2, \ldots, w_L)$ measures the effect of every source datapoint $s$ on the target datapoint $t$\footnote{To estimate the datamodels, we train 71,828 models on different subsets of the source dataset.}. We present the overall procedure in Algorithm~\ref{alg:example_based}.



\begin{algorithm}[!hbtp]
    \caption{Example-based datamodels estimation for transfer learning.}
    \label{alg:example_based}
    \begin{algorithmic}[1]
    \Require source dataset $\mathcal{S} = \cup_{l=1}^{L}~s_l$ (with $L$ datapoints), a target dataset $\mathcal{T} = (t_1, t_2,\cdots,t_n$), training algorithm $\mathcal{A}$, subset ratio $\alpha$, number of models $m$
    \State Sample $m$ random subsets $S_1,S_2,\cdots,S_m \subset \mathcal{S}$ of size $\alpha \cdot |\mathcal{S}|$:
    \For{$i \in  1$ to $m$}
        \State Train model $f_i$ by running algorithm $\mathcal{A}$ on $S_{i}$
    \EndFor
    \State Fine-tune $f_i$ on the training target dataset
    \For{$j \in  1$ to $n$}
        \State Collect datamodels training set $ \mathcal{D}_j = \{ \left(\mathbbm{1}_{\mathcal{S}_i}, f_i(t_j, \mathcal{S}_i) \right) \}_{i=1}^m$
        \State Compute $w_j$ by fitting LASSO on $\mathcal{D}_j$
    \EndFor
    \State
    \Return $w_j~~\forall~~j\in[n]$
    \end{algorithmic}
\end{algorithm}

%% file: sections/app_omitted.tex
\subsection{Per-class influencers}
\label{app:per-class-influencers}
We display for the ImageNet $\rightarrow$ CIFAR-10 the top (most positive) and bottom (most negative) influencing classes for each CIFAR-10 class. This is the equivalent to the plot in Figure~\ref{fig:class-barplots} in the main paper.

\begin{figure}[h!]
    \centering
    \includegraphics[width=0.95\linewidth]{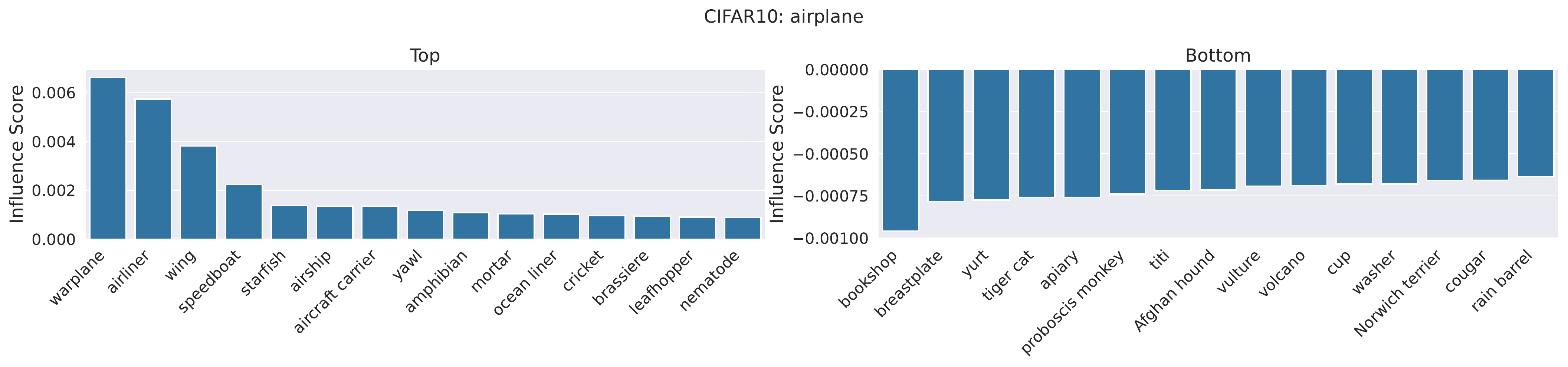}
    \includegraphics[width=0.95\linewidth]{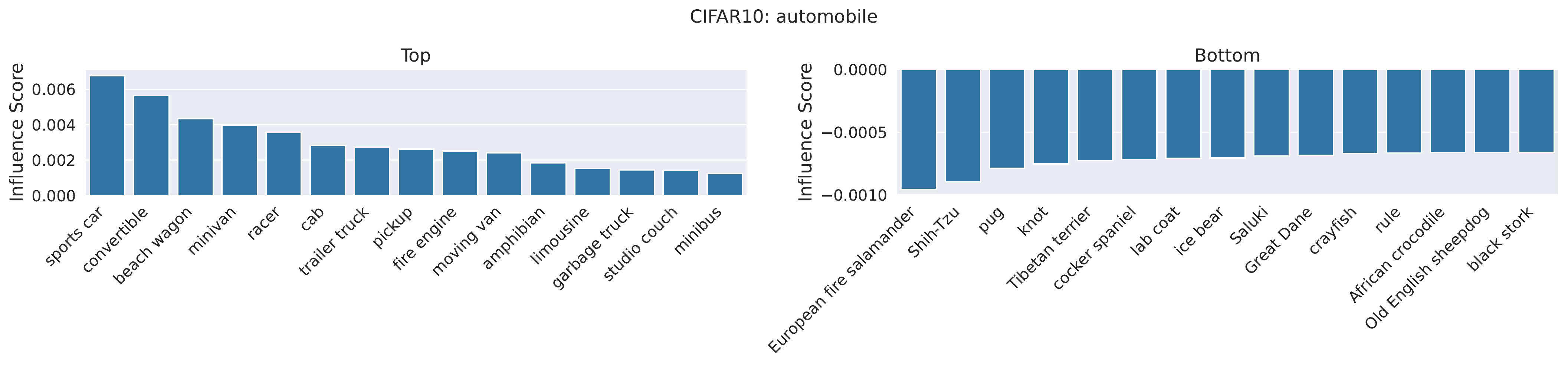}
    \includegraphics[width=0.95\linewidth]{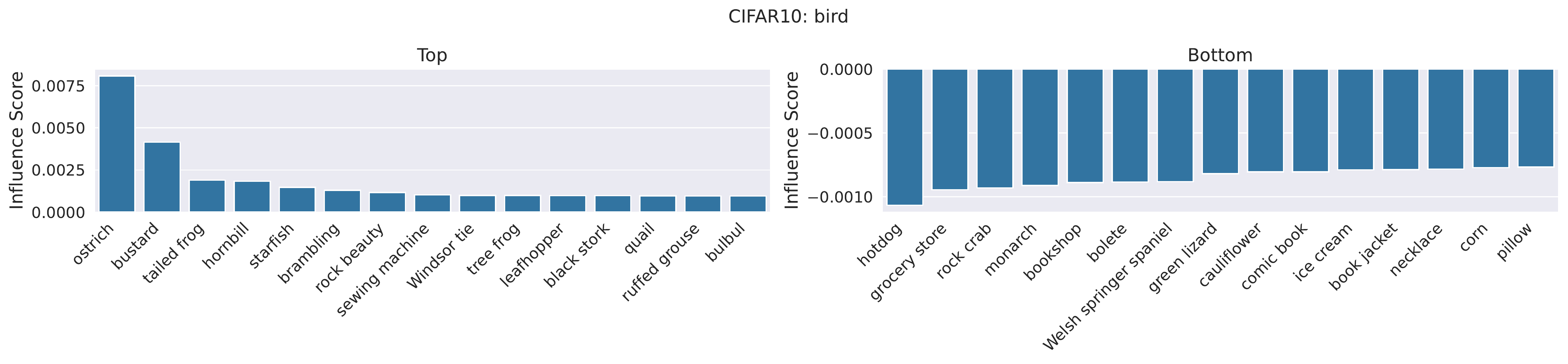}
    \includegraphics[width=0.95\linewidth]{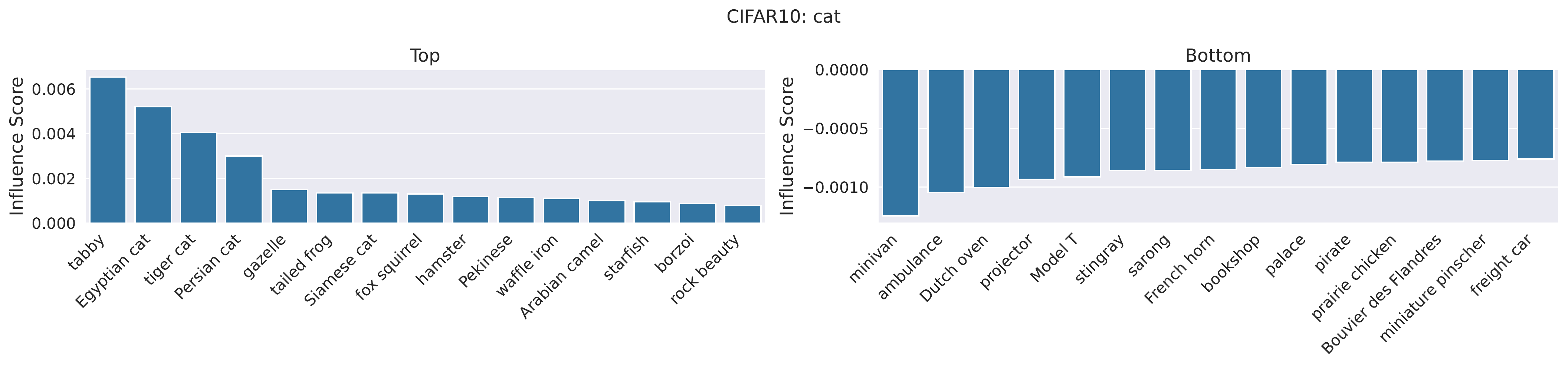}
    \includegraphics[width=0.95\linewidth]{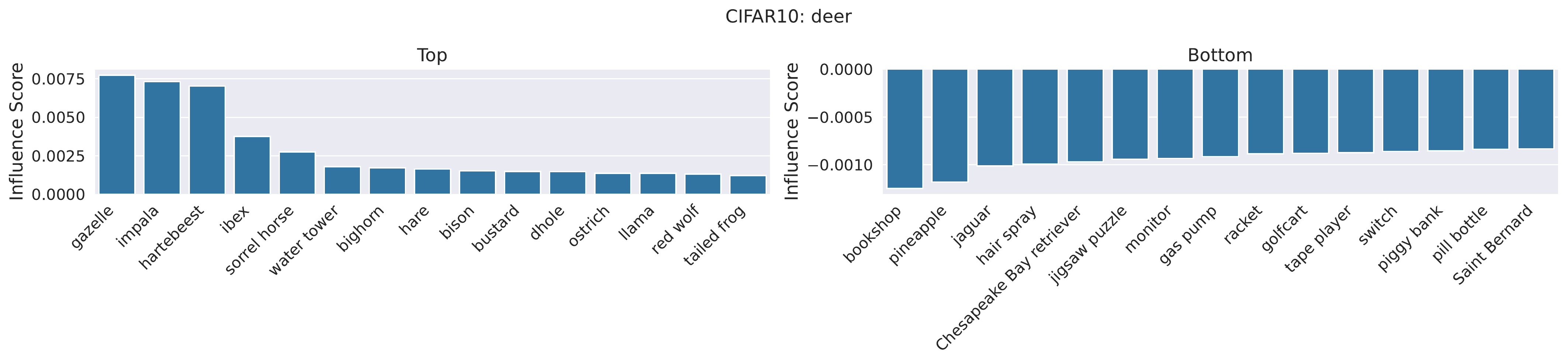}
    \caption{Top and bottom influencing ImageNet classes for all CIFAR-10 classes.}
\end{figure}

\begin{figure}[t!]
    \centering
    \includegraphics[width=\linewidth]{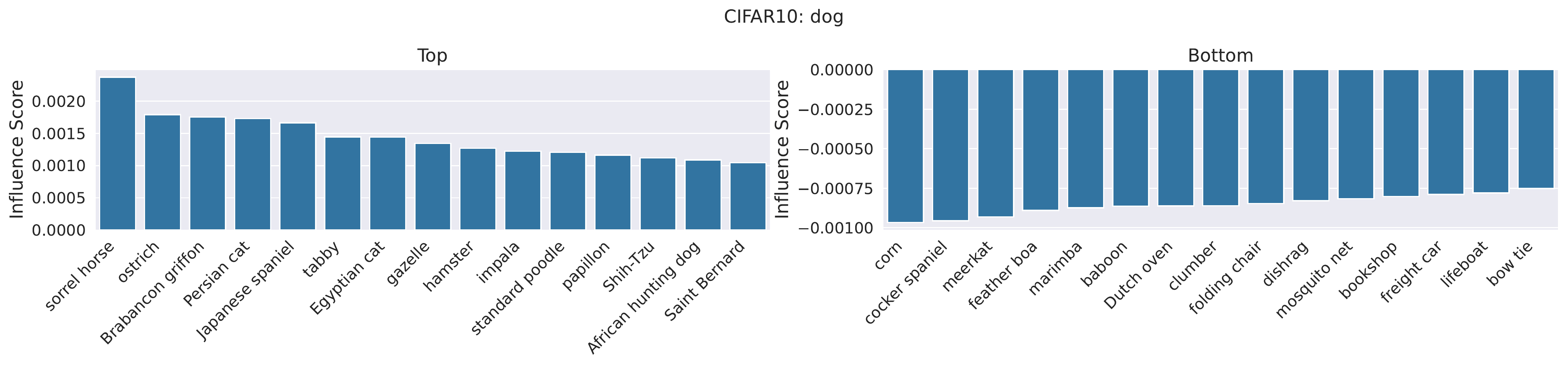}
    \includegraphics[width=\linewidth]{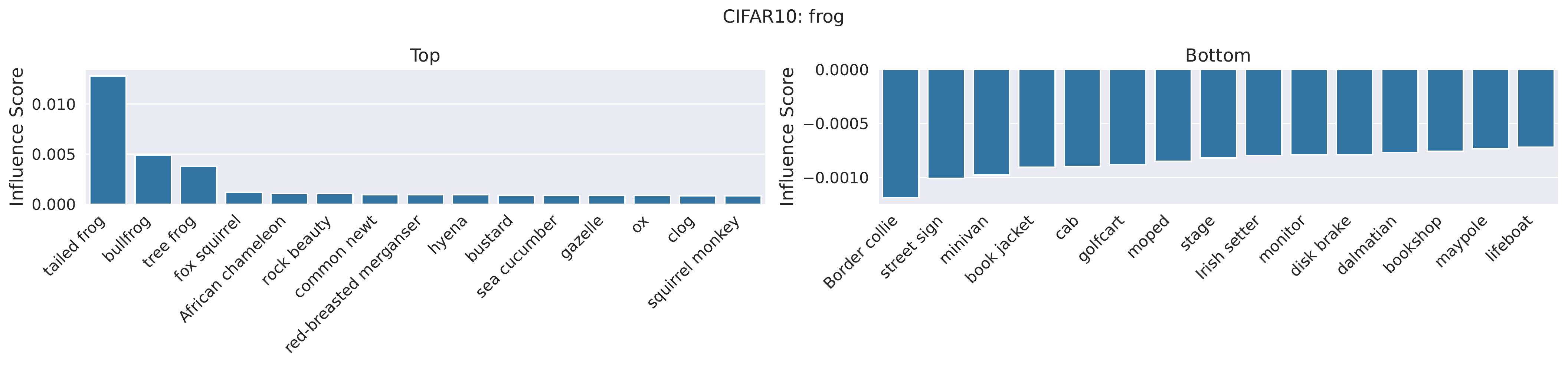}
    \includegraphics[width=\linewidth]{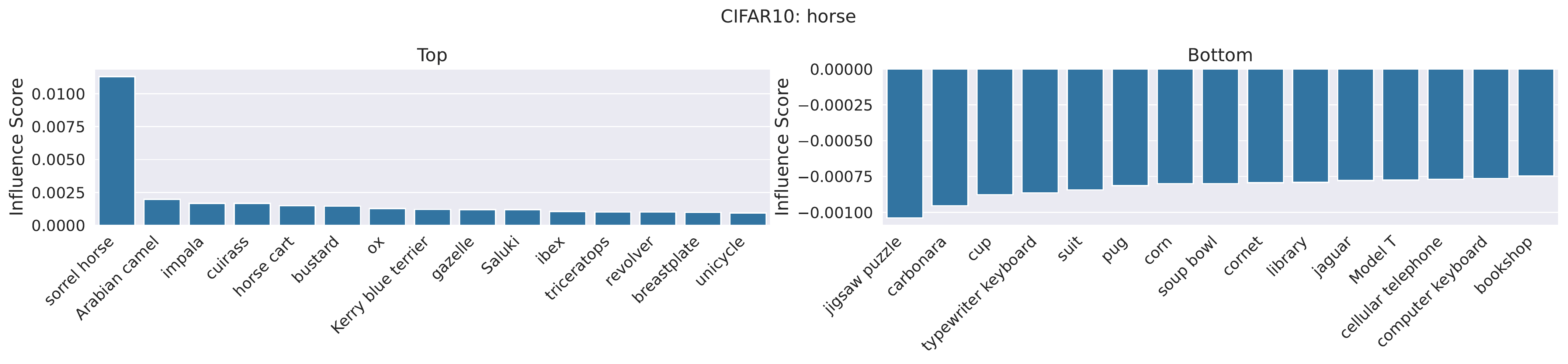}
    \includegraphics[width=\linewidth]{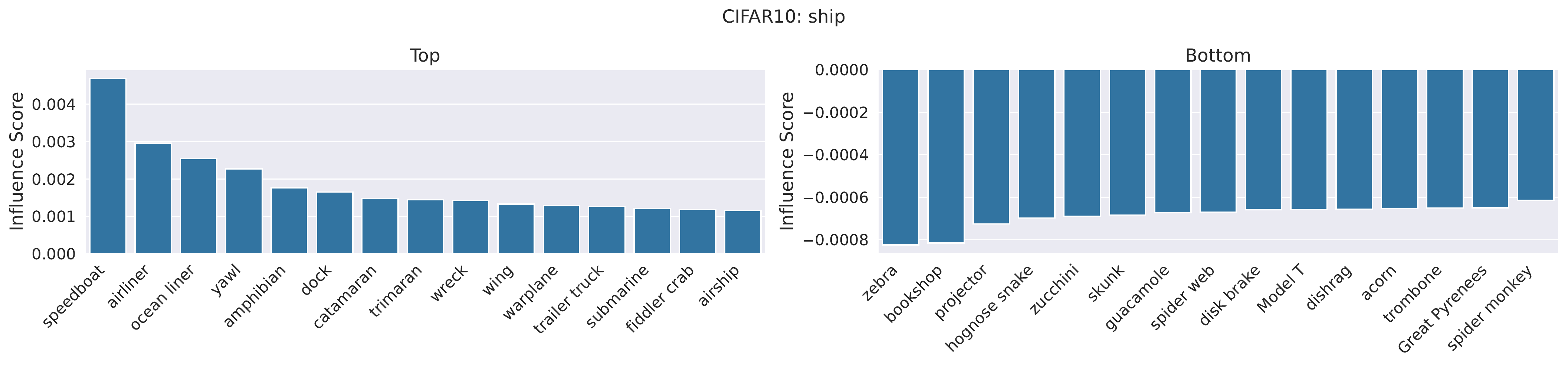}
    \includegraphics[width=\linewidth]{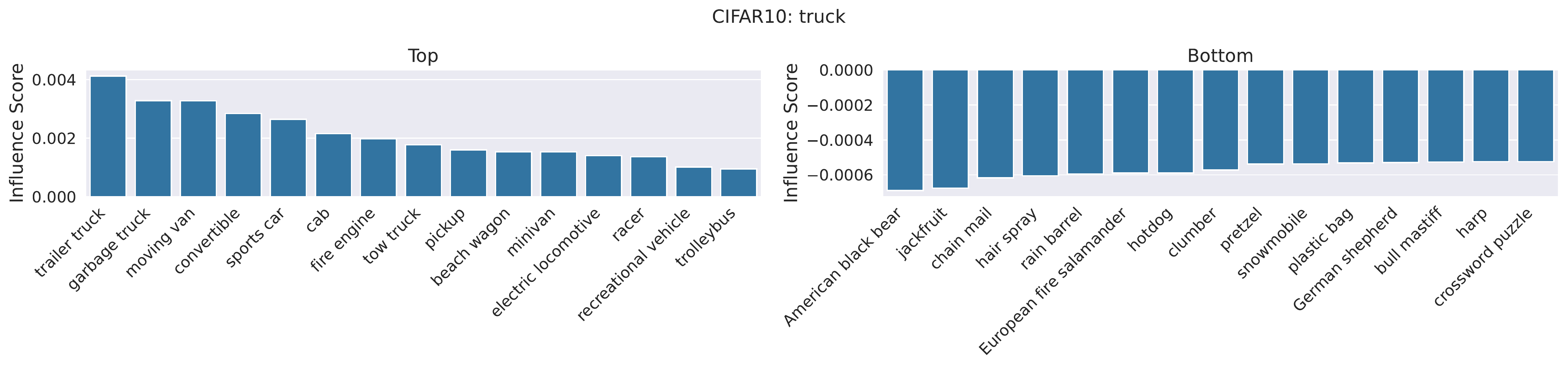}
    \caption{Top and bottom influencing ImageNet classes for all CIFAR-10 classes.}
\end{figure}

\clearpage
\subsection{More examples of extracted subpopulations from the target dataset}
Here, we depict more examples of extracting subpopulations from the target dataset (as in Figure~\ref{fig:extracting-subpopulations} of the main paper).
\label{app:examples-subpop}

\begin{figure}[h!]
    \begin{subfigure}[b]{\linewidth}
        \includegraphics[width=0.45\linewidth]{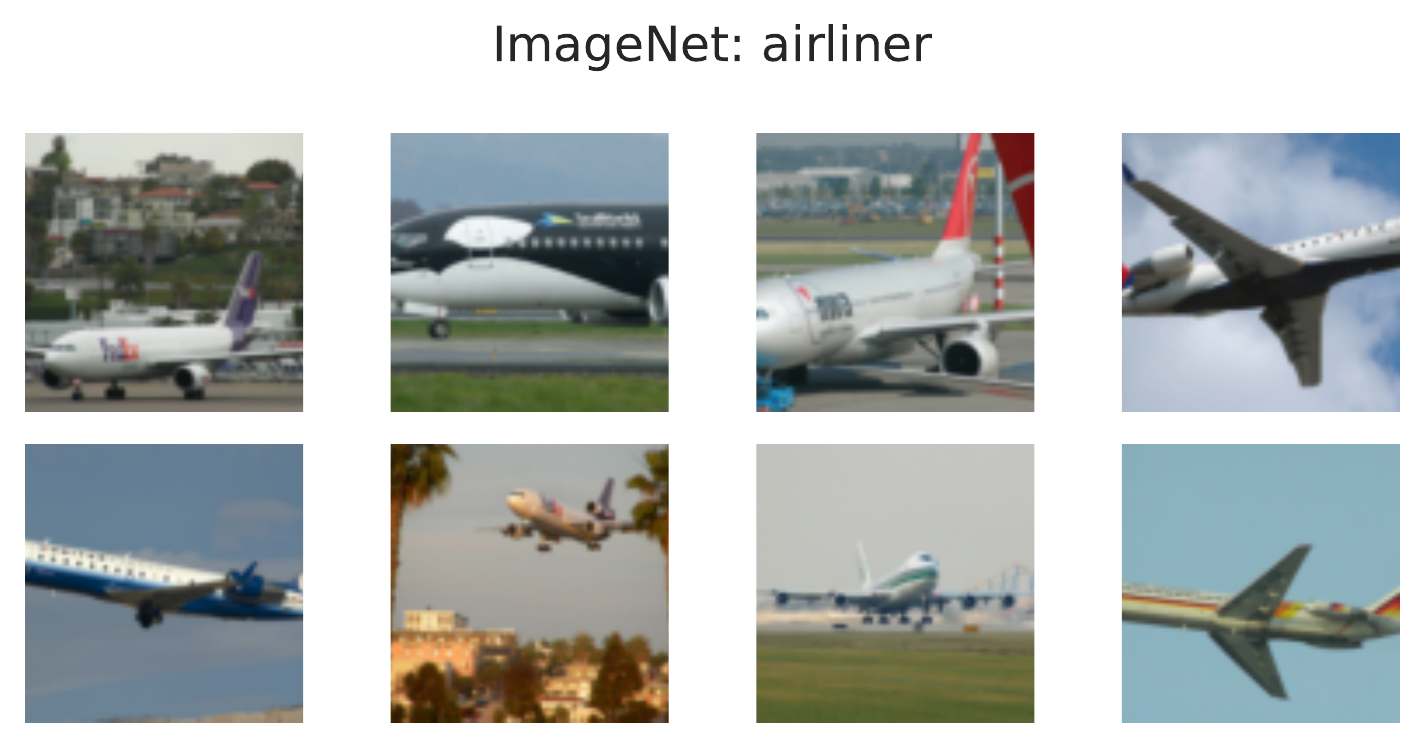}
        \hfill
        \includegraphics[width=0.45\linewidth]{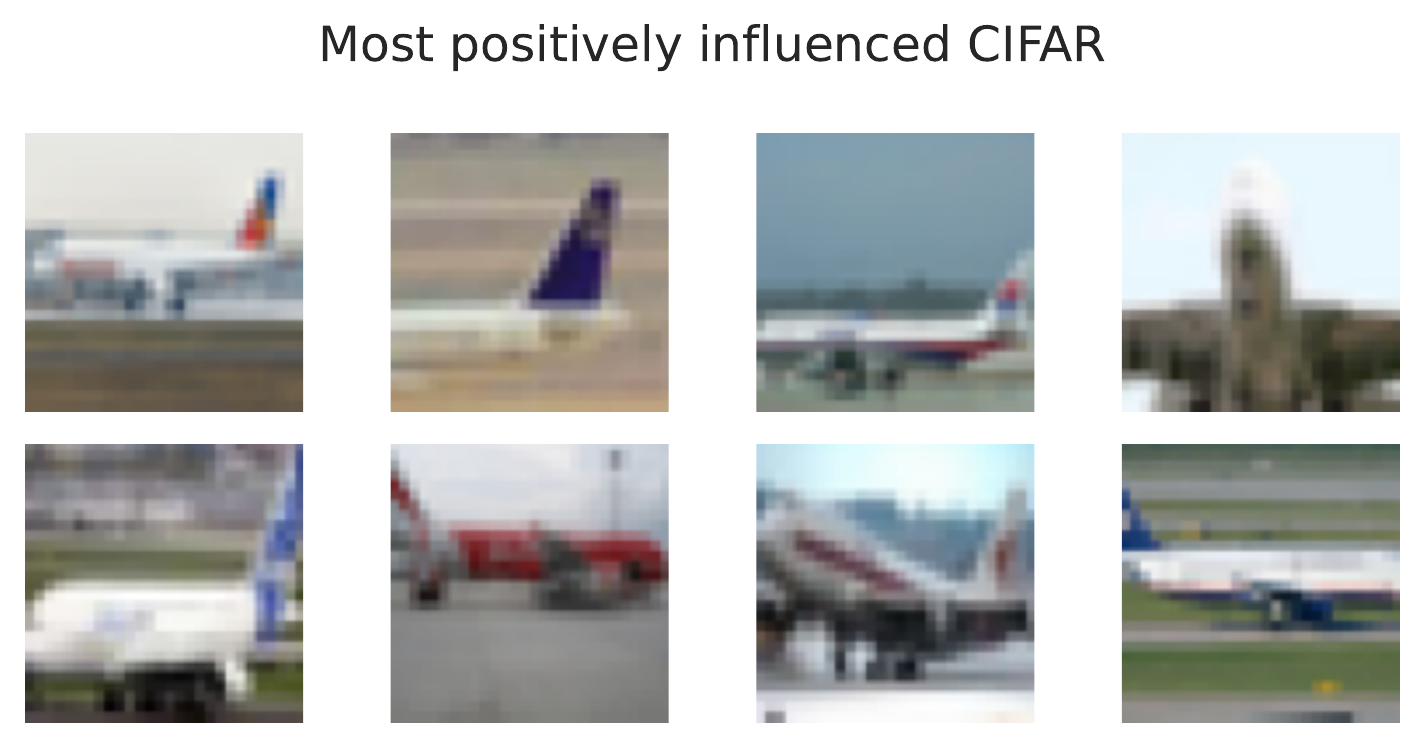}
        \caption{Airliner}
    \end{subfigure}

    \begin{subfigure}[b]{\linewidth}
        \includegraphics[width=0.45\linewidth]{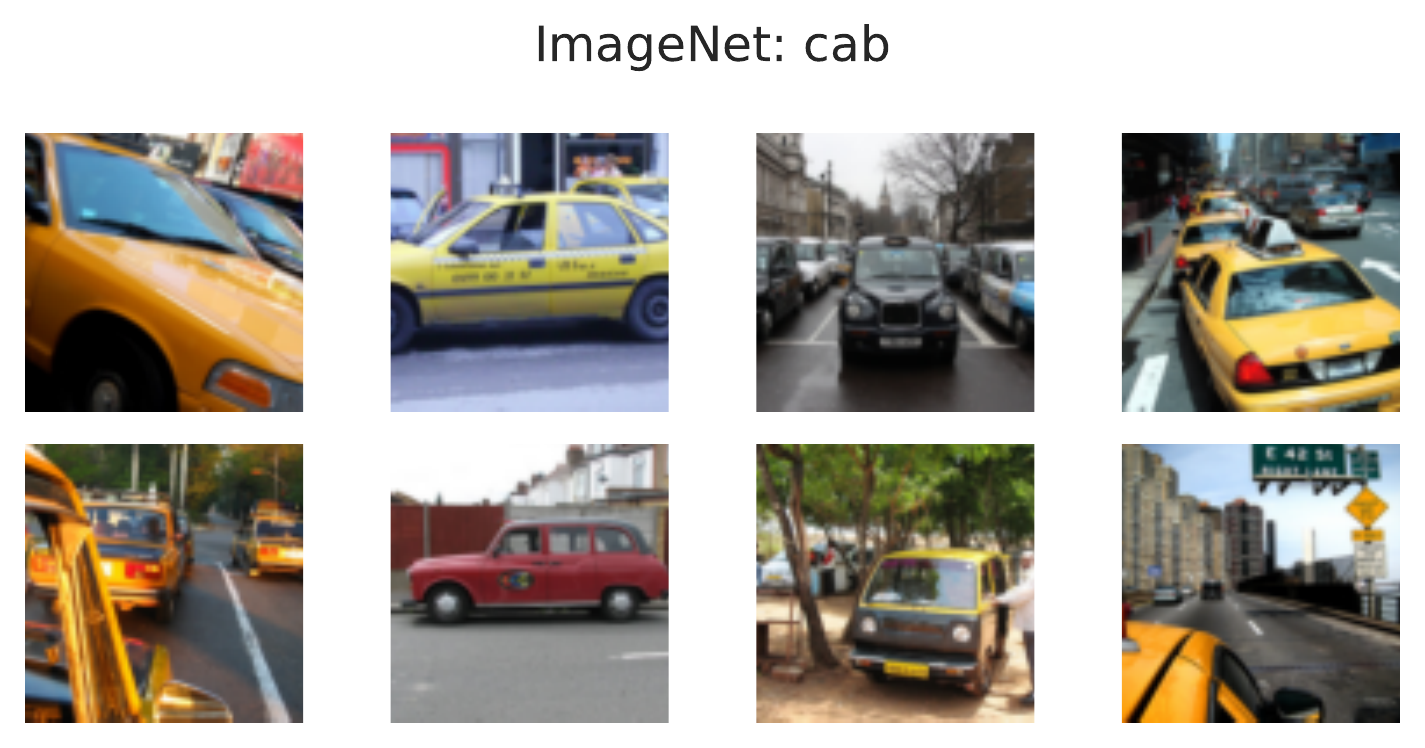}
        \hfill
        \includegraphics[width=0.45\linewidth]{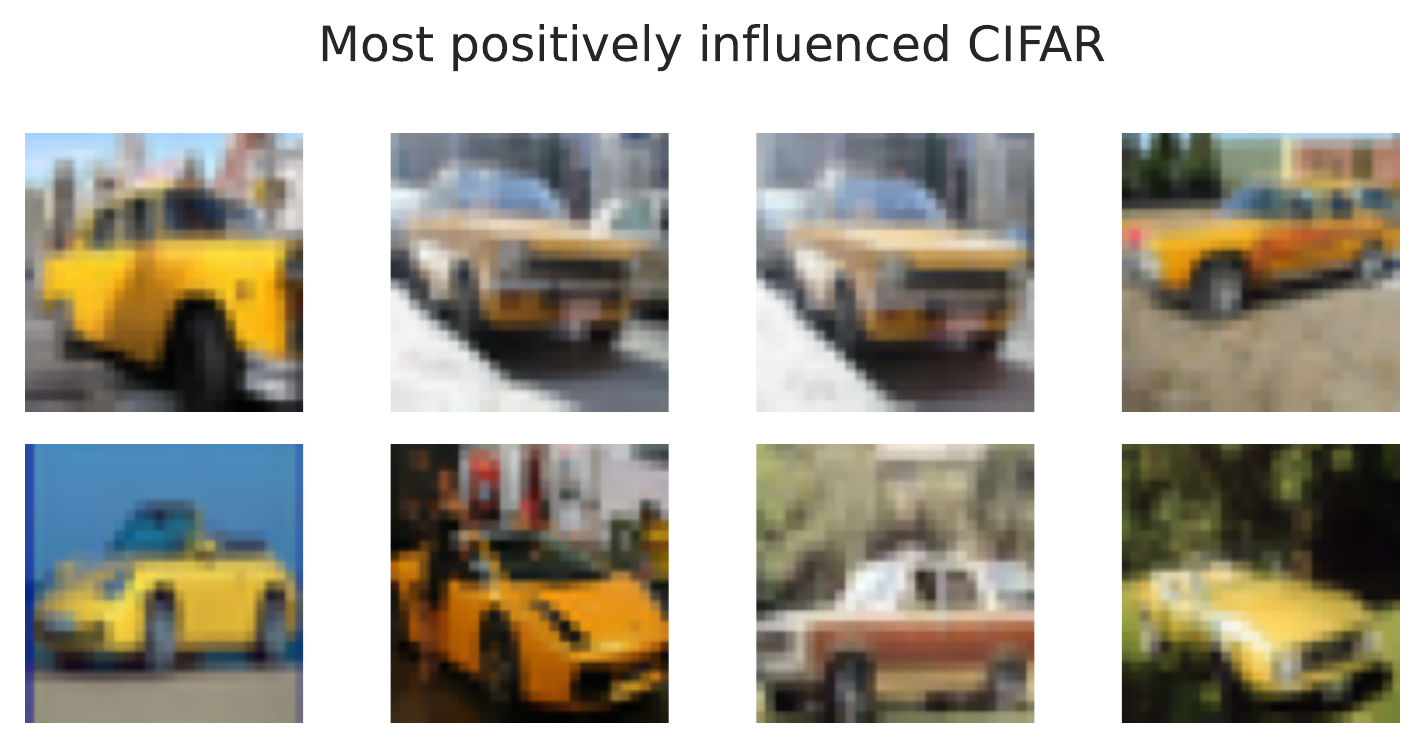}
        \caption{Cab}
    \end{subfigure}

    \begin{subfigure}[b]{\linewidth}
        \includegraphics[width=0.45\linewidth]{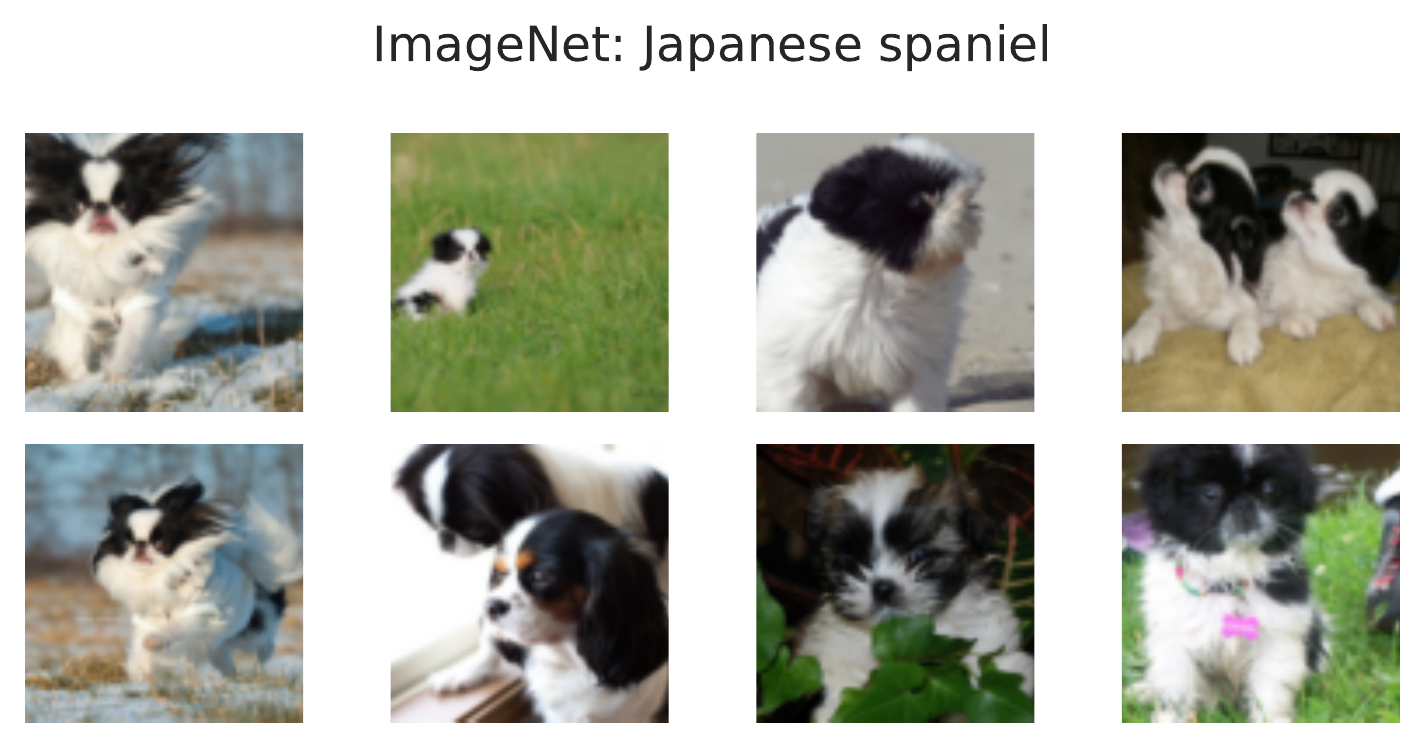}
        \hfill
        \includegraphics[width=0.45\linewidth]{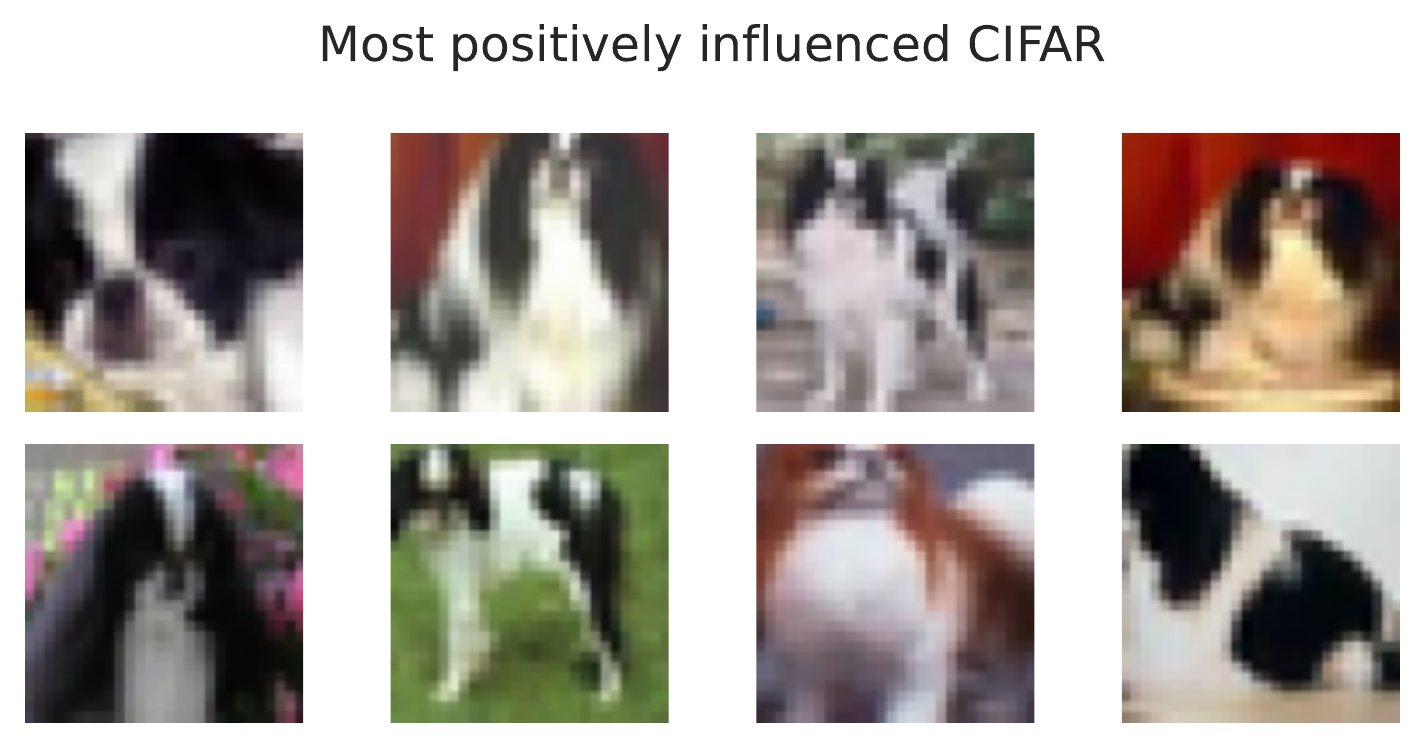}
        \caption{Japanese Spaniel}
    \end{subfigure}

    \begin{subfigure}[b]{\linewidth}
        \includegraphics[width=0.45\linewidth]{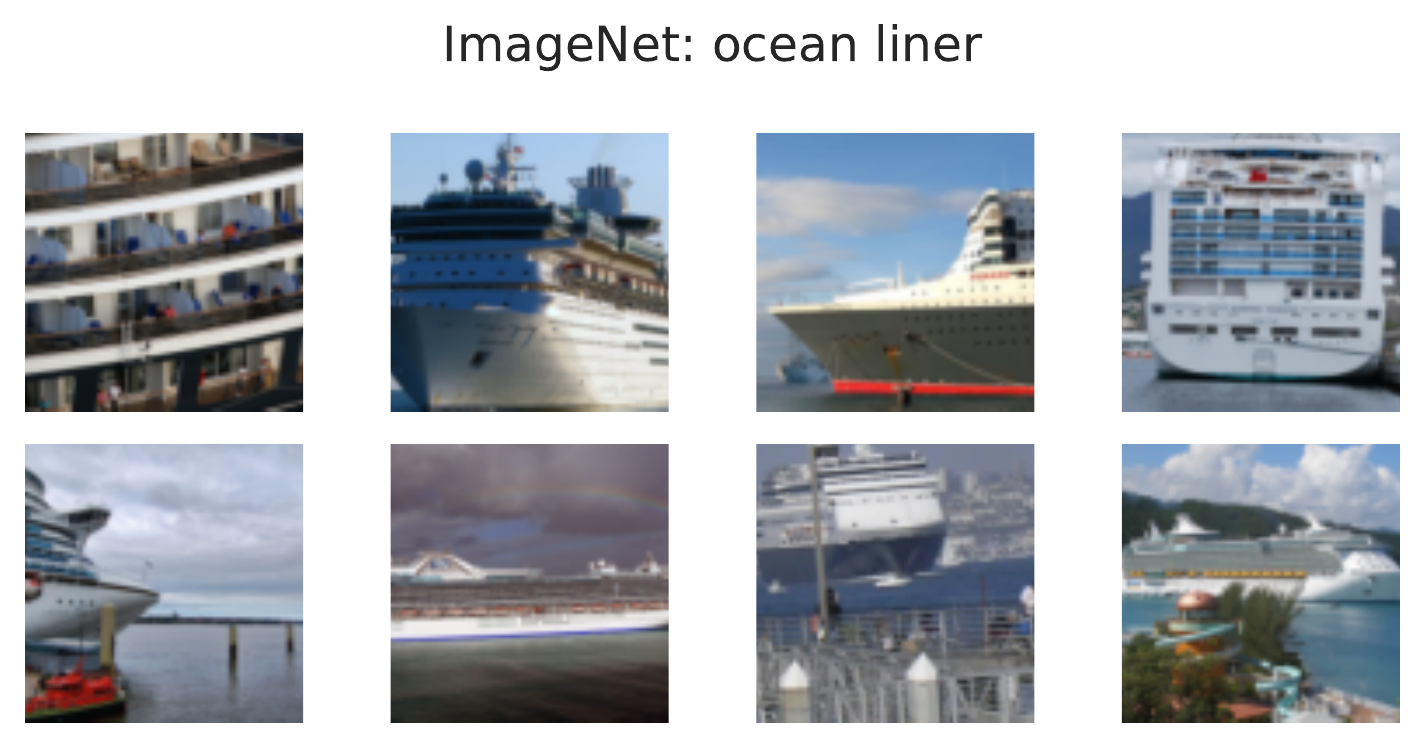}
        \hfill
        \includegraphics[width=0.45\linewidth]{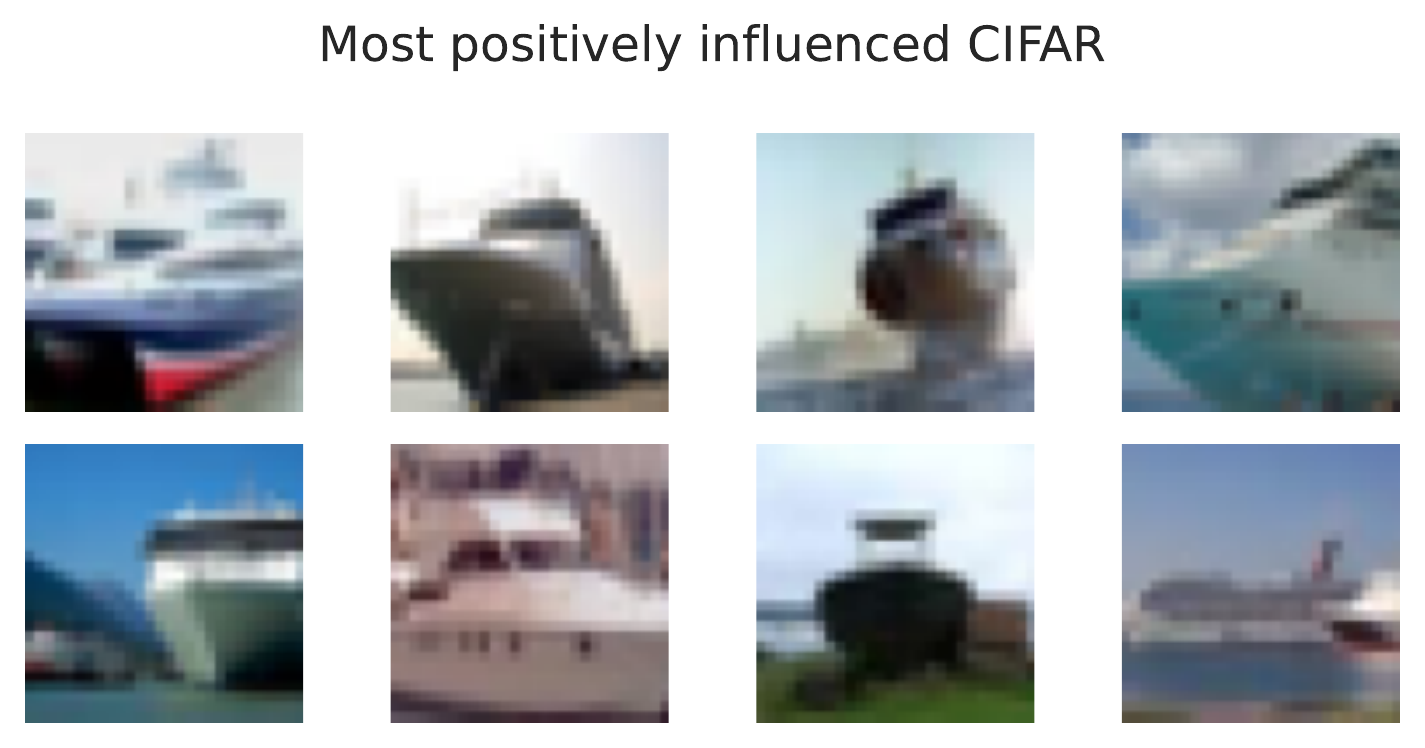}
        \caption{Ocean Liner}
    \end{subfigure}
    \caption{For each ImageNet class, we show the CIFAR-10 examples which were most positively influenced by that ImageNet class.}
\end{figure}

\clearpage
\subsection{More examples of transfer of shape and texture feature}
We depict more examples of ImageNet influencers which transfer shape or texture features (as in Figure~\ref{fig:explaining-imagenet-classes-transfer}).
\label{app:examples-feature-transfer-unrelated}

\begin{figure}[h!]
    \centering
    \begin{subfigure}[b]{0.75\linewidth}
        \centering
        \includegraphics[width=0.45\linewidth]{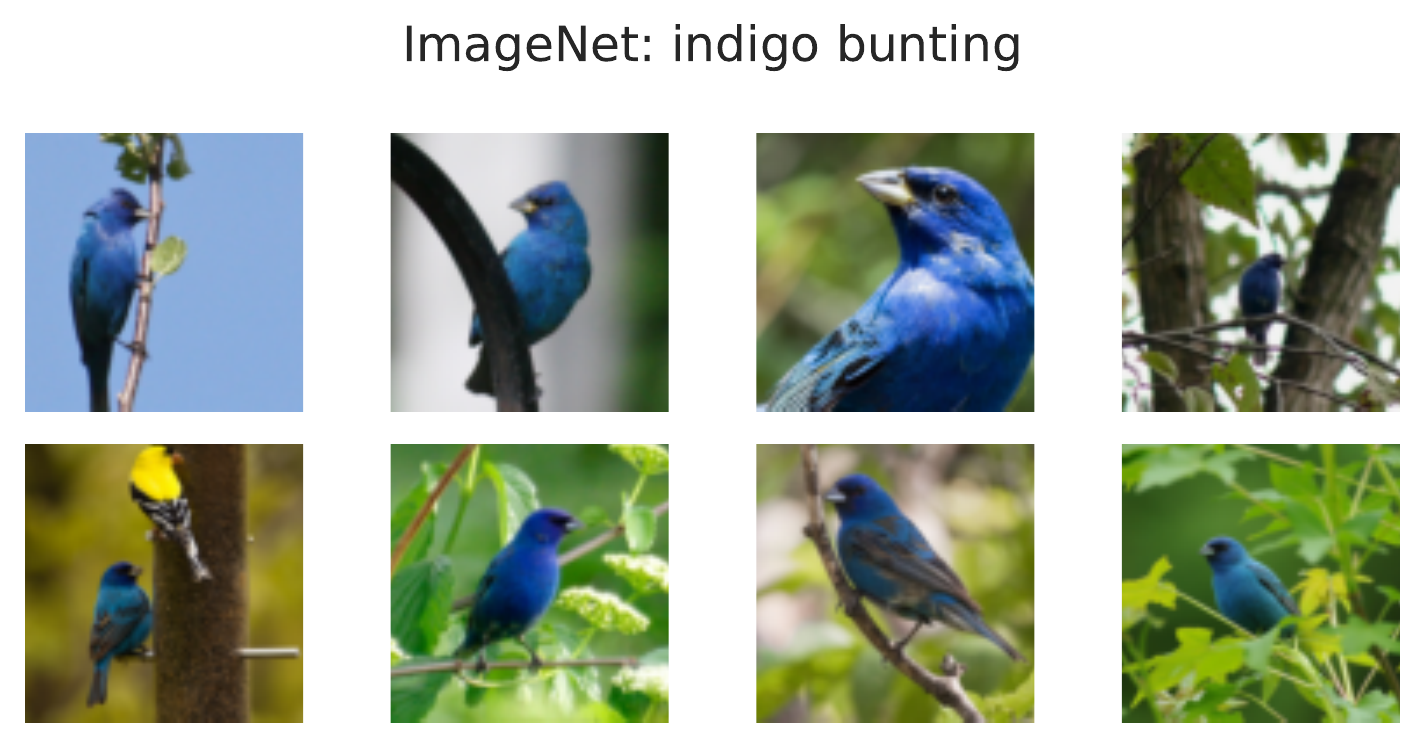}\\
        \includegraphics[width=0.45\linewidth]{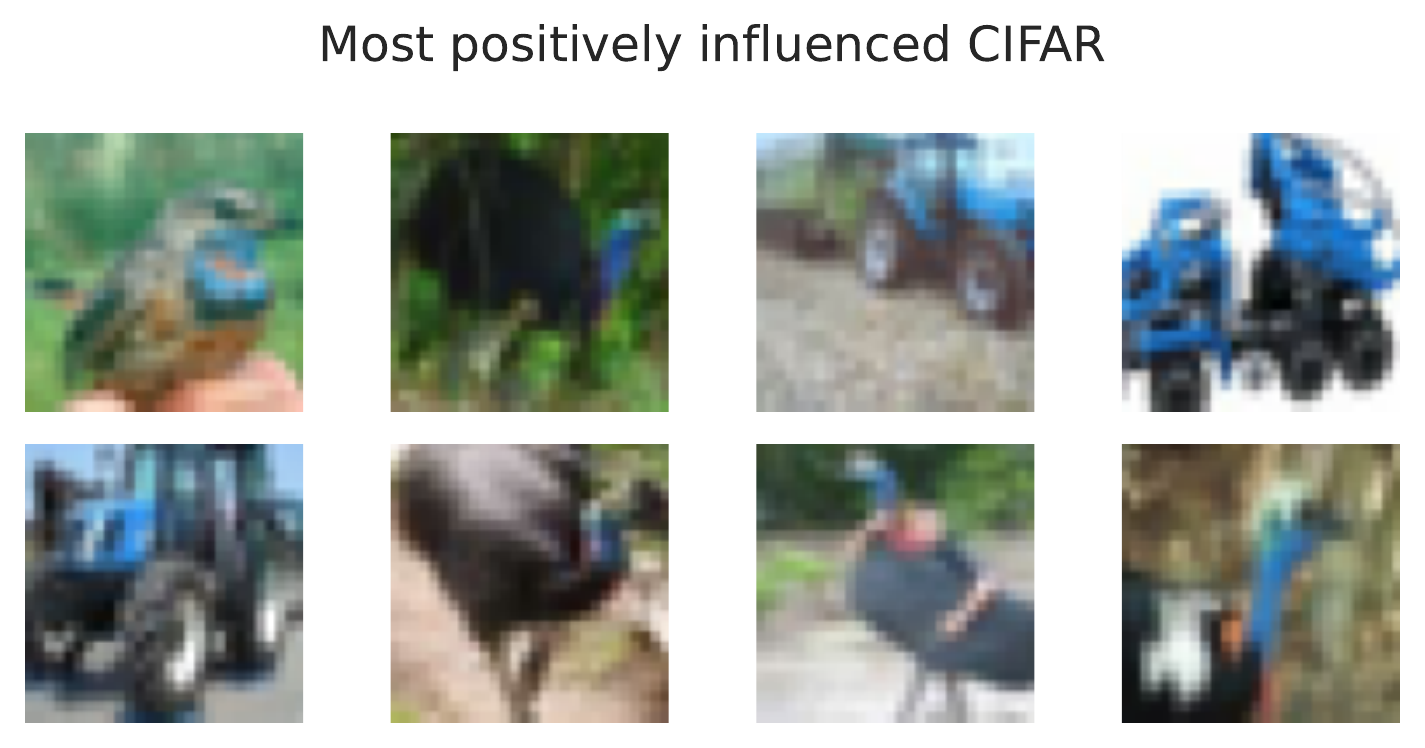}
        \hfill
        \includegraphics[width=0.45\linewidth]{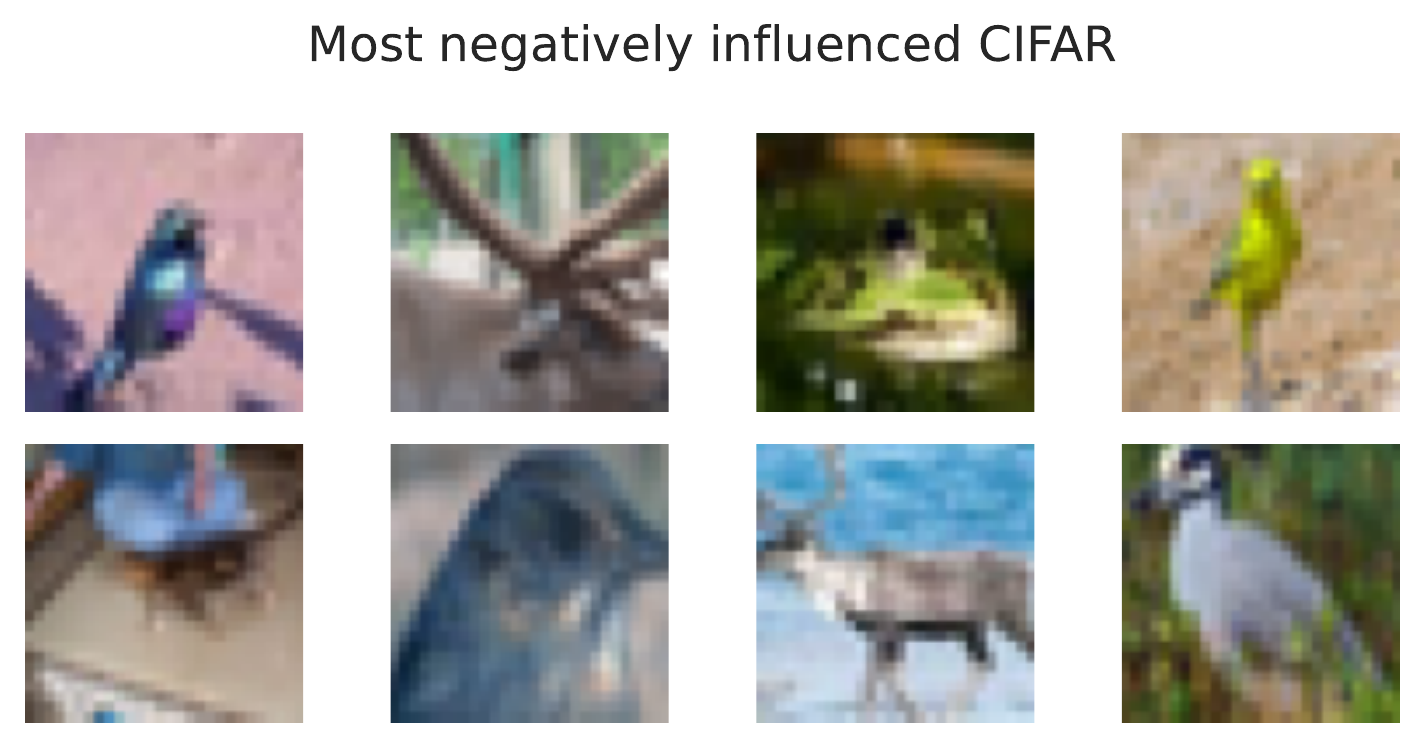}
        \caption{Indigo Bunting}
    \end{subfigure}
    \begin{subfigure}[b]{0.75\linewidth}
        \centering
        \includegraphics[width=0.45\linewidth]{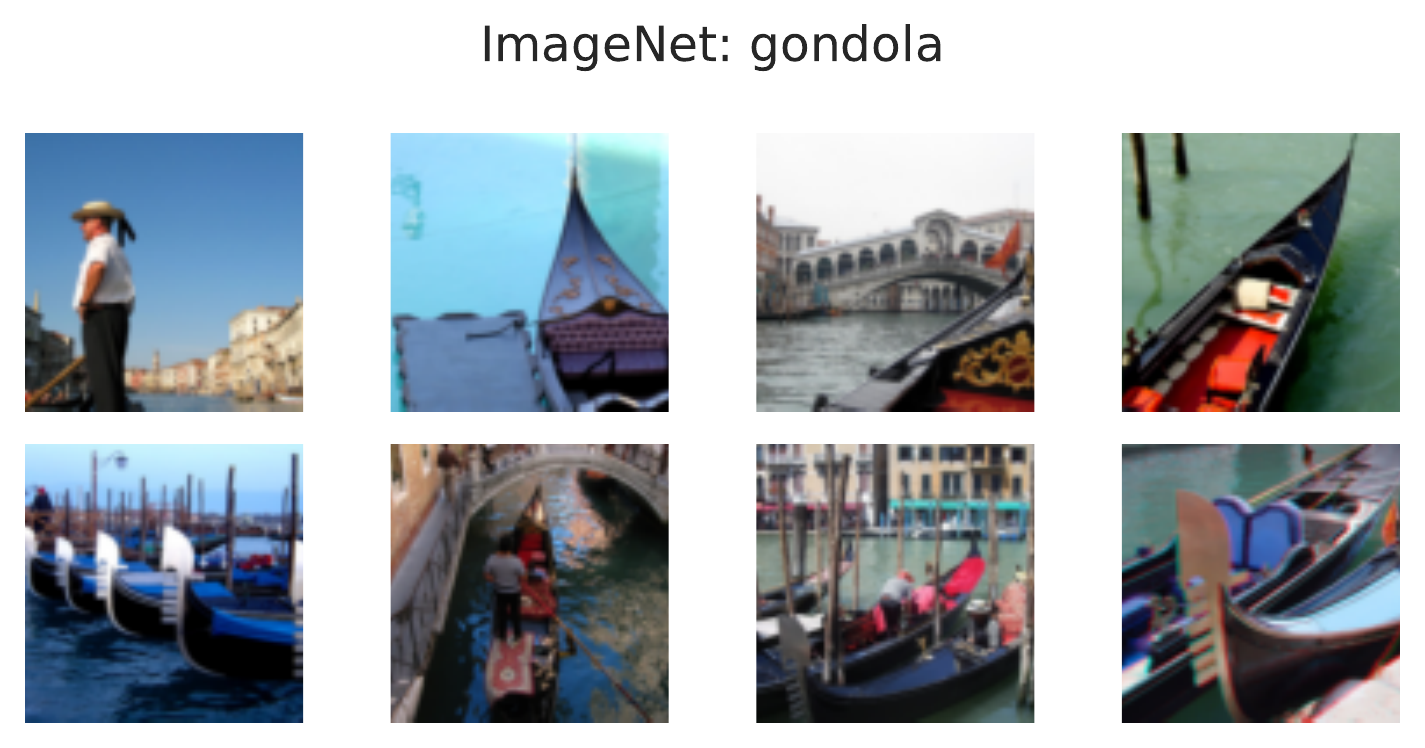}\\
        \includegraphics[width=0.45\linewidth]{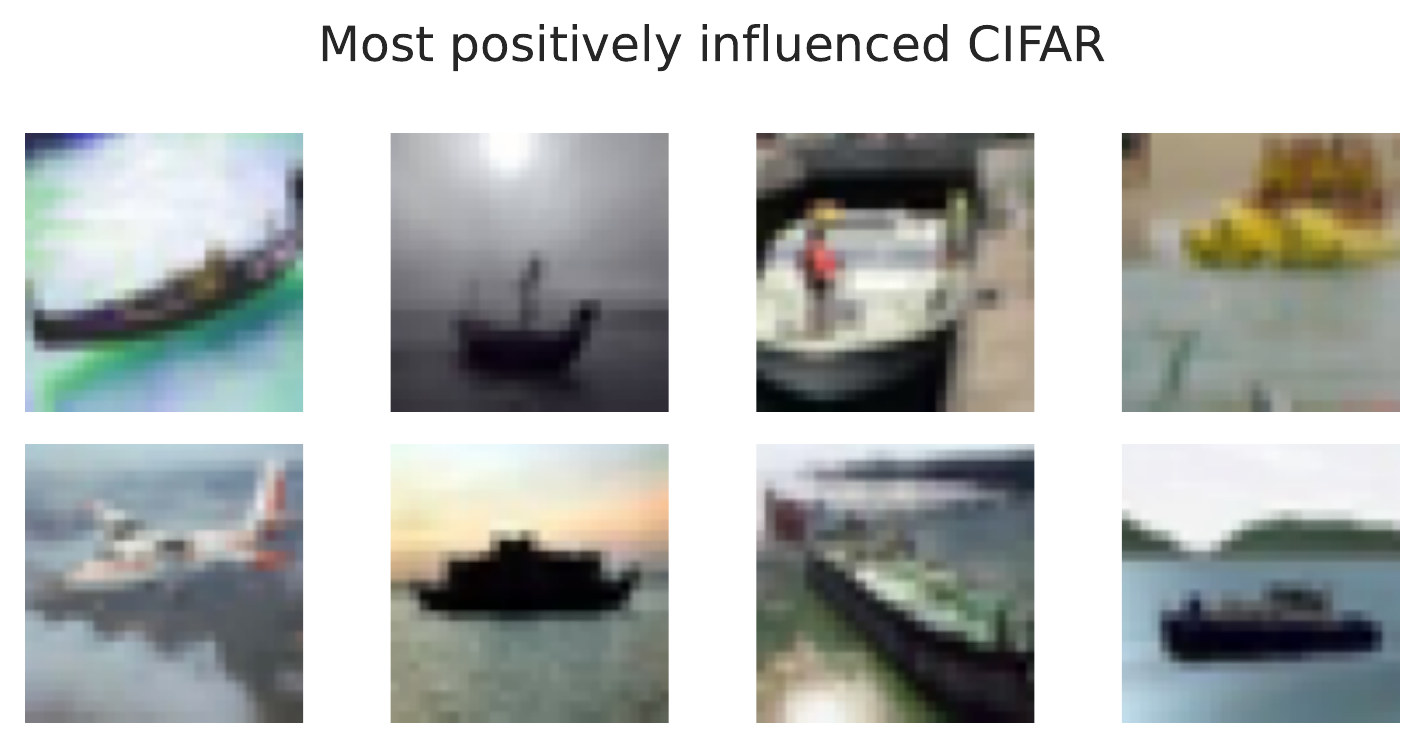}
        \hfill
        \includegraphics[width=0.45\linewidth]{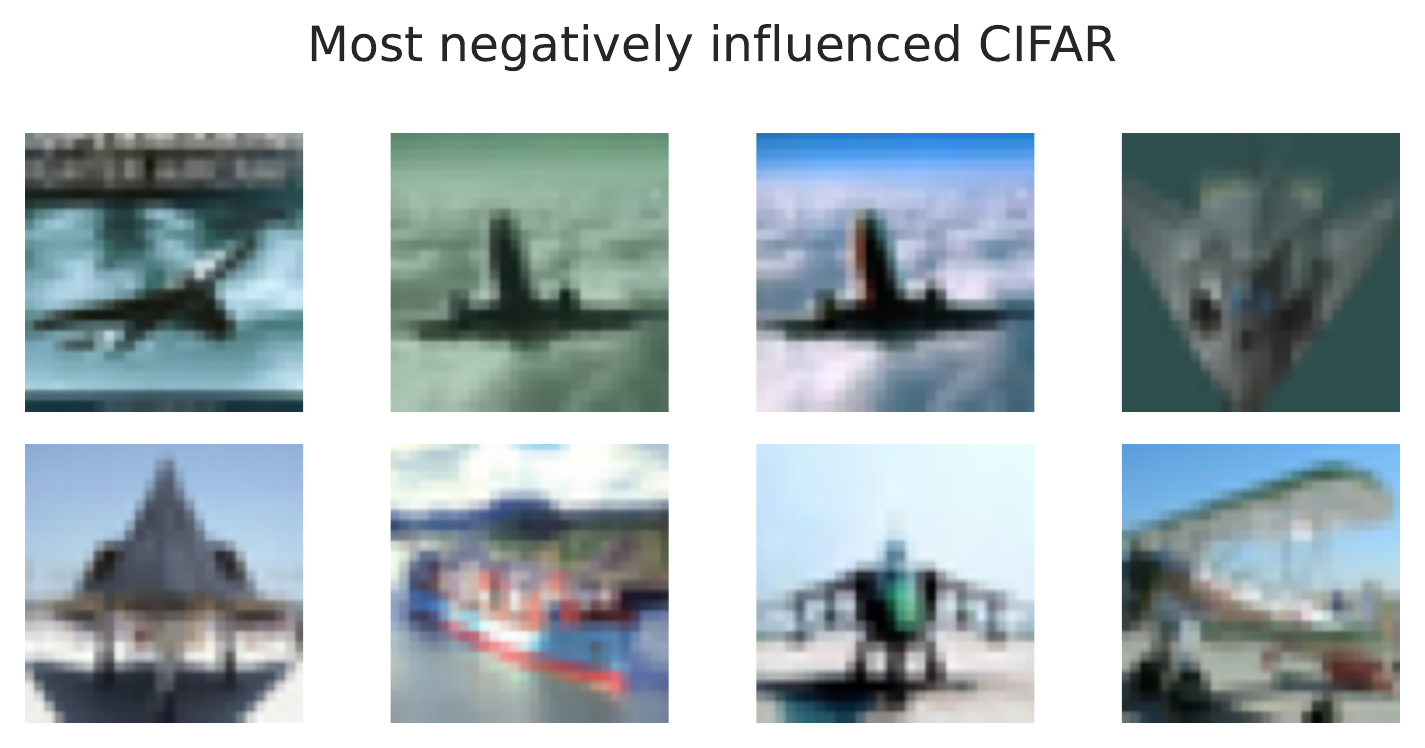}
        \caption{Gondola}
    \end{subfigure}
    \begin{subfigure}[b]{0.75\linewidth}
        \centering
        \includegraphics[width=0.45\linewidth]{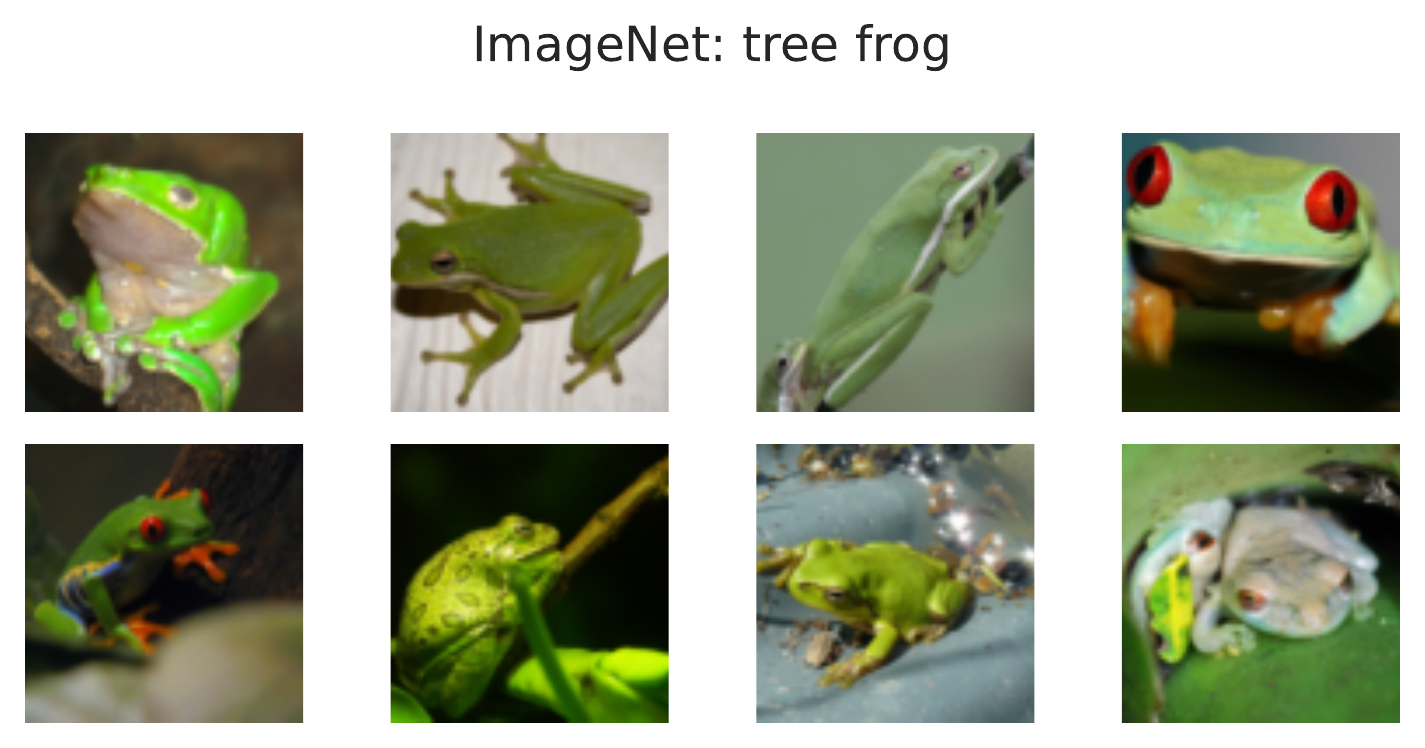}\\
        \includegraphics[width=0.45\linewidth]{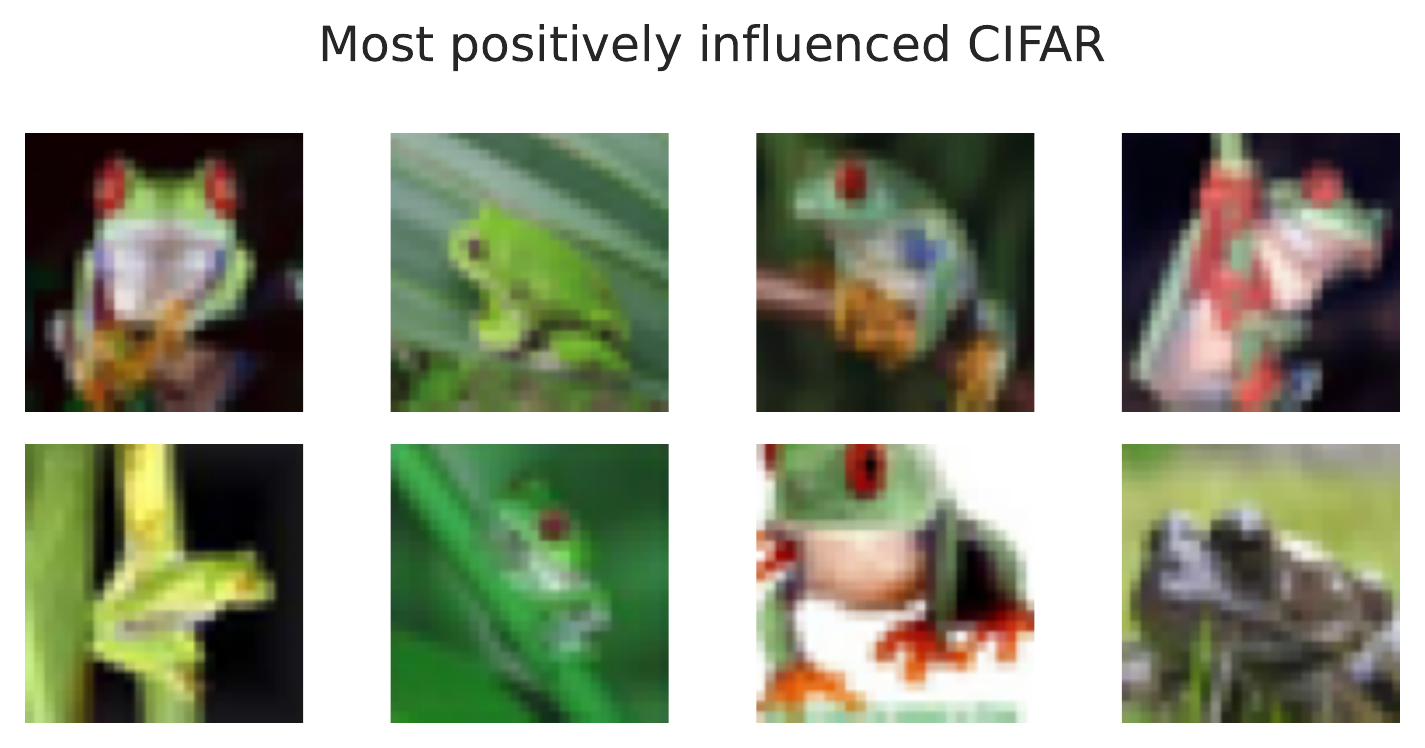}
        \hfill
        \includegraphics[width=0.45\linewidth]{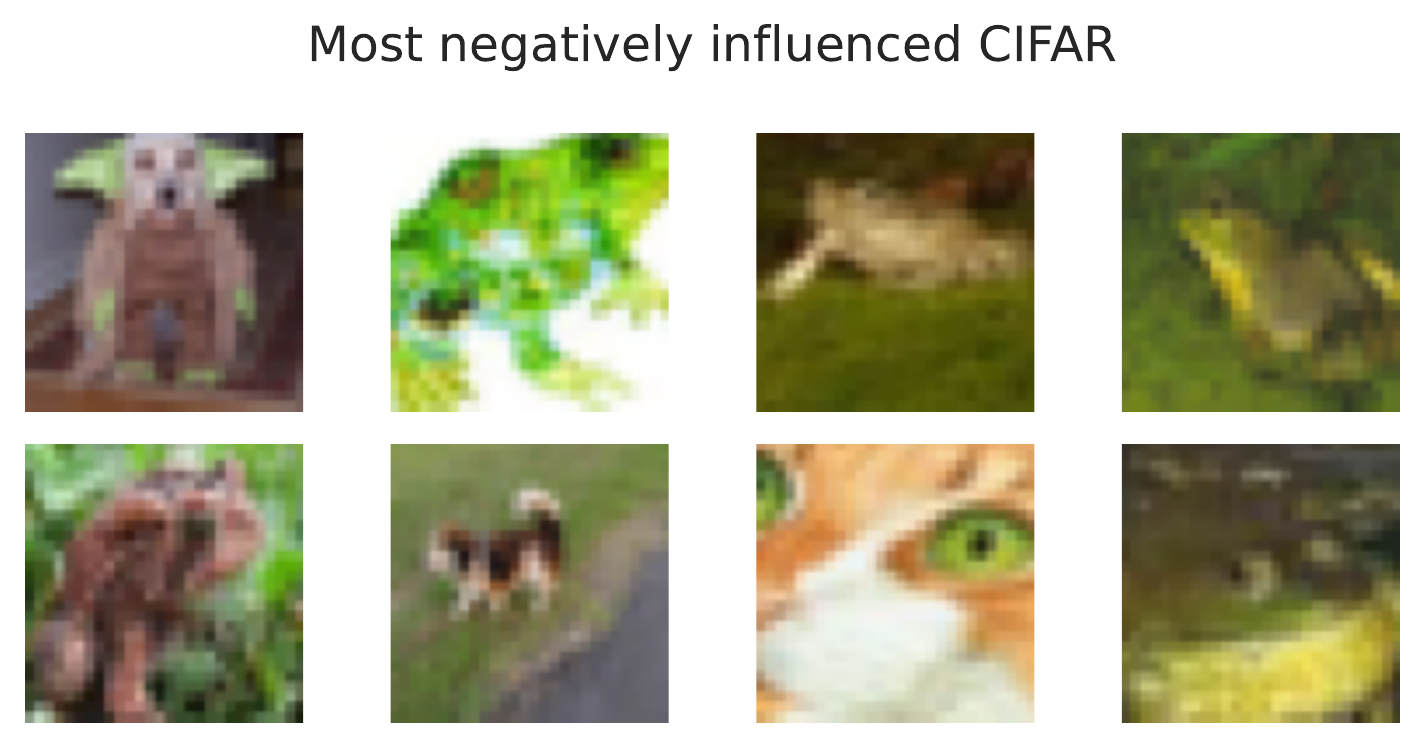}
        \caption{Tree Frog}
    \end{subfigure}
    \caption{For each ImageNet class, we show the CIFAR-10 examples which were most highly influenced by that ImageNet class.}
\end{figure}

\clearpage
\subsection{More examples of debugging mistakes of transfer model using influencers}
\label{app:debugging-mistakes}
\begin{figure}[htbp!]
\begin{subfigure}[b]{\linewidth}
    \centering
    \includegraphics[width=0.75\linewidth]{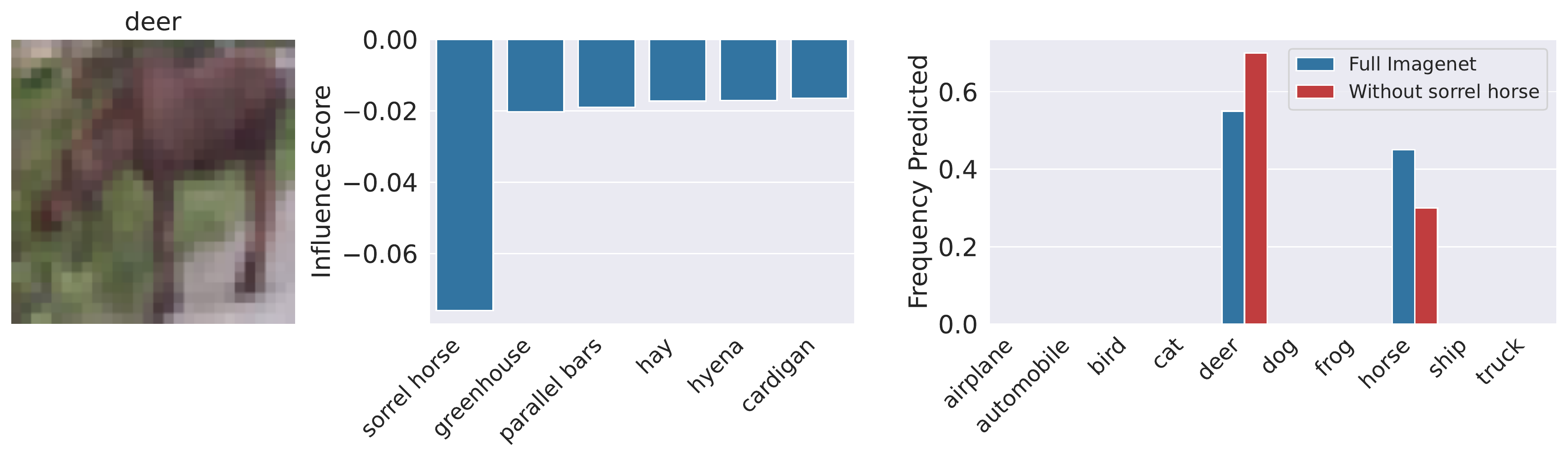}
    \caption{}
    \label{app_fig:debug_1}
\end{subfigure}
\begin{subfigure}[b]{\linewidth}
    \centering
    \includegraphics[width=0.75\linewidth]{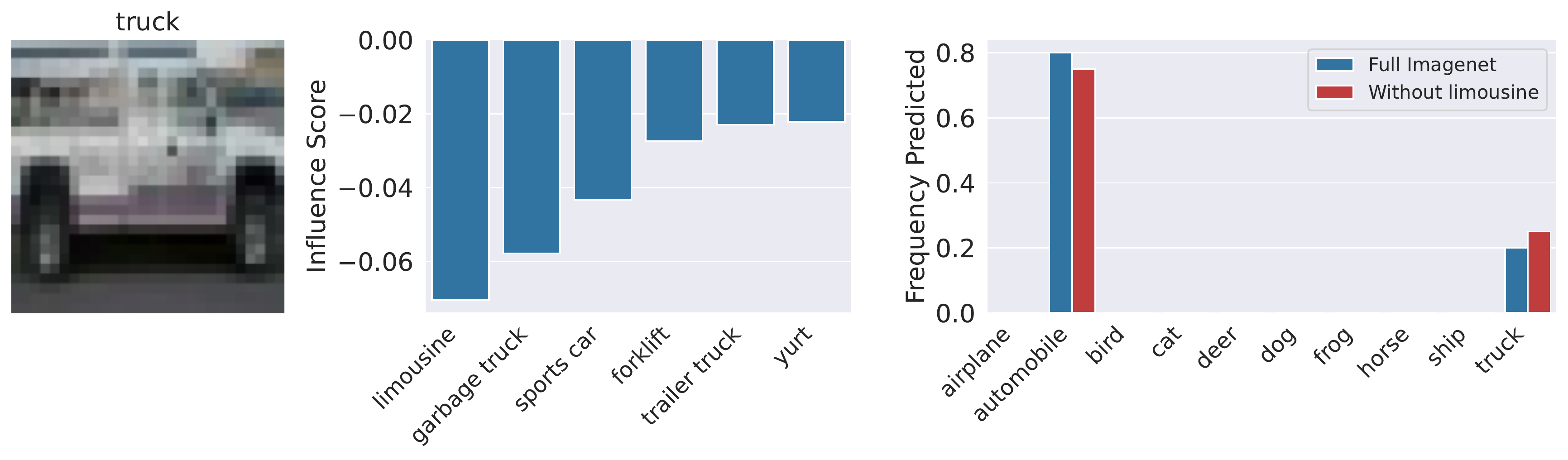}
    \caption{}
    \label{app_fig:debug_2}
\end{subfigure}
\begin{subfigure}[b]{\linewidth}
    \centering
    \includegraphics[width=0.75\linewidth]{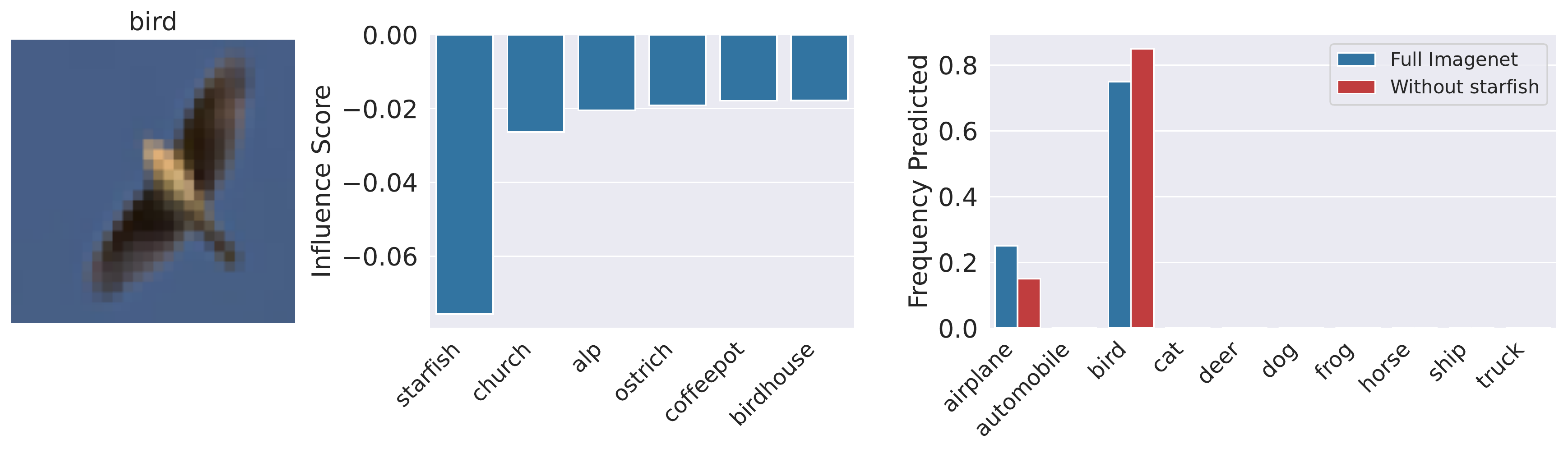}
    \caption{}
    \label{app_fig:debug_3}
\end{subfigure}
\begin{subfigure}[b]{\linewidth}
    \centering
    \includegraphics[width=0.75\linewidth]{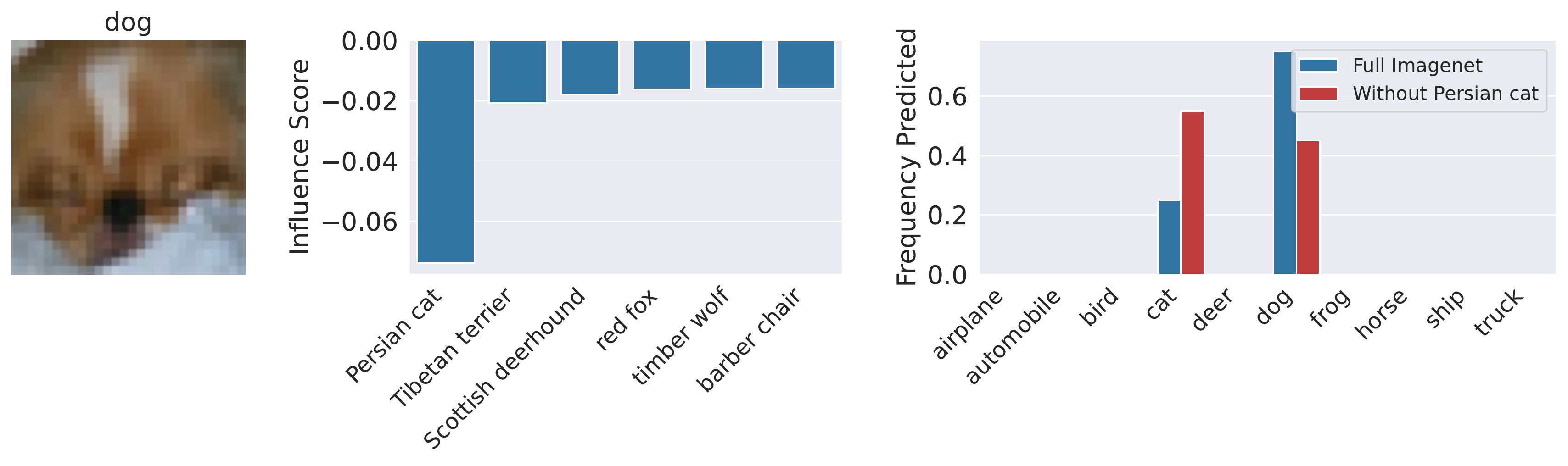}
    \caption{}
    \label{app_fig:debug_4}
\end{subfigure}
\caption{More examples of debugging transfer mistakes through our framework (c.f. Figure~\ref{fig:explaining-mistakes} in the main paper). For each CIFAR-10 image (\textbf{left}), we plot their most negative influencers  (\textbf{middle}). On the \textbf{right}, we plot for each image the fraction (over 20 runs) of times that our transfer model predicts each class with and without the most negative influencer. While in most cases (a, b, c) we find that removing the most negative influencer helps the model predict the image correctly, in some cases removing the negative influencer does not outweigh the impact of removing images from the source dataset (d).}
\label{app_fig:debug}
\end{figure}

We display more examples of how our influences can be used to debug the mistakes of the transfer model, as presented in Figure~\ref{fig:explaining-mistakes} in the main paper. We find that, in most cases (Figure~\ref{app_fig:debug_1}, \ref{app_fig:debug_2}, \ref{app_fig:debug_3}), removing the top negative influencer improves the model's performance on the particular image. There are a few examples where removing the top negative influencer hurts the model's performance on the image (Figure~\ref{app_fig:debug_4}).

\clearpage

\paragraph{Quantitative analysis.} How often does removing the most negative influencer actually improve the prediction on an image? For each of the following 14 classes, we run 20 runs of the ImageNet $\rightarrow$ CIFAR-10 fixed transfer pipeline while excluding that single class from the source dataset: [``sorrel horse'', ``limousine'', ``minivan'', ``fireboat'', ``ocean liner'', ``Arabian camel'', ``Persian cat'', ``ostrich'', ``gondola'', ``pool table'', ``starfish'', ``rapeseed'', ``tailed frog'', ``trailer truck'']. We compare against running the pipeline with 20 runs of the entire ImageNet dataset. Then, we look at individual CIFAR-10 images which were highly negatively influenced by one of these ImageNet classes, and check whether the images were predicted correctly more or less often when the top negative influencers were removed from the source dataset.

Of the 30 most negatively influenced ImageNet class/CIFAR-10 image pairs, 26 of them had the most negative ImageNet influencer in the above classes. Of those, 61.5\% were predicted correctly more often when the negatively influential ImageNet class was removed, 34.6\% were predicted incorrectly more often, and 3.9\% were predicted correctly the same number of times.

We then examine the top 8 most influenced CIFAR-10 images for each of the above 14 ImageNet classes. Of those 112 images, 53\% were predicted correctly more often when the image was removed, 34\% were predicted incorrectly more often,  and 14\% were predicted correctly the same number of times.

We thus find that, for the most influenced CIFAR-10 images, removing the top negative influencer usually improves that specific image's prediction (even though we are removing training data from the source dataset).

\clearpage
\subsection{Do Influences Transfer?}
\label{app:influence-transfer}
\input{sections/app_influence_transfer.tex}

%% file: sections/app_influence_transfer.tex
\subsubsection{Transfer across datasets}
In this section, we seek to understand how much task-specific information is in the transfer influences that we compute. To do so, we use the transfer influences computed for CIFAR-10 in order to perform the counterfactual experiments for other datasets. We find that while using the CIFAR-10 influence values for other target datasets is more meaningful than random, they do not provide the same boost in performance when removing bottom influencers as using the task-specific influences. We thus conclude that the influence values computed by our framework are relatively task-specific.
\begin{figure}[h!]
    \centering
    \begin{subfigure}[c]{0.32\linewidth}
        \includegraphics[width=0.93\linewidth]{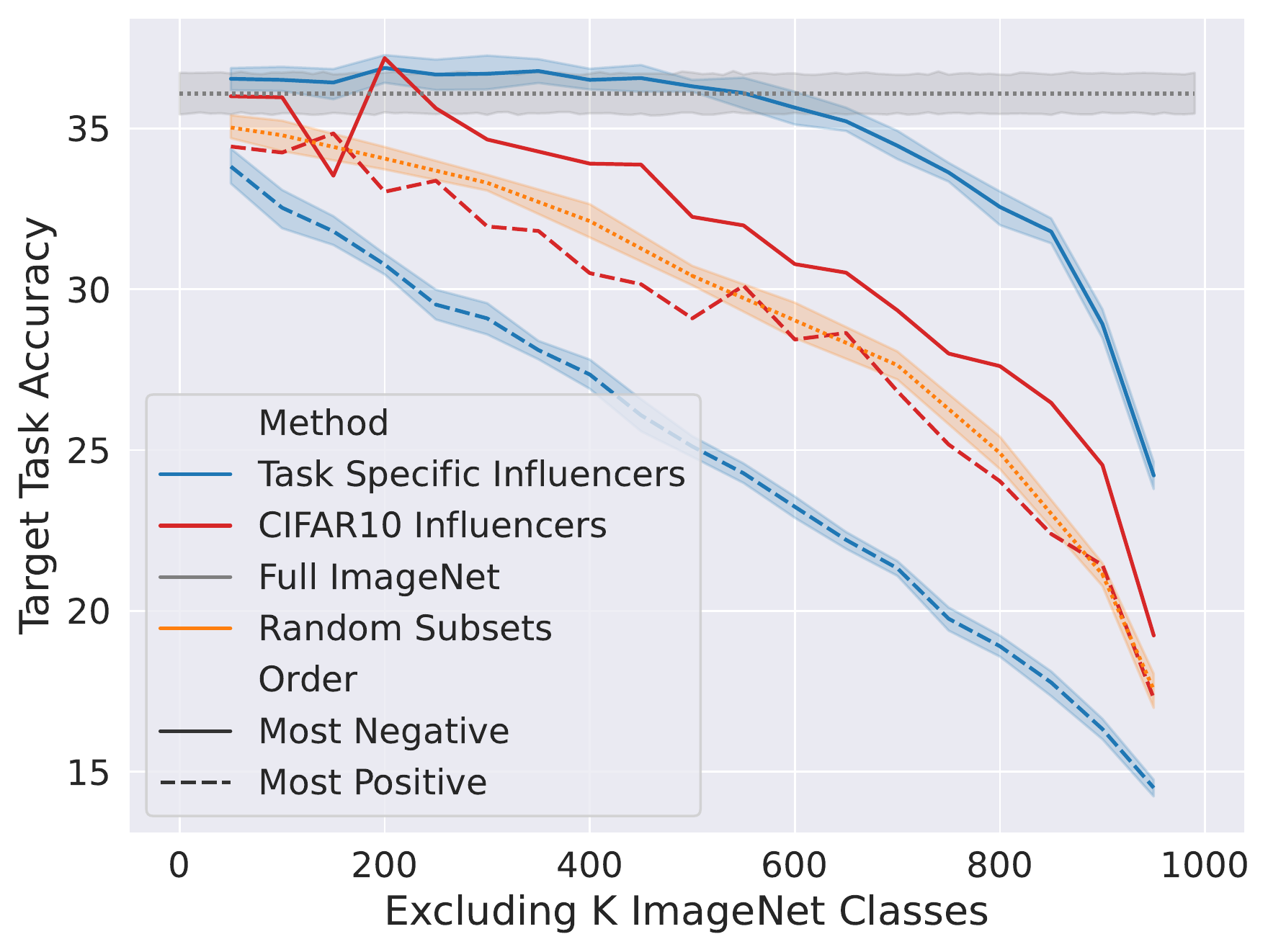}
        \caption{AIRCRAFT}
    \end{subfigure}\hfill
    \begin{subfigure}[c]{0.32\linewidth}
        \includegraphics[width=0.93\linewidth]{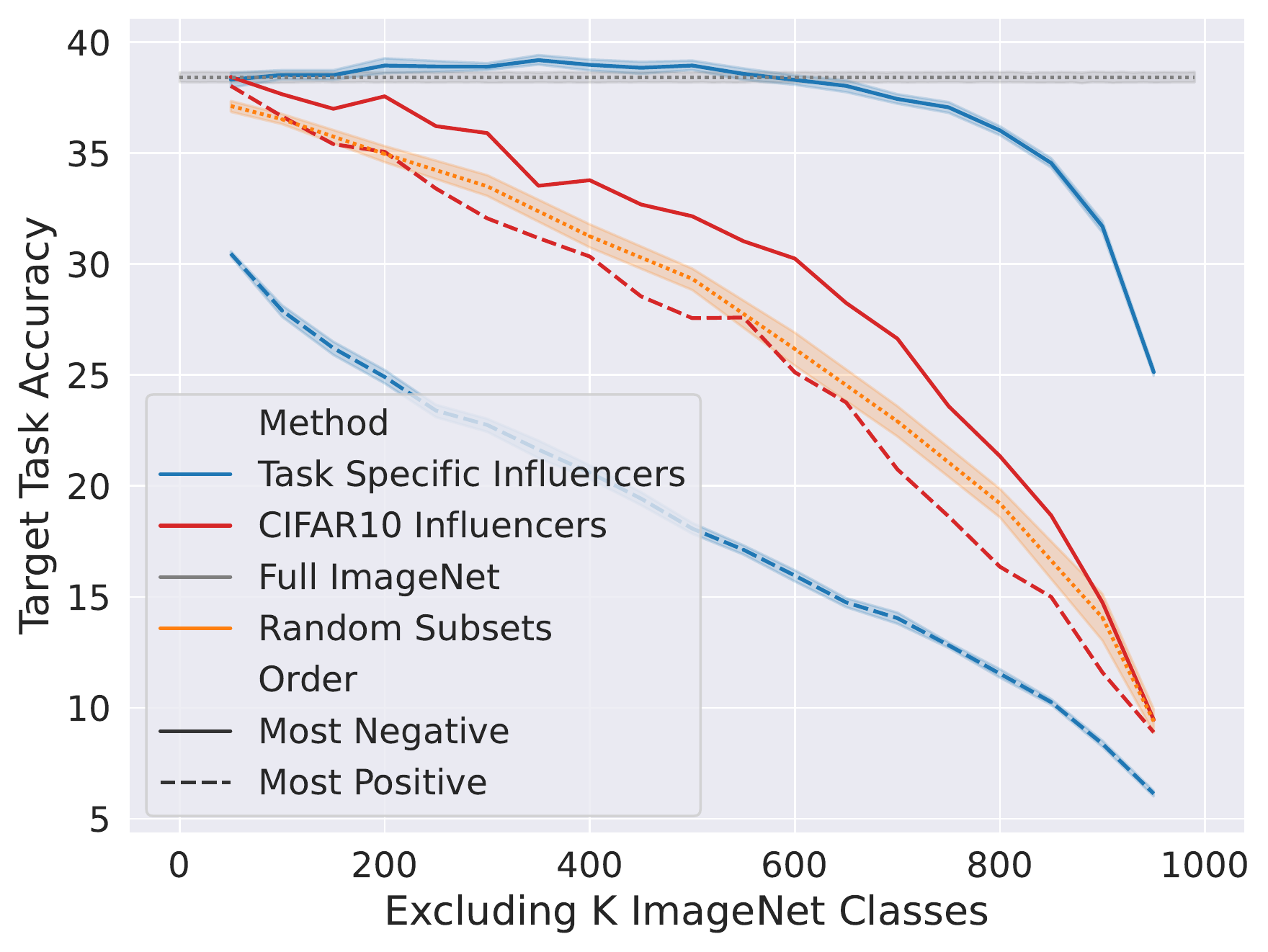}
        \caption{BIRDSNAP}
    \end{subfigure}
    \begin{subfigure}[c]{0.32\linewidth}
        \includegraphics[width=0.93\linewidth]{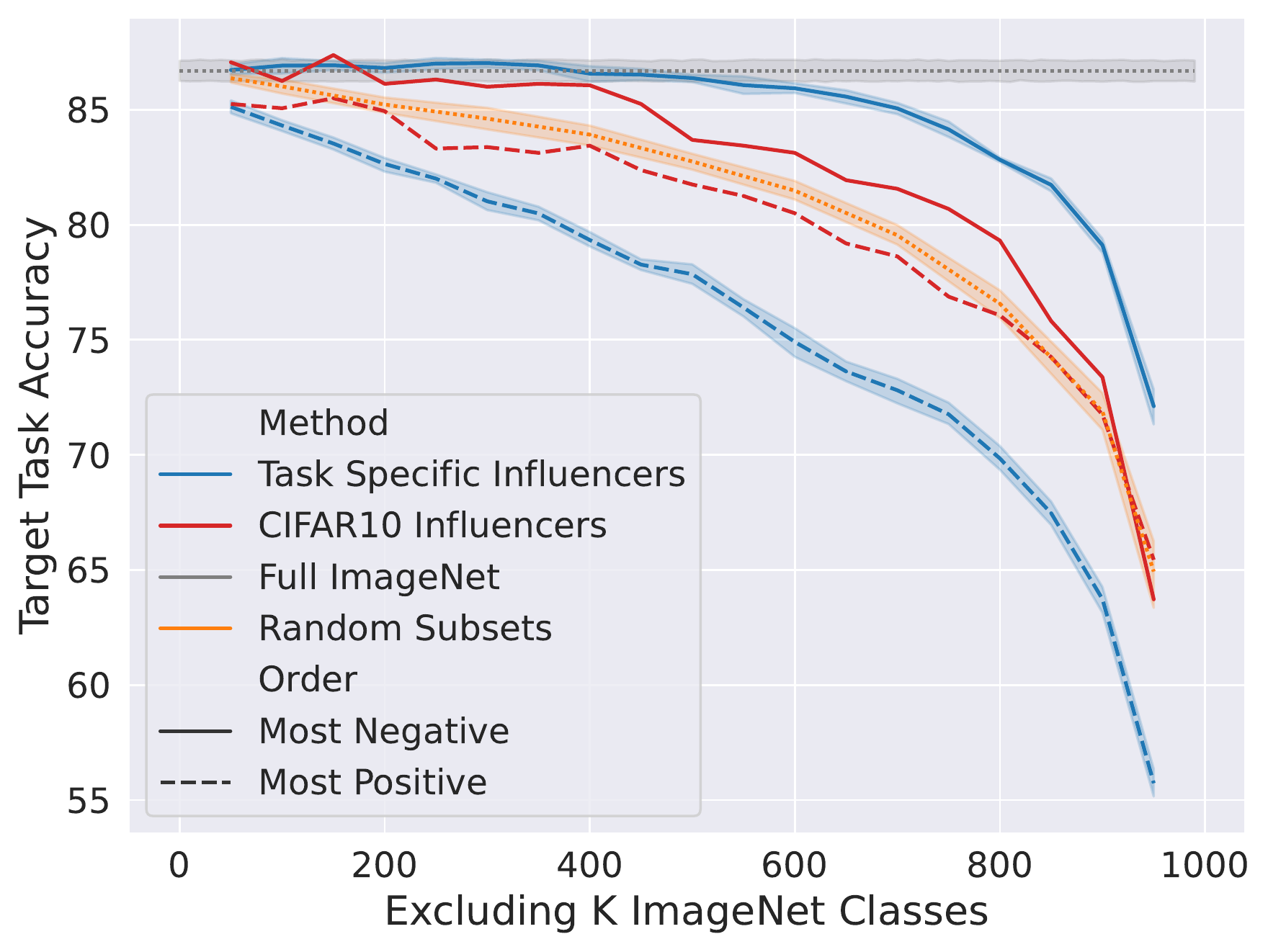}
        \caption{CALTECH101}
    \end{subfigure}\hfill
    \begin{subfigure}[c]{0.32\linewidth}
        \includegraphics[width=0.93\linewidth]{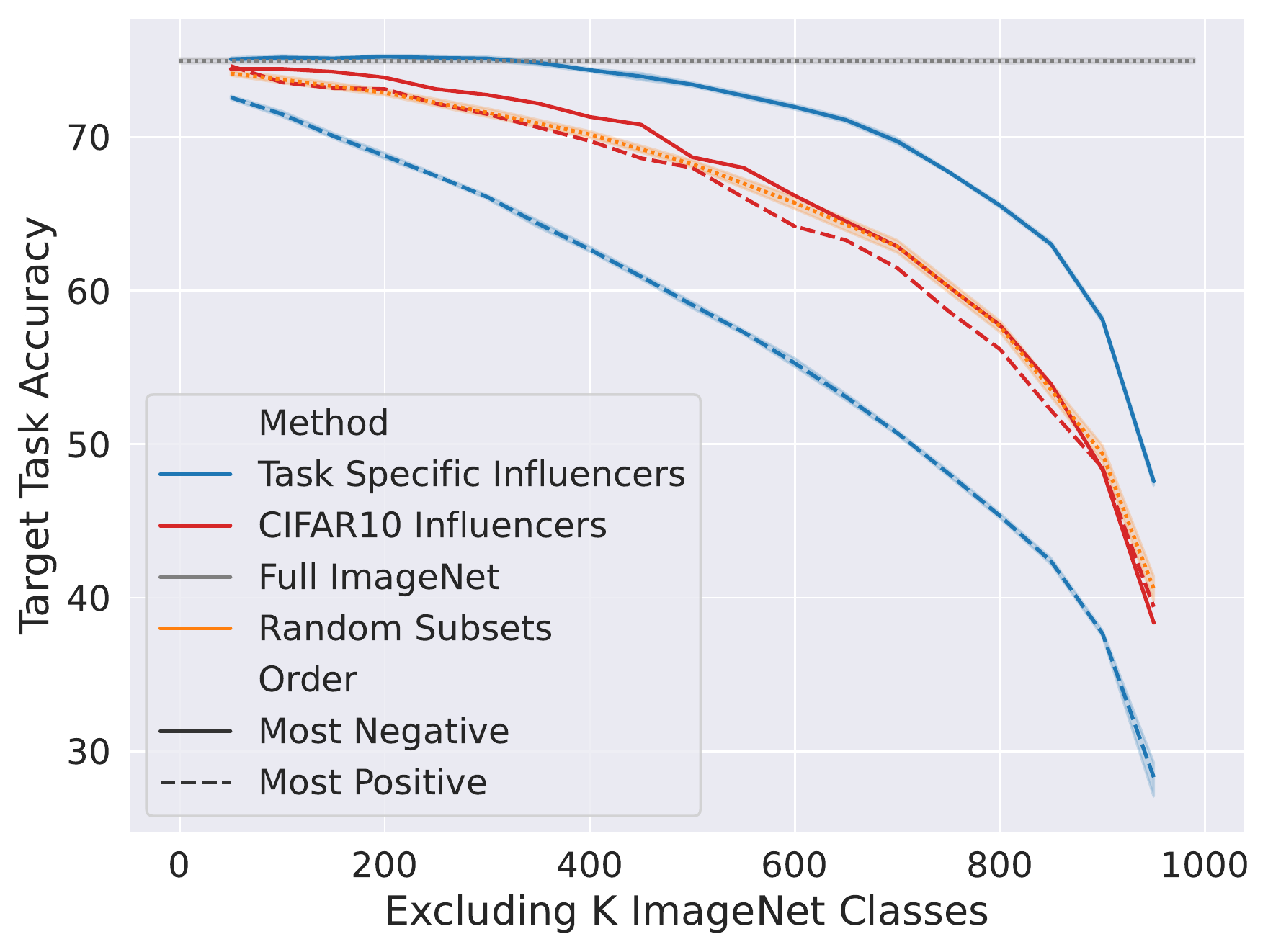}
        \caption{CALTECH2556}
    \end{subfigure}\hfill
    \begin{subfigure}[c]{0.32\linewidth}
        \includegraphics[width=0.93\linewidth]{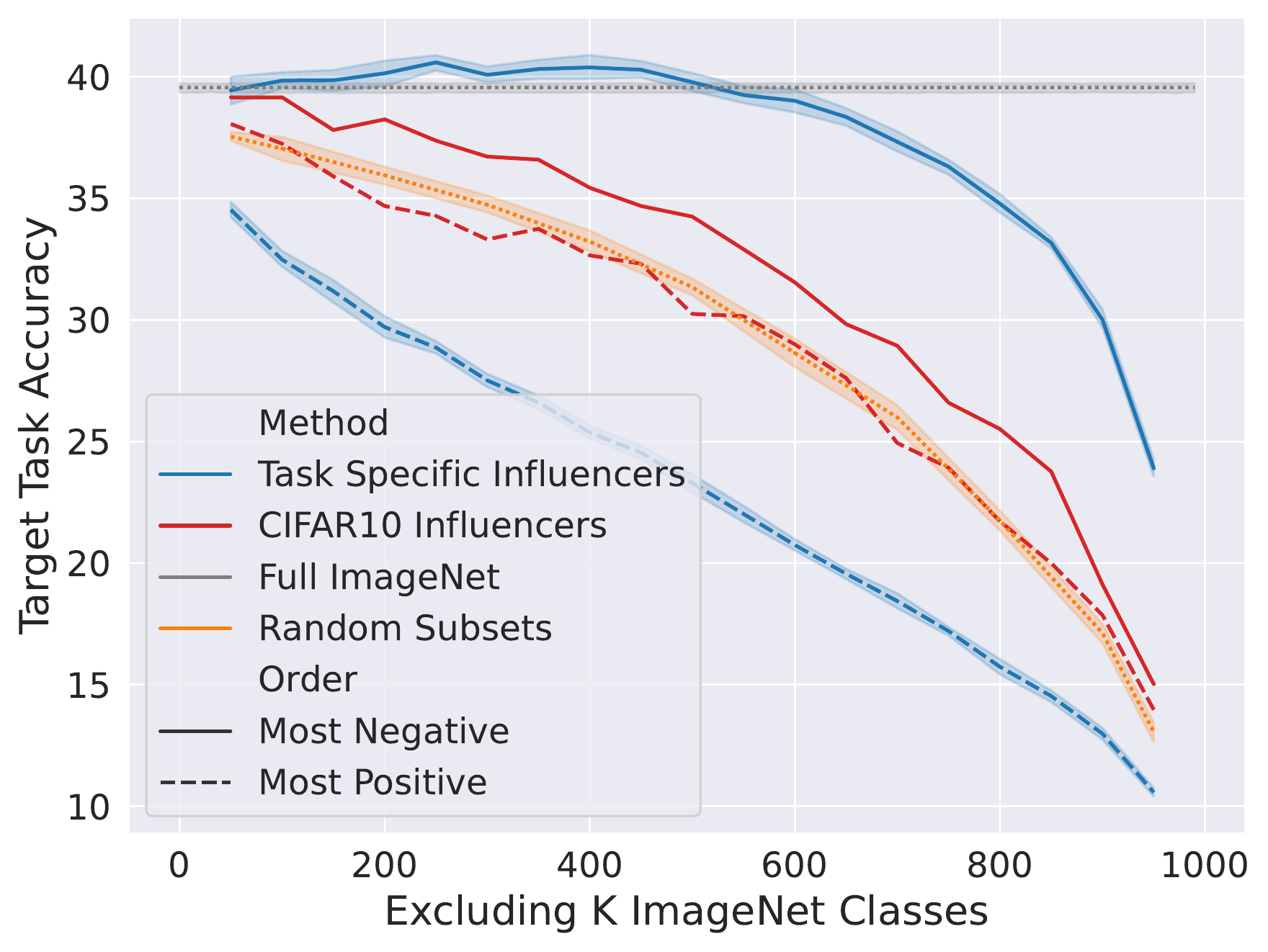}
        \caption{CARS}
    \end{subfigure}\hfill
    \begin{subfigure}[c]{0.32\linewidth}
        \includegraphics[width=0.93\linewidth]{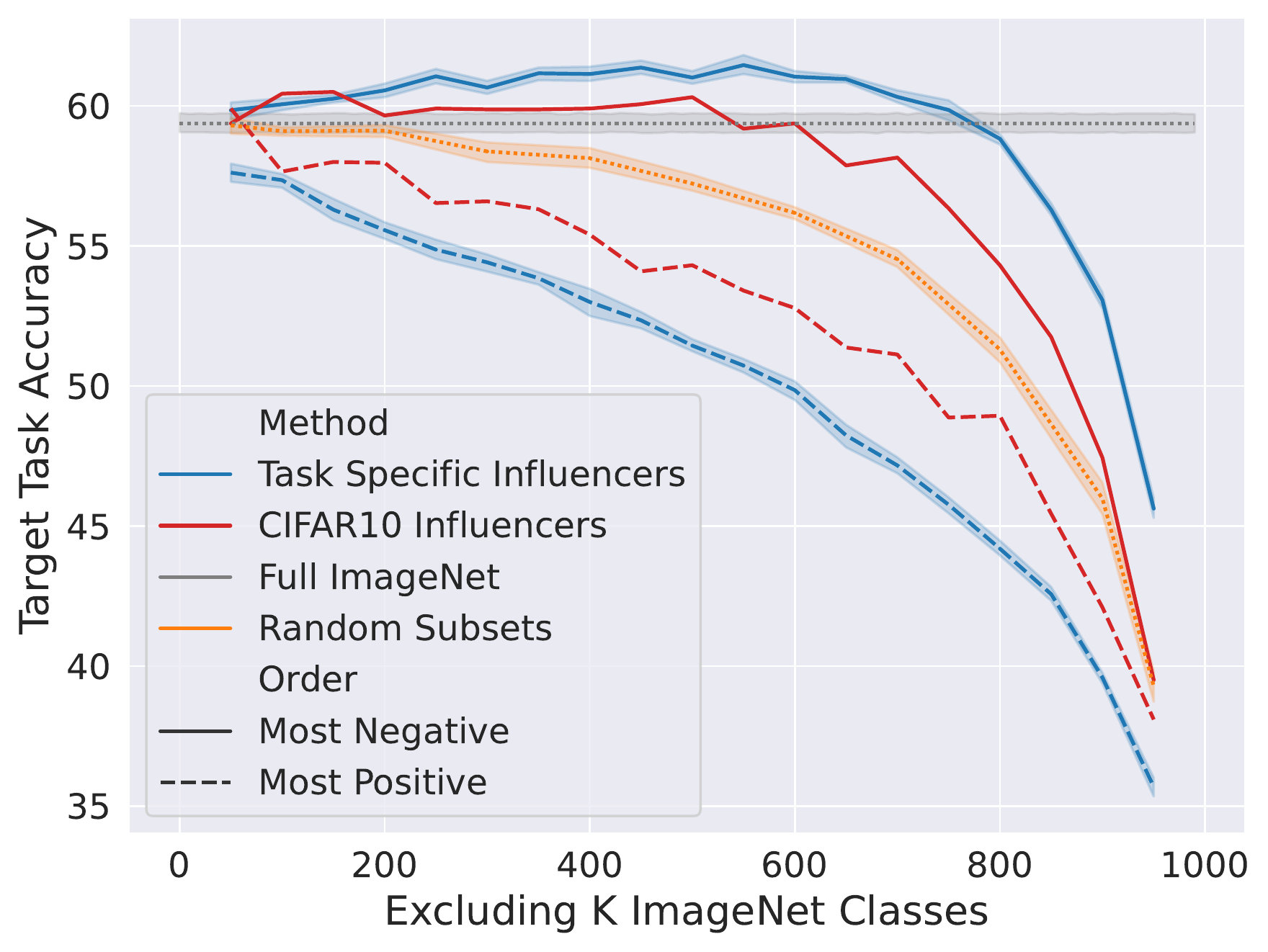}
        \caption{CIFAR100}
    \end{subfigure}\hfill
    \begin{subfigure}[c]{0.32\linewidth}
        \includegraphics[width=0.93\linewidth]{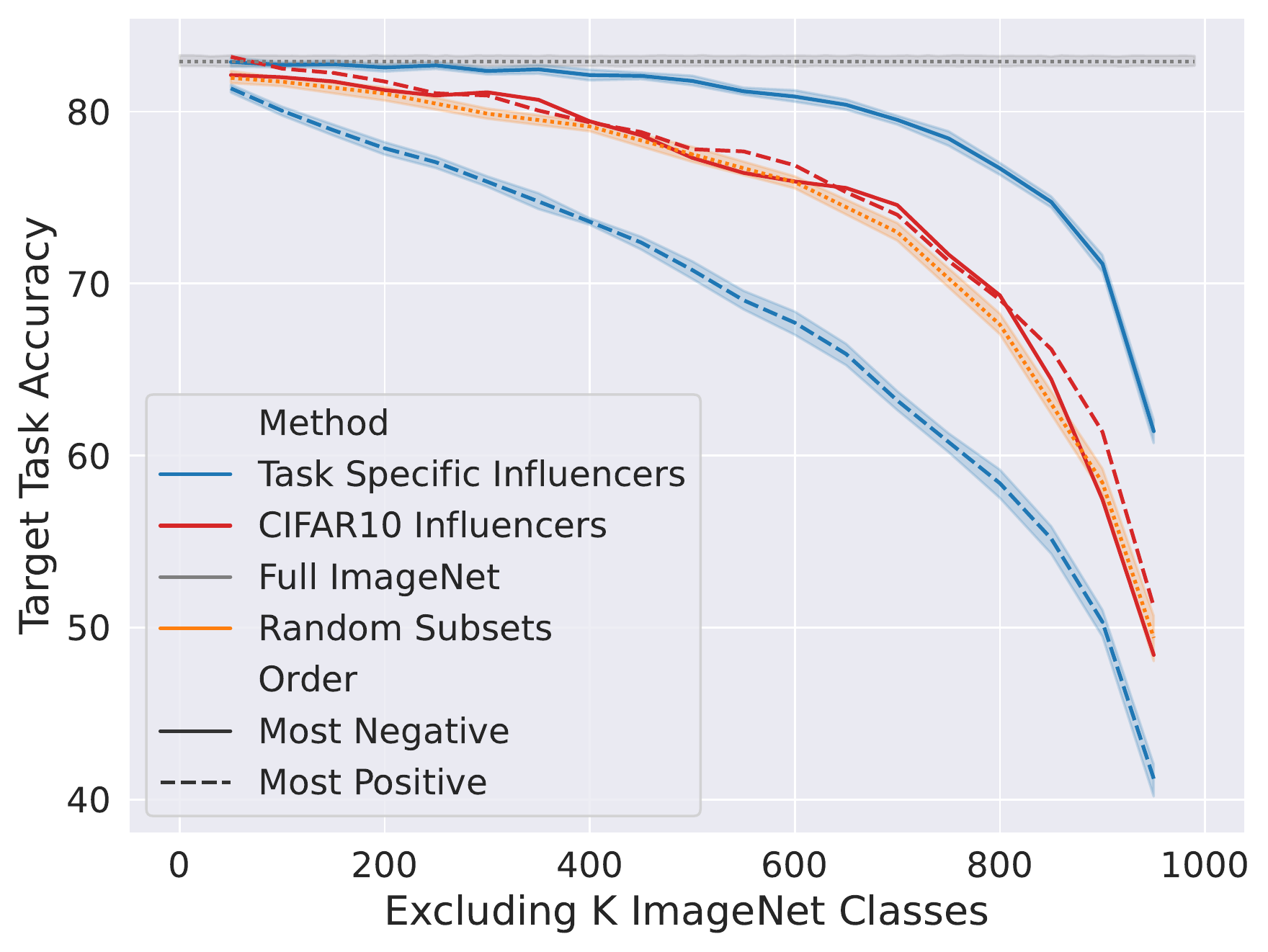}
        \caption{FLOWERS}
    \end{subfigure}\hfill
    \begin{subfigure}[c]{0.32\linewidth}
        \includegraphics[width=0.93\linewidth]{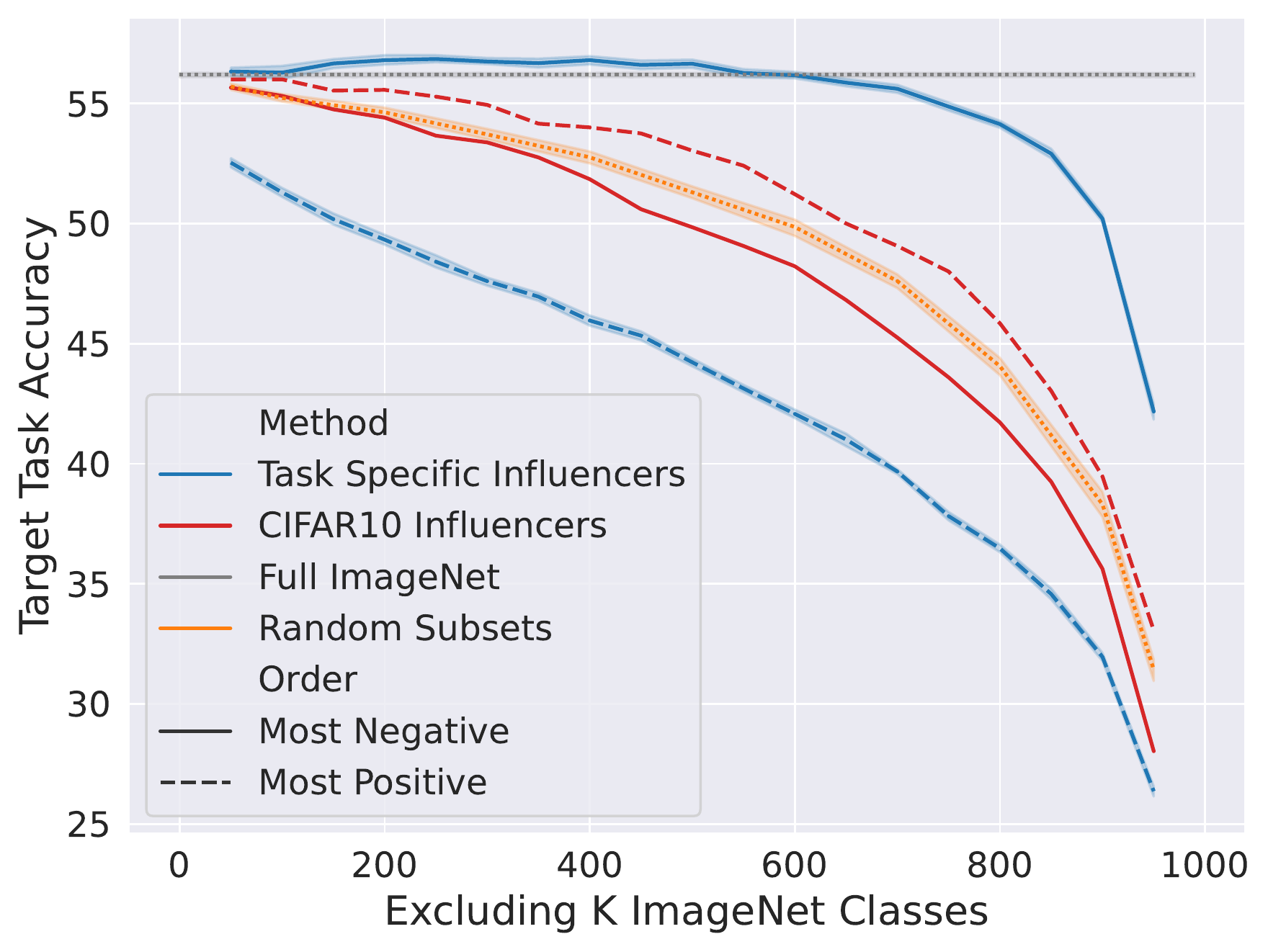}
        \caption{FOOD}
    \end{subfigure}\hfill
    \begin{subfigure}[c]{0.32\linewidth}
        \includegraphics[width=0.93\linewidth]{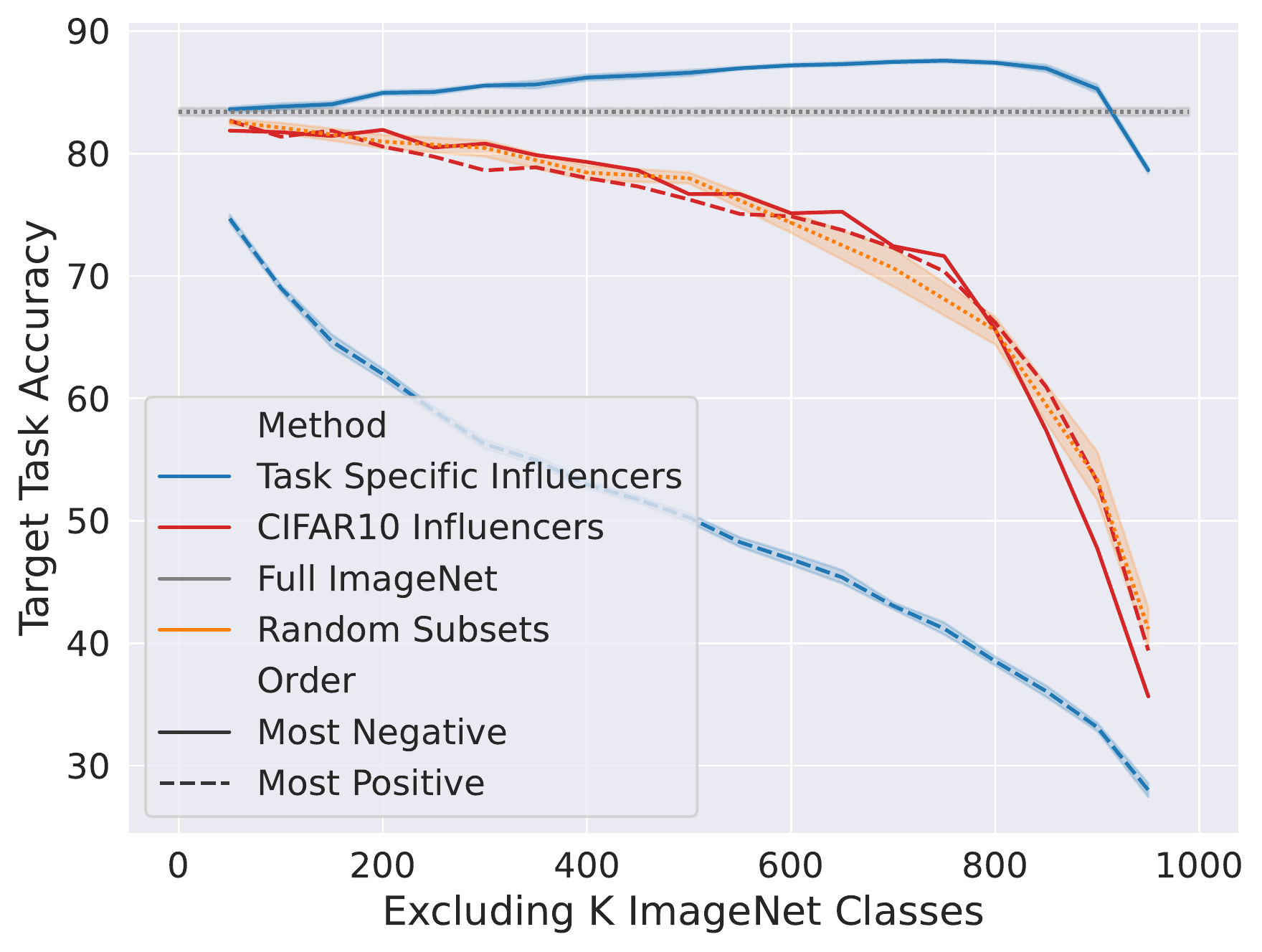}
        \caption{PETS}
    \end{subfigure}\hfill
    \begin{subfigure}[c]{0.32\linewidth}
        \includegraphics[width=0.93\linewidth]{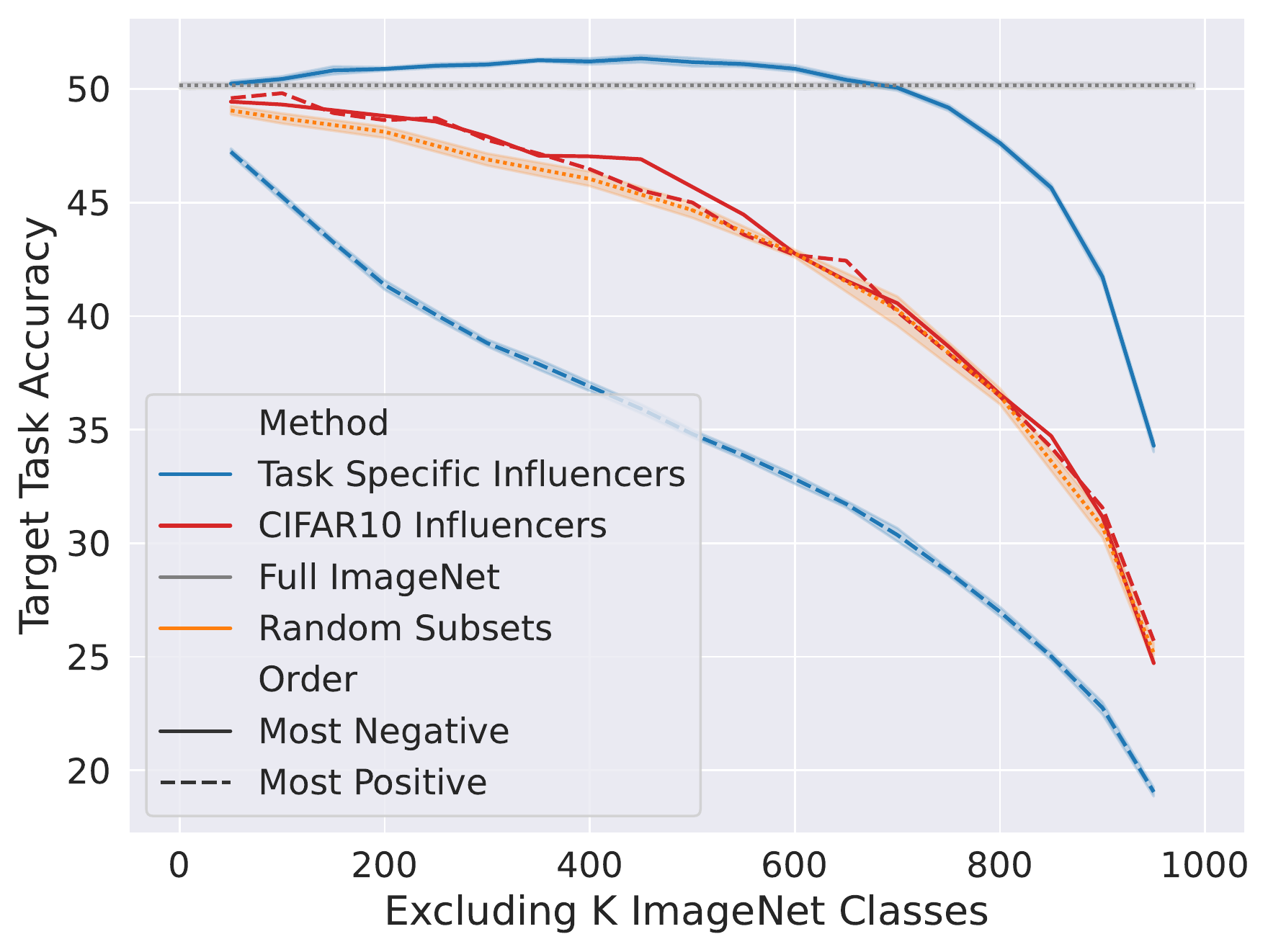}
        \caption{SUN397}
    \end{subfigure}\hfill
    \caption{We repeat the counterfactual experiments (c.f. Figure~\ref{fig:summary_fixed_cf} from the main paper) but using the influence values computed for CIFAR-10 on other target datasets.}
\end{figure}

\clearpage
\subsubsection{Transfer across architectures}
How well do our transfer influences work across architectures? Recall that we computed our transfer influences using a ResNet-18. We now repeat the counterfactual experiment from Figure~\ref{fig:summary_fixed_cf}, using these influences to remove classes from the source dataset when training a ResNet-50. We find that these influences transfer relatively well.

\begin{figure}[h!]
    \centering
    \includegraphics[width=0.5\linewidth]{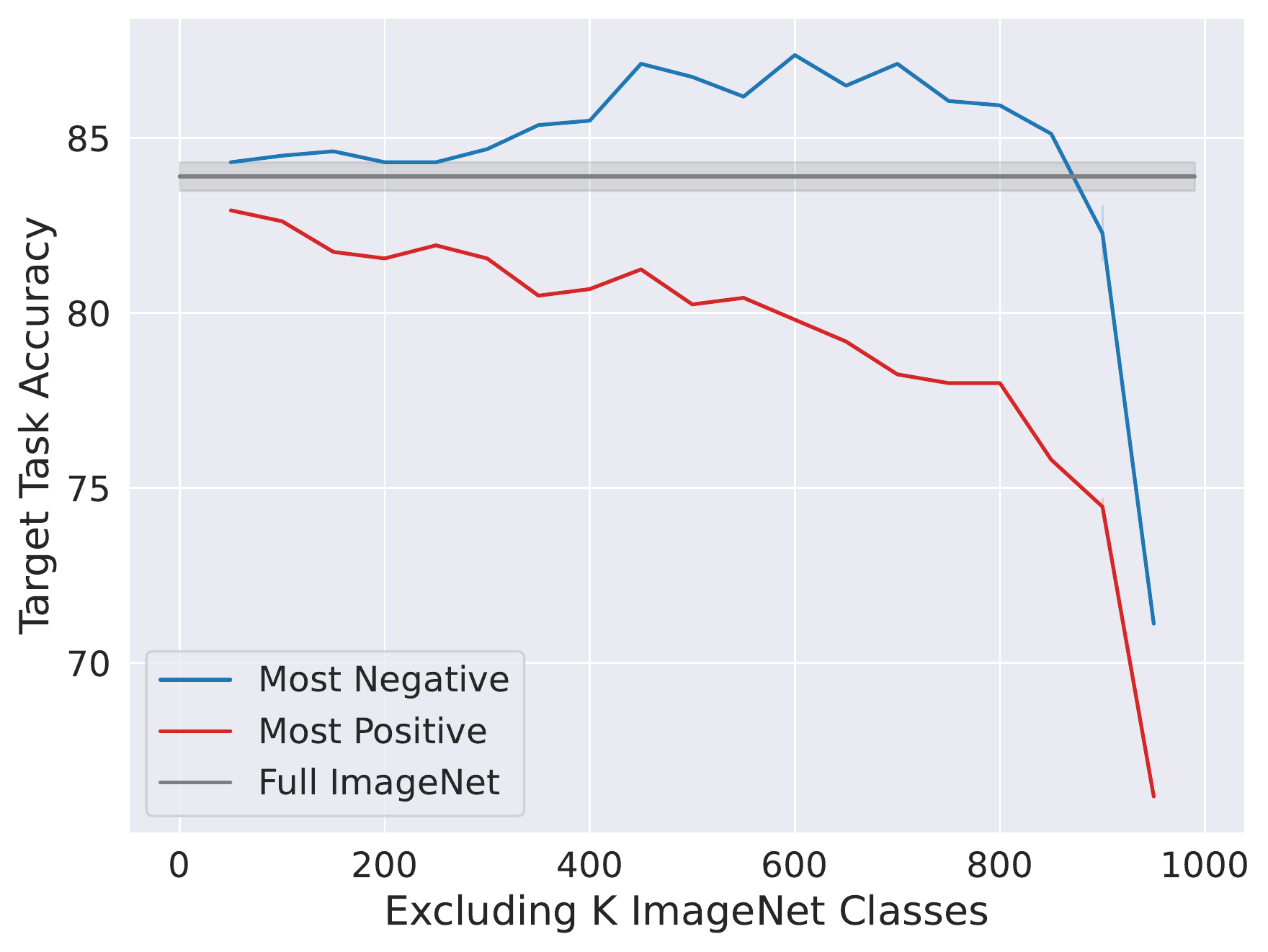}
    \caption{We repeat the counterfactual experiments (c.f. Figure~\ref{fig:summary_fixed_cf} from the main paper) but use our influence values computed using a ResNet-18 on a ResNet-50.}
\end{figure}

%% file: sections/app_convergence.tex
In this section, we analyze the sample complexity of our influence estimation in more detail.
In particular, our goal is to understand how the quality of influence estimates improves with the number of models trained. We also look at the impact of different choices of model outputs.

Instead of directly measuring our downstream objective (transfer accuracy on target dataset after removing the most influential classes), which is expensive, we design two proxy metrics to gauge the convergence of our estimates:

\paragraph{Rank correlation.}
As \citet{ilyas2022datamodels} shows, we can associate influences with a particular linear regression problem: given features $\mathbbm{1}_{\mathcal{S}_i}$, an indicator vector of the subset of classes in the source dataset, predict the labels $f(t; \mathcal{S}_i)$, the model's output after it is finetuned on target dataset $\mathcal{T}$. In fact, we can interpret influences as weights corresponding to these binary features for presence of each class.
That is, the influence vector $w_t = \{\text{Infl}[C_i\rightarrow t]\}_i$ defines a linear function that, given a subset of classes that the source model is trained on, predicts the corresponding model's output when trained on that subset.
Here, we focus on analyzing average model accuracy, so in fact we consider the aggregate output $\sum_i f(t; \mathcal{S}_i)$ and the corresponding aggregated influences $w = \{\text{Infl}[C_i]\}_i$.

Given this view, we can measure the quality of the influence estimates by measuring their performance on the above regression problem on a held-out\footnote{We split the 7,540 models into a training set of 6,000 and a validation set of the remainder.} set of examples $\{\mathcal{S}_i, \sum_i f(t; \mathcal{S}_i)\}$. In order to make different choices of model outputs (logit, confidence, etc.) comparable, we measure performance with spearman rank correlation between the ground truth model outputs and the predictions of the linear model (whose weights are given by the influences).

\begin{figure}[!h]
    \centering
    \includegraphics[width=1\linewidth]{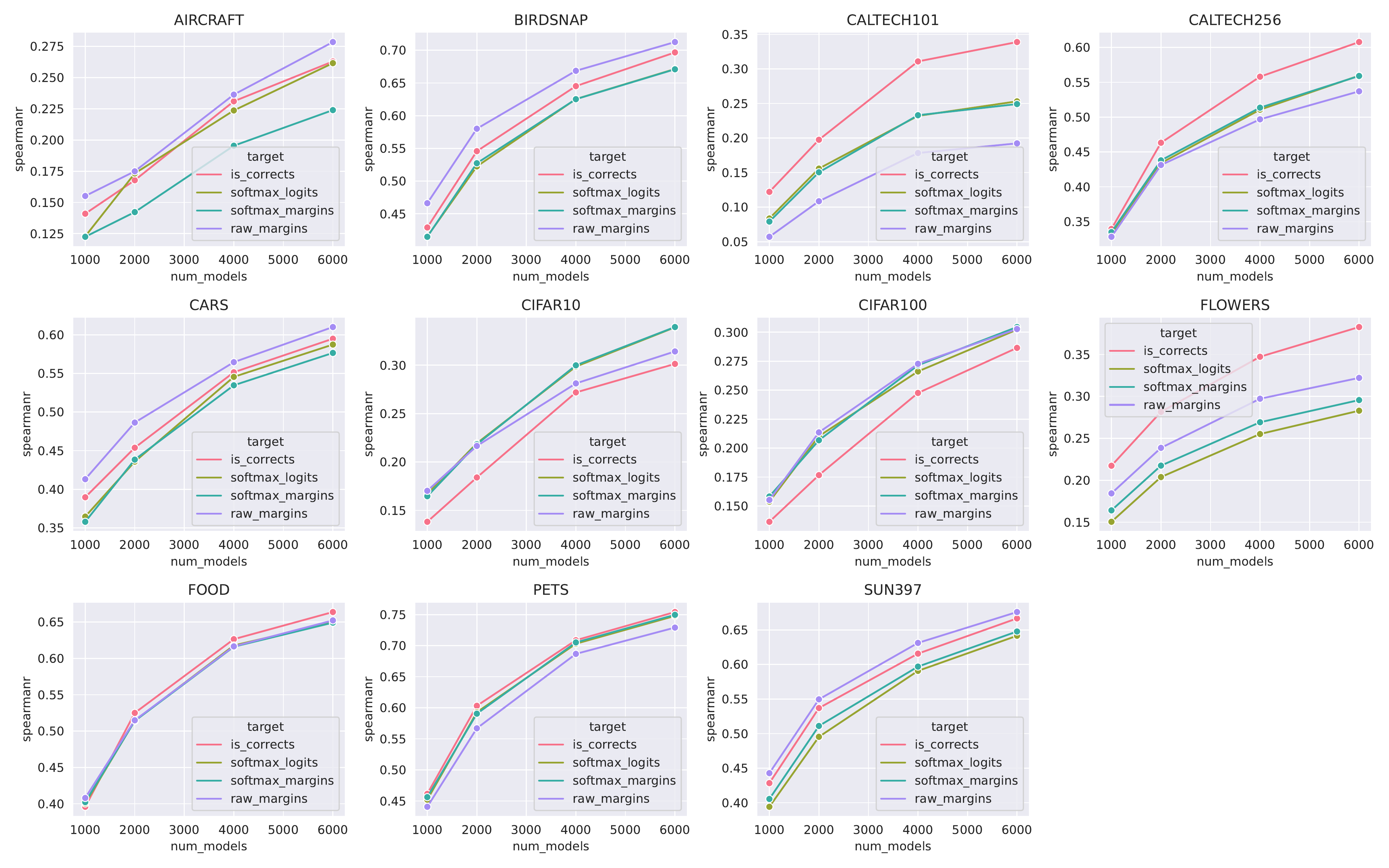}\hfill
    \caption{Measuring improvement in influence estimates using the \textbf{rank correlation} metric. The rank correlation here measures how well influence estimates perform in the underlying regression problem of predicting target accuracy from the subset of classes included in the training set. We evaluate on a held-out set of subsets independent from those used to estimate influences. Across all datasets, correlation improves significantly with more trained models. }
    \label{fig:sample_complexity_correlation}
\end{figure}

We measure this correlation while varying both the number of trained models used in the influence estimation and the choice of the model output, and the results are shown in Figure~\ref{fig:sample_complexity_correlation}.

\begin{figure}[!h]
    \centering
    \includegraphics[width=1\linewidth]{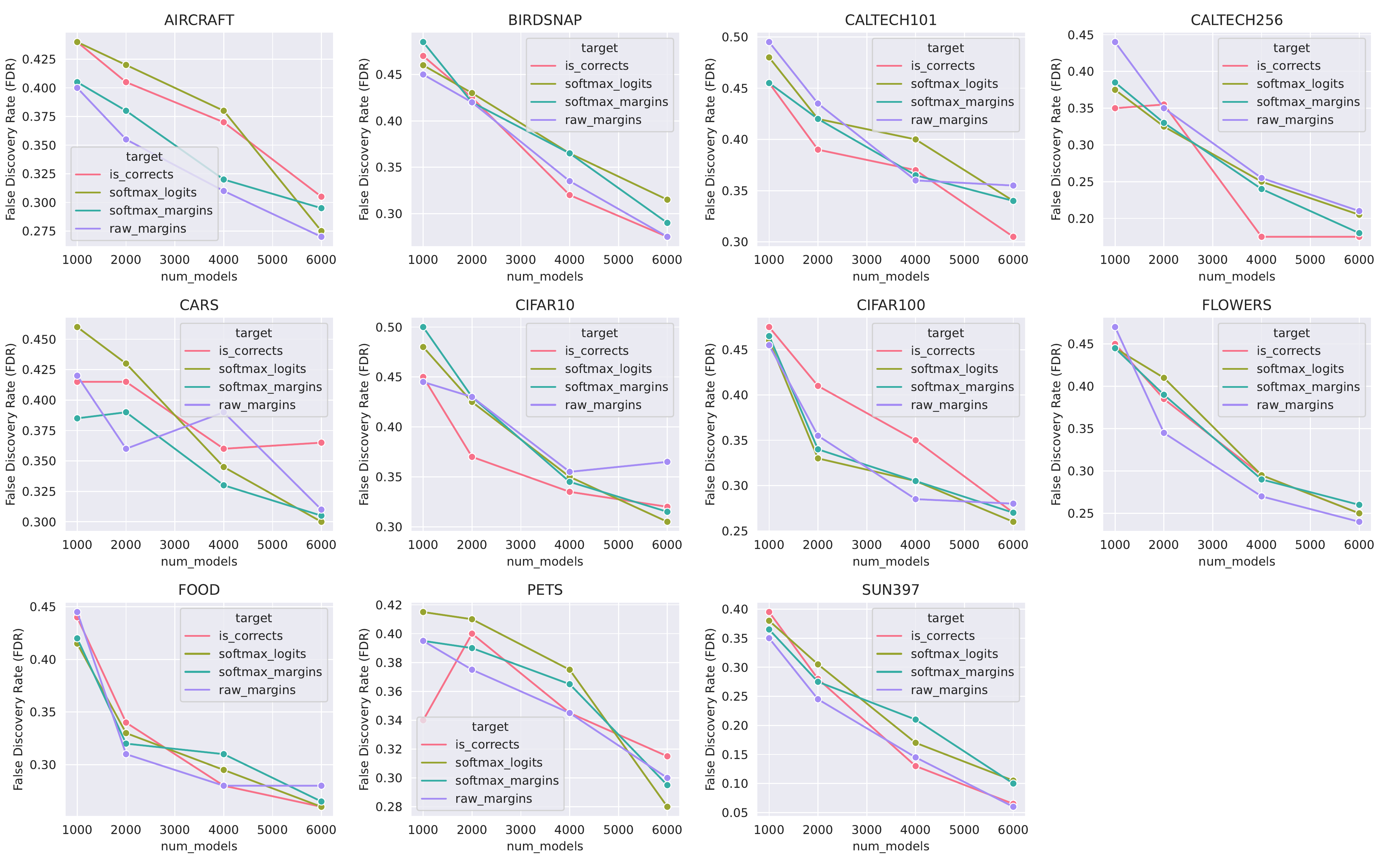}\hfill
    \caption{Measuring improvement in influence estimates using the \textbf{False Discovery Rate} heuristic. Using a procedure (loosely) based on the Knockoffs framework, we estimate the proportion of false discoveries within the top 100 features ranked by esteimated influences. Across all datasets, FDR decreases generally with more trained models.}
    \label{fig:sample_complexity_fdr}
\end{figure}

\paragraph{False discovery rate.}
Above we considered a measure based on predictive performance. Here, we focus on a more parameter-centric notion of False Discovery Rate (FDR). Intuitively, FDR here quantifies the following: how often are the top influencers actually just due to noise?

The Knockoffs \citep{candes2018panning} framework allows one to perform feature selection and also estimate the FDR. At a high level, it consists of two steps: First, one constructs ``knockoff'' versions of the original features which are distributed ``indistinguishably'' (more formally, exchangable) from the original features, and at the same time are independent of the response by design. Second, one applies an estimation algorithm of choice (e.g., OLS or LASSO) to the augmented data consisting of both the original and the knockoff features. Then, the relative frequency at which a variable $X_i$ has higher statistical signal than its knockoff counterpart $\tilde{X}_i$ indicates how likely the algorithm chooses true features, and this can be used to estimate the FDR (intuitively, if $X_i$ is independent of the response $y$, $X_i$ is indistinguishable from its knockoff $\tilde{X}_i$ and both are equally likely to have higher score).

We adapt this framework here (particularly, the verion known as model-X knockoffs) as follows:
\begin{enumerate}
    \item Sample an independent knockoff matrix $\tilde{X}$ consisting of 1,000 binary features from the same distribution as the original mask matrix $X$ (namely, each instance has 500 active features).\footnote{Technically, this procedure is not exactly valid in the original FDR framework as $X_i$ and $\tilde{X}_i$ are exchangable due to depenencies in the features. Nonetheless, it is accurate up to some approximation.}
    \item Estimate influences for both original and knockoff features using the difference-in-means estimator.
    \item Consider the top $k=100$ features by positive influence, and count the proportion of features that are knockoff. This yields an estimate of FDR among the top 100 features. An FDR of 0.5 indicates chance-level detection.\footnote{This is different from the usual manner of controlling the FDR, but we look at this alternative metric for simplicity.}
\end{enumerate}

As with the previous metric, we measure the above FDR for each target dataset while varying the number of trained models and the target output type (Figure~\ref{fig:sample_complexity_fdr}).

\paragraph{Discussion.}
We observe the following from the above analyses using our two statistics:
\begin{itemize}
\item There are significant gains (higher correlation and lower FDR) with increasing number of trained models.
\item But neither metric appears to have plateaued with 6,000 models, so this indicates that we can improve the accuracy of our influence estimates with more trained models, which may in turn improve the max improvement in transfer accuracies (Section~\ref{sec:counterfactuals}), among other results.
\item The choice of the target type does not appear to have a significant or consistent impact across different datasets, which is also consistent with the results of our counterfactual experiments (Appendix \ref{app:vary_model_outputs}).
\end{itemize}